%% file: ms.tex
\documentclass{new_tlp}
\usepackage[latin1]{inputenc}
\usepackage{aopmath}
\usepackage{amsfonts}
\usepackage{amssymb}
\usepackage{graphicx}
\usepackage{pgfplots}
\usepackage{varwidth}
\usepackage{lscape}
\usepackage[boxed]{algorithm2e}
\usepackage{fancyvrb}

\usepackage[english]{babel}
\usepackage{color}
\usepackage{lineno}

\newtheorem{definition}{Definition}[section]
\newtheorem{mytheorem}{Theorem}[section]

\newcounter{example}[section]

\newcommand{\mytab}{\mbox{\ \ \ \ \ }}

\newcommand{\R}{\mbox{$\cal{R}$}}
\newcommand{\C}{\mbox{$\cal{C}$}}
\newcommand{\Aux}{\mbox{${\cal A}ux$}}

\author[Broda, Sadri and Butler]{KRYSIA BRODA \and FARIBA SADRI\and STEPHEN BUTLER\\
Imperial College London\\
\email{\{k.broda,f.sadri,stephen.butler14\}@imperial.ac.uk}}
\title{Reactive Answer Set Programming}
\begin{document}
\maketitle  	
%
\begin{abstract}		
Logic Production System (LPS) is a logic-based framework for modelling reactive behaviour. Based on abductive logic programming, it combines reactive rules with logic programs, a database and a causal theory that specifies transitions between the states of the database. This paper proposes a systematic mapping of the Kernel of this framework (called KELPS) into an answer set program (ASP).
For this
purpose a new variant of KELPS with finite models, called $n$-distance KELPS, is introduced.  A formal definition of the mapping from this $n$-distance KELPS to ASP is given and  proven sound and complete.  The Answer Set Programming paradigm allows to capture additional behaviours to the basic reactivity of KELPS, in particular   proactive, preemptive and prospective behaviours. These are all discussed and illustrated with examples. Then a hybrid framework is proposed that 
integrates KELPS and ASP, allowing to combine the strengths of both paradigms.
\end{abstract}
\input{Intro_Final_corr}

\input{Background_Final_corr}

\input{Translation_Final_corr}

\input{Formal_Final_corr}

\input{Remainder_Final_corr}
\input{Hybrid_corr}

\input{Related_Final_corr}

\section*{Acknowledgements}
We thank the anonymous reviewers for their helpful and insightful comments. 
We  also thank Bob Kowalski for useful discussions on an early draft of the paper. 
\bibliographystyle{acmtrans}
\bibliography{reactivebibliography_v2}
\appendix
\input{Iclingo_appendix_corr}

%
\end{document}

%% file: Intro_Final_corr.tex
\section{Introduction}\label{sec:intro}
Reactivity plays a major part in many areas of computing. 
For instance, it is an important feature in situated agent systems  and it forms the foundation of many state transition systems.
 It also plays a part in constraint handling rules  and abstract state machines   and reactive programming in general~\cite{CIFF,Early,DALI,Alferes,Agentspeak,fruhwirth,gurevich}. Reactivity can take several different forms, such as event-condition-action rules, for instance in active databases, or condition-action rules, for example in production systems,  or transition rules in abstract state machines~\cite{activedatabases,Statelog,Fernandes,Russell,ASMs}. 
Reactivity is implicit in some systems, such as BDI agents, whereas in other systems it is explicit and core, for example Reaction RuleML and LPS~\cite{BDI,RuleML,LPSAbductive,LPSReactive}.
Consider, for example, an agent situated in an environment. The agent may have some initial goals towards which it may plan and execute actions. But, to be effective, it also needs to take account of the changes in its environment and react to them by setting itself new goals and adjusting its old goals and any already constructed (partial) plans. 
This is also in the spirit of   (Teleo)reactive systems, which have the primary objective  to  increase the responsiveness and resilience of computer systems without the need for replanning~\cite{Keith1,Keith2,Nilsson,Sanchez}.

ASP (Answer Set Programming) \cite{ASP,ASP_Glance} is a paradigm of declarative programming, rooted in logic programming, that has gained popularity in recent years and has been applied in several interesting domains, such as planning,  semantic web, computer-aided verification and health care~\cite{applications}.
Given a logic program, ASP computes its models, called answer sets, which can be considered solutions to the problem captured by the logic program.

In this work we show how reactivity can be incorporated within the ASP  paradigm by adopting the notion of reactivity from another logic-based paradigm, KELPS (KErnel of LPS) \cite{kelps_paper}. 
We call the resulting framework {\em Reactive ASP}.
KELPS (and  LPS)  is based on abductive logic programming and combines reactive rules with logic programs, a database and a causal theory that specifies transitions between the states of the database. 
We chose KELPS as the basis for Reactive  ASP   for two  reasons. 
Firstly, the logic-based syntax of KELPS  and its notion of reactivity are, in our opinion, quite general and  intuitive {\em--} for example they subsume both event-condition-action rules and condition-action rules.
Secondly, KELPS provides a formal definition of reactivity which  guided our design of Reactive ASP and  its formal verification.

\begin{figure}[htbp]\label{fig1}
\begin{center}
\vspace{-0.3cm}
 \includegraphics[width=11cm, height=7cm]{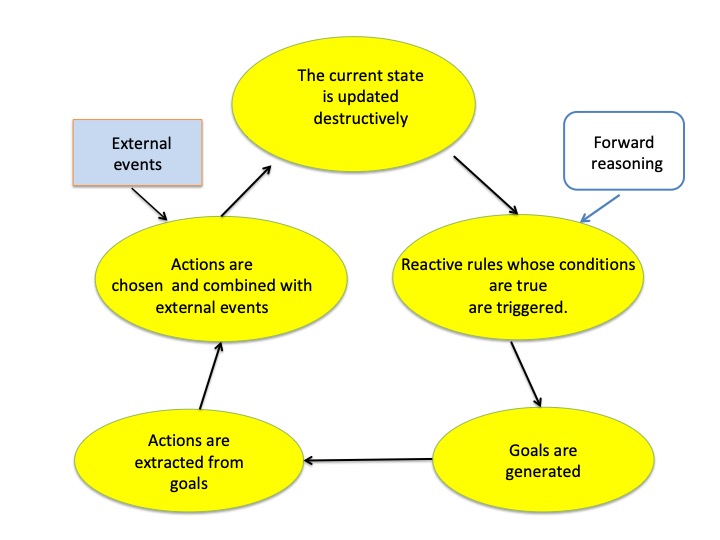}
\end{center}
\vspace{-0.5cm}
\caption{KELPS operational semantics  cycle.}
\end{figure}

The operational semantics of KELPS is based on the cycle illustrated in Figure~\ref{fig1}. Starting from an initial state (database), 
incoming external events and the framework's own generated actions are assimilated and the state is updated, and reactive rules whose conditions have become true given the history of events and states so far are triggered. This  results in new goals to satisfy, and their incremental partial solutions, in turn, result in more actions  generated to be executed, thus iterating through the cycle.
As well as an operational semantics KELPS has a model-theoretic 
semantics, and KELPS computation is aimed at generating models for the 
reactive rules in dynamic environments. These models can be infinite.

The paper makes several contributions:
\begin{itemize}
\item
 It shows how to facilitate reactivity in ASP by first defining a variant of KELPS with a finite number of state changes in any model, called $n$-distant KELPS, and  then giving a mapping from this $n$-distant KELPS to Reactive ASP. We prove that the mapping is sound and complete in the sense that, given any initial state and any ensuing sets of external events, the answer sets of the resulting Reactive ASP programs correspond exactly to the $n$-distant KELPS reactive models.
\item
The mapping is then used to shed light on 
 possible relationships and  synergies between the two different paradigms.
In particular, we show  that it facilitates for free additional alternative control strategies and a \textit{prospective} style of programming \cite{prospective}, which allows  to consider future consequences of present decisions in order to choose what to do presently. 
\item
Furthermore, we propose an approach that combines features of KELPS with ASP into one unified 
architecture that enjoys the benefits of KELPS operational semantics and its destructive updates of the state where no frame axiom reasoning is needed, together with the flexibility of ASP that allows prospective and preference reasoning. 
\end{itemize}
Notions of reactivity have  been considered in ASP in other work
but these approaches are different from the form presented here.
For example, in a tool for maritime traffic control, Vaseqi and Delgrande  \cite{maritime} explain how they used oclingo,  a particular form of 
reactive ASP that handles external data, now superseded by \textit{clingo} 4 \cite{oclingo},
 in order to handle histories efficiently inside the ASP system. Thus the work in generating answer sets  at each time-step is reduced by discarding information from previous time-steps that
is no longer useful.  
In  \cite{ribeiro}     the notion of reactivity in ASP is used to mean efficiency in reasoning by partitioning the knowledge base so that reasoning is done only with the part relevant to the context, whereas 
Brewka, at the end of his paper  \cite{Brewka}, sketches a notion of reactivity closer to ours. He  proposes as future work the use of rules that have `operational statements in their heads', where these operations are `read off' from the generated answer sets and the program is modified accordingly.
Recently, some applications using iclingo (an ASP system that 
incorporates incremental grounding)  to simulate reactivity have been described in~\cite{Recenticlingo}. 

The rest of the paper is structured as follows. Section~\ref{sec:background} provides background on KELPS and ASP. Section~\ref{sec:translation} describes and exemplifies the mapping between the two, while Section~\ref{sec:theory} provides theoretical results.
Section~\ref{sec:extensions} discusses how Reactive ASP allows  more complex control strategies and   prospection. In Section~\ref{sec:combined} we then propose a hybrid of the two 
paradigms of KELPS and ASP that combines their advantages. We illustrate 
this hybrid system with examples and discuss two different ways of realising it. 
Section~\ref{sec:related}  reviews the mappings discussed earlier in the paper and 
provides further insights in the comparison of the incremental behaviour 
of KELPS and Reactive ASP.  It then briefly introduces an alternative 
incremental mapping to ASP and its implementation in \textit{clingo} 4,   and gives a brief empirical  evaluation of our approach (with more detail in Appendix B). It finally discusses related work by looking 
at other approaches to reactivity and prospection in logic programming.
In  Section~\ref{sec:conclusion} we discuss future work and conclude.

%% file: Background_Final_corr.tex
\section{Background}\label{sec:background}
This section  introduces the KELPS framework  and  reviews relevant features of Answer Set Programming.

\subsection{KELPS}\label{sec:KELPS}
KELPS \cite{kelps_paper}
 is a logic-based state transition framework combining reactivity with a destructively updated database and a causal theory
that has both an {\it operational semantics} and a {\it model-theoretic semantics}.
KELPS is a subset of the full LPS language \cite{LPSAbductive}.  
LPS has been implemented in Prolog, Java, Python and SWISH and has been used in currently ongoing industry-based applications in smart contracts and industry production control.
The SWISH proptotype implementation of LPS \cite{SWISH}
is downloadable and together with LPS examples  can be found at 
\texttt{http://lps.doc.ic.ac.uk}.

\subsubsection{The KELPS Vocabulary}\label{sec:vocabulary}
KELPS uses a first-order sorted language including a sort for linear and discrete time, in which
the   predicate symbols (and consequently atoms) of the language are partitioned into sets representing fluents, events, auxiliary predicates and
meta-predicates.
\textit{Fluent predicates} are used to represent time-dependent facts in the KELPS state. In their time-stamped form, $p(t_1,...,t_n, i)$, their last argument $i\geq 0$ represents the time  of the state $S_i$ to which the fluent belongs. The  atom $p(t_1, ..., t_n)$ is called the unstamped fluent.
\textit{Event predicates} capture  events, both observed external events and events generated by the framework itself (sometimes called {\em actions} to distinguish them).
Events contribute to state transitions {\em--} that is, they map one state into a successor state. 
In their timestamped form, $e(t_1,...,t_n, i)$, their last argument $i\geq 1$ represents the time of the (successor) state
$S_i$ and the event is said to take place in the transition between state $S_{i-1}$ and $S_i$. The  event atom $e(t_1, ..., t_n)$ is called the unstamped event.\footnote{
For ease of notation we sometimes write $p(i)$ and $e(i)$ in place of $p(t_1,...,t_n, i)$ and $e(t_1,...,t_n, i)$.}
\textit{Auxiliary predicates} are of two kinds: (i) \textit{time-independent predicates} (and corresponding atoms) do not include time parameters, and represent properties that are not affected by events, e.g. $isa(book, item)$, denoting that book is an item; and (ii) \textit{temporal constraint predicates} (and corresponding atoms) use only time parameters in arguments and express temporal constraints, including inequalities of the form $T1 < T2$ and $T1 \leq T2$ between timepoints, and functional relationships among timepoints, such as $max(T1,T2,T)$,
denoting that $T$ is the maximum  of $T1$ and $T2$. 
The KELPS \textit{meta-predicates}  \textit{initiates(event, fluent)} and \textit{terminates(event, fluent)} express  the fluents that are initiated and terminated by events.
In general the first argument would be a set of events, to cater for cases where a set of events may have a different impact on state changes from the sum of the effects of each of the constituent events.
In our mapping to ASP we  define these meta-predicates  for single events only.
This caters for all cases where concurrent events are {\em independent}  of each other, that is they do not affect 
the same fluent.

\subsubsection{The KELPS Framework}\label{sec:Framework}
The KELPS framework is specified by a tuple $<\R,\C,\Aux>$ consisting of a set \R~of
reactive rules, a causal theory  \C~specifying preconditions and postconditions of the events that cause state transitions, and a set \Aux~of 
auxiliary  ground atoms.
A reactive rule has the logical form
\begin{equation}\label{eq1}
\forall \overline{X}[antecedent(\overline{X})\rightarrow \exists \overline{Y} consequent(\overline{X1},\overline{Y})]
\end{equation}
\noindent
in which the $consequent$ is a disjunction  $consequent_1  \lor   \ldots \lor consequent_n $, 
the $antecedent$ and each $consequent_{i} $ is a conjunction of \textit{conditions}, where each condition is either a {\em fluent literal}, an {\em event atom},  or an {\em auxiliary literal}.\footnote{In KELPS a condition may also be a {\em state condition} which is a FOL formula involving fluents and auxiliary atoms. In this paper we ignore this generality.}
In Equation~(\ref{eq1}) $\overline{Y}$ is the set of all variables that occur only in $consequent(\overline{X1}, \overline{Y})$, $\overline{X}$ is the set of  remaining variables in the rule and $\overline{X1} \subseteq \overline{X}.$\footnote{Throughout the paper variables start in upper case and a set of variables  is represented as $\overline{X}$.} 
Note that because of the restrictions of the quantification of variables in reactive rules we can omit the quantifier prefixes without ambiguity and write	
$antecedent(\overline{X})\rightarrow consequent(\overline{X1},\overline{Y}) $.
All timestamps in  $consequent$   are equal to, or later than,  all timestamps in $antecedent$.

For example, consider the following policy:
\textit{If a customer (Cust) makes a request for an item (Item) at time $T$, then either the item is available and the agent  allocates the item to the customer at some time $T_1$ later than $T$, and then $processes$ the order, all to be done before 4 units of time after $T$, or the agent   apologises about the item to the  customer at 4 units of time after $T$. Moreover,		
if an item is allocated to a customer, there are fewer than 2 units of that item left afterwards and the item is not already on order, then the agent must order 20 units of it at the next time unit.} 
This can be expressed as two reactive rules in KELPS:
\begin{equation}\label{eq2}
\begin{array}{l}
request(Cust, Item, T)  \rightarrow \\
\mytab [( avail(Item, N, T_1)\wedge allocate(Cust, Item, N, T_2)\wedge T_2=T_1+1 \\
\mytab\mytab \wedge process(Cust, Item, T_3)\wedge T < T_2 < T_3 < T+4)\\
\mytab\mytab \vee (apologise(Cust, Item, T_4) \wedge T_4 = T+4)]\\ \\
allocate(Cust, Item, N,T) \wedge avail(Item, N1, T)\wedge N1 < 2 \\
\mytab \wedge \neg\ on\_order(Item, T) \rightarrow order(Item, 20, T_1) \wedge T_1=T+1
\end{array}\end{equation}
The antecedent of reactive rules can refer to a history of events and states, in a similar way that the consequent can refer to a plan\footnote{We use the term (planning or) plan to mean simply (the generation of) a course of actions that together with external events and resulting states would make reactive rules true.} to be made true over time and several states. 
For example, consider the following requirement for applicants to a degree programme: \textit{an applicant to a degree programme who is offered a place and then accepts the offer must be placed on the pending list immediately and be sent an invoice within 30 days of accepting the offer}. In KELPS:
\begin{equation}\label{eq3}
\begin{array}{l}
[apply(A, Prog, T_1) \wedge offer(A, Prog, T_2) \wedge accept(A, Prog, T)\\
\mytab  \wedge  T_1 < T_2 < T ] \rightarrow\\
\mytab [ add\_pending(A, Prog, T_4) \wedge T_4 = T + 1  \\
\mytab \wedge send\_invoice(A, Prog, T_5) \wedge T_4<T_5  \leq T +30]
\end{array}\end{equation}
Computation in KELPS involves the execution of actions in an attempt to make reactive rules true in a canonical model of the logic program determined by an initial state, sequence of events, and the resulting sequence of subsequent states. The causal theory \C, comprising  $C_{post}$ and $C_{pre}$,
specifies the state transformations caused by the events.
$C_{post}$ uses the meta-predicates $terminates$ and $initiates$ to specify the postconditions of events, 
and $C_{pre}$ is a set of integrity constraints restricting the occurrence and co-occurrence of sets of events. 
These constraints take the form $false \leftarrow body$, where $body$ is a conjunction that will include at least one event atom, and may include fluent and auxiliary literals. All the event atoms will have the same variable  or constant timestamp, and all the fluent literals will have the same variable or constant timestamp, but one unit before the common timestamp of the events.

For example, the postconditions  below specify that  \textit{whenever an item is allocated to a customer the available stock count ($N$) of the item is decremented}, and that whenever an item is ordered (for the stock) it is $on\_order$.
The constraints specify  that \textit{an item cannot be allocated   if it is
out of stock (its quantity is 0)},\footnote{Note that here we follow KELPS notation of using the same identifier for the term representing a fluent or an event and the predicate representing the holding of the fluent and the happening of the event.} 
nor can \textit{the same item be  allocated  at the same time  to two different customers.}

\begin{equation}\label{eq4}
\begin{array}{l} C_{post}:\\
initiates(allocate(Cust, Item, N) ,avail(Item,N-1)) \\
terminates(allocate(Cust, Item,N), avail(Item, N) )\\
initiates(order(Item, N), on\_order(Item)) \\ \\
C_{pre}: \\
false \leftarrow allocate(Cust, Item, N, T+1) \wedge avail(Item, 0, T)\\
false \leftarrow allocate(Cust1, Item, N1, T) \wedge allocate(Cust2, Item, N2, T) \\
 \mytab \wedge Cust1 \neq Cust2 
\end{array}
\end{equation}

\subsubsection{The KELPS Operational and Model Theoretic Semantics}\label{sec:semantics}
Recall the operational semantics (OS) of KELPS from Figure~\ref{fig1}. The OS monitors the stream of states and incoming and self-generated events and actions, to determine whether an instance of a reactive rule antecedent has become true.
For all such true instances  the instances of the consequents are generated as goals to be satisfied in future cycles. The OS attempts to make these goals true by executing actions, that, in turn, change the state.
KELPS keeps a record only of the latest events and states; new states replace the ones they succeed. Because of this the antecedents of reactive rules are processed incrementally with the incoming streams of events and changes of state. To illustrate this point consider the reactive rule in Equation~(\ref{eq3}), in which, more realistically, there is a deadline of 30 days after an offer for accepting it, which is added to the antecedent of the reactive rule. Suppose John applies for MSc at time 1. The reactive rule will provide a residue
\begin{equation}\label{eq3a}
\begin{array}{l}
[offer(john, msc,T_2) \wedge accept(john,msc, T) \wedge T\leq T_2+30
\wedge  1< T_2 < T ]\\
\mytab  \rightarrow[add\_pending(john,msc, T_4) \wedge T_4 = T + 1 \\
\mytab \mytab \wedge send\_invoice(john,msc, T_5) \wedge T_4<T_5  \leq T +30]
\end{array}\end{equation}
If now John is offered a place on the MSc at time 3, then this residue will be further processed to
\begin{equation}\label{eq3b}
\begin{array}{l}
[accept(john,msc,T) \wedge 3 < T\wedge T\leq 33] \rightarrow add\_pending(john,msc, T_4) \\
\mytab \wedge T_4 = T + 1 
  \wedge send\_invoice(john,msc, T_5) \wedge T_4<T_5  \leq T +30]
\end{array}\end{equation}
If John does not accept his offer by the deadline of time 33 the residue will be discarded. Suppose John accepts at time 16, then the instantiated consequent of the reactive rule will be generated as a goal to be solved.
\begin{equation}\label{eq3c}
\begin{array}{l}
[add\_pending(john,msc, 17) 
\wedge send\_invoice(john,msc, T_5) \wedge 17<T_5  \leq 46]
\end{array}\end{equation}

Facts about fluents are updated destructively, without timestamps, giving rise to an event theory ET as an emergent property that is similar to the event calculus~\cite{KowalskiSergot}. This consists of two templates:
\begin{equation}\label{eq5}
\begin{array}{l}
p(i+1)\leftarrow initiates(e,p) \wedge e\in ev_{i+1} \\
p(i+1) \leftarrow p(i) \wedge \neg \exists e (terminates(e,p) \wedge  e\in ev_{i+1} )
\end{array}\end{equation}
where $ev_{i+1}$ represents the set of events  in the transition between state $S_i$ and $S_{i+1}$.
Note that the second template in ET (Equation~(\ref{eq5})) is a \textit{frame axiom} and whereas in KELPS it is an emergent property,  its translation will need to be included explicitly in the Reactive ASP program.

In the KELPS model-theoretic semantics fluents and events are timestamped and combined into a single model-theoretic structure.
The computational task of KELPS is to make the reactive rules true with respect to the model-theoretic semantics in the presence of dynamically incoming external events, by generating actions that also satisfy  the integrity constraints in $C_{pre}$. 
We now describe the KELPS computational task  more formally.

\paragraph{\bf Notation} If $S_i$ is a set of fluents without timestamps, then $S_i^*$ represents the same set of fluents with timestamp $i$. Similarly, if $ev_i$ is a set of events without timestamps taking place in the transition from state
$S_{i-1}$ to state $S_i$, then events  $ev_i^*$ represents the same set of events with timestamp $i$; likewise the set  of timestamped external events $ext_i^*$ and the set of timestamped agent's own actions $acts_i^*$, where $ext_i$ and $acts_i$ are the external events and agent's actions occurring in the transition between states $S_{i-1}$ and $S_i$.

\begin{definition}\label{def:modelstructure}
Let $<\R,\C,\Aux>$ be a KELPS  framework, $ext^*=\cup_{i} ext_{i}^{*}$  
be a given set of external events   
and  $S_0$ be the initial state. The KELPS computational task is to generate a set  of actions, $acts_{i}$ (and corresponding set of states $S_{i}$), for all $i\geq 1$, satisfying the following properties:
\begin{itemize}
\item $\R \cup \C_{pre}$ is true in the Herbrand model  $\Aux \cup S^* \cup ev^* $, where $ev_i = ext_i \cup acts_i$,   $ev^* = \bigcup_{i\geq1}ev^*_{i} $ and $S^*=   \bigcup_{i\geq 0} S^*_{i} $.
\item State $S_{i+1}$, $ i\geq 0$, is generated from $S_i$, $ev_{i+1}$ and $C_{post}$ and given by $S_{i+1} = $\\
$(S_i - \{p: terminates(e,p) \in C_{post} \wedge e \in ev_{i+1}\})$ $ \cup$$ \{p: initiates(e,p) \in C_{post} \wedge e \in ev_{i+1}\}$.\footnote{Notice that this implies that if some ground fluent $p$ is both initiated and terminated by  actions at $T+1$, then that fluent will hold at $T+1$.} 
\end{itemize}
\end{definition}

Consider the KELPS program described by Equations~(\ref{eq2}) and (\ref{eq4}).
Assume that initially, according to $S_0$, we have  6 copies of `Hamlet' and 2 copies of `Emma' and external events occur at times 1 and 2
as follows:
\begin{tabbing}
abc\=indentmore\=\kill
\>$S_0 = \{available(hamlet, 6), available(emma, 2)\}$\\
\>$ext_1=\{request(john, hamlet), request(john, emma), request(bob, emma)\}$\\
\>$ext_2=\{request(tom, emma)\}$
\end{tabbing}
In order to solve the goals represented by the reactive rules in Equation~(\ref{eq2}) in the light of the external events, the KELPS program will produce a sequence of states and events. 
The OS has some choices {\em --} for example at each cycle it can decide whether or not to execute an action and which actions to execute (concurrently). 
One possible sequence is the following.
\begin{tabbing}
abc\=indentmore\=\kill
\>$acts_1=\{\ \}$,\ \ $S_1 = S_0$, \ \ $acts_2=\{allocate(john, hamlet,6), allocate(john, emma,2)\}$ \\
\>$S_2 = \{available(hamlet, 5), available(emma, 1)\}$\\
\>$acts_3=\{process(john, hamlet), process(john, emma)$, \\
\>\>$allocate(bob, emma,1), order(emma, 20)\}$\\
\>$S_3 = \{available(hamlet, 5), available(emma, 0), on\_order(emma)\}$\\
\>$acts_4=\{process(bob, emma)\}$,\ \ $S_4=S_3$\\
\>$acts_5=\{\ \}$,\ \ $S_5 = S_4$\\ 
\>$acts_6= \{apologise(tom,emma)\}$,\ \ $S_6=S_5$
\end{tabbing}	

Other outcomes and models are also possible and might be generated by the KELPS operational semantics.  One such could issue  an apology to John with respect to his first order instead of processing it, also making the reactive rules true.
We will see later how the ASP translation facilitates specifying preferences between models that could, for example, favour allocating and processing orders wherever possible, rather than issuing apologies.

\subsubsection{KELPS Reactivity}\label{reactivity}
The reactive rules \R~in KELPS are implications and can, in principle, be satisfied (made true) in one of three
ways:  (i) by ensuring their antecedents are false, (ii) by ensuring their consequents are true, (iii) by ensuring their consequents become true whenever their antecedents are true.
We call these possibilities, respectively,  \textit{preemptive}, \textit{proactive} and \textit{reactive}.
For example, consider the second reactive rule in Equation (\ref{eq2}). This rule can be satisfied {\em reactively} as just illustrated above, i.e. by ordering 20 copies of items whenever their number falls below 2 after an allocation. The rule can be satisfied {\em proactively} by ordering 20 copies of all items at all times, 
thus ensuring that the consequent of the rule is always true. The rule can be satisfied {\em preemptively} by ordering at least 2 copies of all items at all times, thus ensuring that the antecedent is never true.
As we will see next, KELPS OS is designed to  generate  only reactive models.
This is also the behaviour we will model in Section~\ref{sec:translation} in our mapping to Reactive ASP. But later in Section~\ref{sec:extensions} we also show how the other two types of behaviour can be achieved in Reactive ASP.

In KELPS  reactive (Herbrand) interpretations are those in which every agent generated action is \textit{supported}, in the sense that it originates from a reactive rule whose $earlier$ parts have already been made true. This is formally defined in Definition~\ref{reactivemodel} and uses the notion of sequencing from Definition~\ref{ordering}.
 
\begin{definition} \label{ordering}
Let $earlier$ and  $later$ be conjunctions of conditions. Then the conjunction $earlier \wedge later$ is  said to respect a {\em sequencing}, denoted $earlier < later$  
if, and only if, there exists a ground substitution $\theta$ for all  time variables in $earlier \wedge later$ such that:
\begin{itemize}
\item all  temporal constraints in $earlier \theta \wedge later \theta$ are true in \Aux, and
\item all timestamps in conditions occurring in $earlier \theta$ are earlier  than ($<$)  all time-stamps occurring in conditions in $later \theta$.
\end{itemize}
\end{definition}
\noindent We can now define a reactive model.

\begin{definition} \label{reactivemodel}
Let $<\R,\C,\Aux>$ be a KELPS framework with initial state $S_0$, and set $ev^*$ of timestamped events partitioned into external events $ext^*$ and actions $acts^*$.
Let $I$ be the (Herbrand) interpretation $I = S^* \cup ev^* \cup \Aux$ and let $\C_{pre}$ be true in $I$.
Then $I$ is a {\em reactive interpretation} if, and only if, for every action $action \in I$, there exists a rule $r \in  \R$~of the form $antecedent \rightarrow [other \vee [earlier \wedge act \wedge rest]]$, and there exists a ground substitution $\theta$ such that $r\theta $ supports $action$, in the sense that all the following hold:
\begin{description}
\item[(i)] $action$ is $act\theta$ 
\item[(ii)] ($antecedent\theta \wedge earlier\theta$) $<$ ($act\theta \wedge rest\theta$) 
\item[(iii)] $I$ satisfies $antecedent\theta \wedge earlier\theta \wedge act\theta$
\end{description}	
$I$ is a \textit{reactive model} of $S^* \cup ev^* \cup \Aux$ if, and only if, $I$ is a reactive interpretation and
\R~is  true in $I$. 	
\end{definition}
\noindent
Note that in Definition~\ref{reactivemodel} condition (iii)  allows $rest\theta$ to be false in interpretation $I$, although it must be possible for $rest$ to become true in the future, in the sense of not violating the temporal constraints. This is ensured by the sequencing in condition (ii).

The KELPS OS allows the agent to perform actions from a consequent more than once, as long as the temporal constraints are respected. This is beneficial, for instance, when an action succeeds, but subsequent ones do not, and the action needs to be repeated.
For example: {\em Suppose a customer requests an item and the seller notifies the customer of a delivery window (W), but the delivery cannot subsequently be performed. Then the seller can do another notification, of a new delivery window, and make another attempt at delivery.}
This is captured by the reactive rule
\begin{equation}\label{eq11}
\begin{array}{l}
request(Cust, Item, T)  \rightarrow \\
\mytab notify\_window(Cust, Item, W, T1) \wedge deliver(Cust, Item, W, T2)\\
\mytab\mytab \wedge  T<T1<T2
\end{array}\end{equation}
A slightly more involved example is the following: if  a fluent is initiated to make the precondition of a later action true, but an external event subsequently terminates that fluent, the initiating action can be re-executed  thus re-establishing the fluent (subject to time constraints). 
Reactive models can also include unnecessary actions  because of this possible repeated execution. Such actions can often be prevented through the use of integrity constraints in $C_{pre}$.

\subsection{Answer Set Programming}\label{ASP}
Answer Set Programming  \cite{Gelfond1988,ASP_Glance,ASP}  is an approach to declarative problem solving. A problem is expressed as a normal logic program with some additional constructs {\em --} the mapping from KELPS presented in this paper makes use of the additional ASP constructs  of choice rules and constraints.
In an answer set program a rule with variables  is viewed as a first order schema and represents the set of ground instances of the rule that are formed by substituting ground terms for the variables (called a grounding). The  ground terms are the constants or functional terms constructed from the signature of the program, and the grounding (even if infinite) must be equivalent to a finite set of ground rules. In what follows, without loss of generality, we consider a program to be a set of ground rules.
A normal rule, $r$, over a set of atoms $A$ is of the form $h:- b_1,...,b_m, not  \: b_{m+1},...,not \: b_n$,
where $0 \leq m \leq n$,  $h$ and each $b_i$, $0 \leq i \leq n$, are   atoms in $A$, and for any atom $a$,  $not \: a$ is default negation.
A \textit{literal} is  $a$ or $not \: a$, where $a$ is an atom in $A$. 
For a normal rule $r$:
$head(r) = \{h\}$;
$body(r) = \{b_1,\ldots,b_m, not  \: b_{m+1},\ldots,not \: b_n \}$;
$body(r)^{+} = \{b_{1},\ldots,b_{m}\}$;
$body(r)^{-} = \{ b_{m+1},\ldots, b_{n} \}$.
$atoms(r) = head(r) \cup body(r)^{+} \cup body(r)^{-} $.
The informal meaning of a  normal rule is that if the body is true, then so is the head. 

A \textit{choice rule}  is of the form
		$l\{h_1,...,h_k\}u:- b_1,...,b_m,not \: b_{m+1},...,not \:b_n$,
where $0 \leq m \leq n$, each $h_i$ and $b_i$ is an atom in $A$, and $l$ and $u$ are non-negative integers satisfying $l \leq u$. For a choice rule $r$, $head(r)=\{h_1,...,h_k\}$, 
$body(r) = \{b_1,...,b_m, not  \: b_{m+1},...,not \: b_n \}$, and $r$ is true if between \textit{l} and \textit{u} atoms from the set $\{h_1,...,h_k\}$ in the head are true when the body  is true. 
An \textit{integrity constraint} $r$ is of the form
$:-b_1,...,b_m, not  \: b_{m+1},...,not \: b_n$,
where $1 \leq m \leq n$ and each  $b_i$ is an atom in $A$.
For a constraint $r$,   $head(r)$ is empty (implicitly false) and $body(r) = \{b_1,...,b_m, not  \: b_{m+1},...,not \: b_n \}$. (Note that for a  choice rule or constraint $r$, $body(r)^{+}$, $body(r)^{-}$ and $atoms(r)$ are defined as for a normal rule $r$.)
An integrity constraint $r$ is {\em satisfied} if $body(r)$ is false.
By insisting that integrity constraints are satisfied in an answer set, they effectively rule out answer sets for which this is not the case. 
Furthermore, in the original (non-ground) program, every normal rule, choice rule or constraint $r$ must be \textit{safe}; that is every variable which occurs in $r$ must occur at least once in $body(r)^{+}$.

ASP also allows  weak constraints, which are used to define a preference ordering over the answer sets of a program. Usually, one looks for the  best answer sets in the ordering.
A \textit{weak constraint} is  of the form
$:\sim l_1, \ldots, l_{m}$.$[penalty@level, t_1, \ldots, t_{n}]$,
where each $t_i$ is a term occurring in the set   of body literals, $l_i$, $level$ is a positive integer representing a priority and $penalty$ is an integer.
In our programs we  assume that the   terms $t_{i}$  are exactly the variables occurring in the body literals. 
With this (simplified) assumption, for each answer set, and for each level, the penalty for each weak constraint instance whose body is satisfied by the answer set  is accumulated into a total $W$ for that answer set. This total is the penalty the answer set pays for making the body of the constraint instances true at the given level.
The returned answer sets are those with minimal overall accumulated penalty value, where penalties at the highest level are considered first, and  lower level penalties are only considered if all higher level penalties for two answer sets are equal. If maximal accumulated weight values are required, then the penalties would be given as negative integers.
An example is provided by the trading program introduced in Equation~(\ref{eq2}) {\em --} it is preferable to allocate an item if one is available than to issue an apology. This can be captured by  the following weak constraint:
\begin{equation}\label{eq21}
\begin{array}{l}
:\sim apologise(Cust,Item,T)\mbox{.}[1@1,Cust,Item,T]
\end{array}\end{equation}
\noindent which states that a penalty of 1 is paid every time an apology is made.\footnote{Later we use a reified notation for event occurrences.} 
In this case the optimal answer set would contain an instance of the atom $allocate(Cust,Item,N,T)$ (which has no penalty), where possible in ensuring a reactive rule  is satisfied, rather than a corresponding instance of $apologise(Cust,Item,T)$.

The semantics of an ASP program $P$ is given by its \textit{answer sets}, which are defined in terms of the reduct of  $P$. We use the definition of reduct from \cite{MarkLawReduct}, repeated below, which is adequate for our purposes.

\begin{definition}\label{MarkReduct}
The reduct of a   program $P$ with respect to an interpretation $I$, denoted $P^{[I]}$, is constructed through the following 4 steps.
\begin{enumerate}
\item Remove any normal rule, choice rule, or constraint whose body contains 
$not \: a$ for some 
$a \in I$ and remove any negative literals from the remaining rules or constraints.
\item For any constraint $r$ replace $r$ with $\bot:\mbox{-} body(r)^{+}$ ($\bot$ is a new atom which cannot appear in any answer set of $P$).
\item For any choice rule $r$ , 
$l\{h_1,...,h_{k}\}u:\mbox{-} body(r)^{+}$,  such that $l \leq  \mid I \cap \{h_1, \ldots, h_{k}\}\mid \leq u$, replace $r$ with the set of rules  
$\{h_{i} :\mbox{-} body(r)^{+}  \mid$ 
$h_{i} \in I \cap \{h_1, \ldots, h_{k}\}\}$.
\item For any remaining choice rule $r$,  replace $r$ with the constraint $\bot:\mbox{-} body(r)^{+}$.
\end{enumerate}
The {\em Answer Sets} of a program $P$ are those interpretations  $I$ that are minimal models of 
the reduct of $P$ with respect to $I$.
\end{definition}
In Section~\ref{sec:theory}, where we give some formal results, we will make use   of the notions of splitting set and partial evaluation, which we recall next (taken from \cite{Splitting}).
\begin{definition}\label{splittingSet}
Let $P$ be a ground normal  program and $U$ be a set of ground atoms. $U$ is called a {\em splitting set of $P$} if for every rule $r \in P$, if $head(r) \cap U\neq \emptyset$ then $atoms(r) \subseteq U$. The set of rules $r \in P$ such that $atoms(r) \subseteq U$ is called the {\em bottom of $P$ w.r.t. $U$}, denoted $bot_{U}(P$), and the set $top_{U}(P) = P-bot_{U}(P)$ is called the {\em top of $P$ w.r.t. $U$}.
\end{definition}

\begin{definition}\label{partialEval}
Let $U$ and $X$ be sets of atoms and $P$ be a ground  normal  program. Then the set of rules $ev_{U}(P,X)$ is obtained from $P$ by {\em partial evaluation} as follows.  
For each rule $r \in P$ such that  $body^{+}(r) \cap U\subseteq X$ and $body^{-}(r) \cap U$ is disjoint from $X$, then the rule $r'$, where $head(r')=head(r)$, $body^{+}(r') = body^{+}(r)-U$, and $body^{-}(r') = body^{-}(r)-U$ belongs to $ ev_{U}(P,X)$.
\end{definition}

The Splitting Set Theorem \cite{Splitting} allows for the answer set of a locally stratified program to be constructed iteratively. The theorem states that $I$ is the answer set of $P$, split by $U$, iff it can be written as the union of $X$ and $Y$, where $X$ is the answer set of $bot_{U}(P)$ and $Y$ is the answer set of $ev_{U}(top_{U}(P),X)$.

In Section~\ref{sec:theory} the Splitting Set Theorem will only be applied to  reducts of 
programs with no constraints. In particular, these will be locally stratified programs with  no negation, and which therefore will have a unique answer set \cite{GelfondChapter}. The set construction  is  relatively simple and is described in Lemma~\ref{answerSetconstruction} (adapted from \cite{Graham} Lemma 5.17) in order to introduce some notation.
The idea is to compute the answer set in steps. Beginning with the  lowest strata 0, the atoms in strata 0 are used to split the program then the answer set of the bottom part is used to partially evaluate the remaining strata. The process is repeated using atoms in strata $\{0,1\}$ as a splitting set, continuing in a similar way through all strata until the final strata  is reached. The answer set is then the union of the answer sets computed at each step.

\begin{lemma}\label{answerSetconstruction}
Let $P$ be a ground locally stratified positive program with $n+1$ strata labelled $0\ldots n$, where $R_j$ is the set of rules in strata $j$ and ${\cal H}_j$ is the set of atoms occurring in strata 0 to $j$, then its unique answer set $A$ can be constructed iteratively using the Splitting Set Theorem.
\begin{proof}
Let $\pi_0=P$.
Then ${\cal H}_0$, the set of atoms in the rules in $R_0$, splits    $\pi_0$  into $bot_{{\cal H}_0}(\pi_0)$ and $top_{{\cal H}_0}(\pi_0)$. 
Define $\pi_1 = ev_{{\cal H}_0}(top_{{\cal H}_0}(\pi_0),S_0)$, where $S_0$ is the answer set of $R_0$ (=   $bot_{{\cal H}_0}(\pi_0)$).
The lowest strata of $\pi_1$ is strata 1 of $\pi_0$, but with its rules partially evaluated by $S_0$. By the stratification these rules are not dependent on atoms in higher strata.
More generally, given $S_0$ is an answer set of $bot_{{\cal H}_0}(\pi_0)$, for $1 \leq j \leq n$ we can split $\pi_{j-1}$ using ${\cal H}_{j-1}$ into $bot_{{\cal H}_{j-1}}(\pi_{j-1})$ and $top_{{\cal H}_{j-1}}(\pi_{j-1})$, and define $\pi_j=ev_{{\cal H}_{j-1}}(top_{{\cal H}_{j-1}}(\pi_{j-1}),S_{j-1})$, where $S_{j-1}$ is the answer set of $bot_{{\cal H}_{j-1}}(\pi_{j-1})$.

Then by the Splitting Set Theorem $S=\cup_{j=0}^{n} S_j$ is the answer set of $\pi_0$.
\end{proof}
\end{lemma}

\paragraph{Controlling Grounding and Solving in \textit{clingo} 4.}
 In this work, we have used the \textit{clingo}  implementation, version 4 \cite{clingo_guide}.\footnote{The \textit{clingo} 4 webpage (https://potassco.org/clingo/) states that \textit{clingo} 4 adheres to the ASP language standard \cite{ASPCore2}. Since our use of ASP is compatible with the structures in ASP-Core-2 our programs also adhere to the standard.} 
 The standard mode of computation in ASP is the following (called {\em single shot}):
first it generates a finite propositional representation of the program (called a {\em grounding}), and then it computes the answer sets of the resulting propositional program. 
 We assume this  computational mode in the following sections.
However, it is not the only way KELPS could be simulated. 
There is an alternative   incremental computation (called {\em multi-shot})  provided by \textit{clingo} 4, which can also be used. 
In this mode of operation, the program is structured into parametrisable subprograms. The grounding and assembly of these subprograms are modular and controllable using one of the embedded scripting languages.\footnote{Either Python or Lua may be used; in our implementation we used  Lua.}
Control of which rules to include in the subprogram assembly is achieved through the use of an \textit{external} atom
that can be set to true or false before each iteration of the multi-shot solving process. 
In Section~\ref{sec:inc} we explain why we kept to the standard mode in our mapping. Nevertheless, for completeness we outline the incremental mapping method in the Appendix.

%% file: Translation_Final_corr.tex
\section{Mapping KELPS to ASP}\label{sec:translation}

In this section we  show how a  KELPS  program  $P$ can be systematically mapped  into an ASP program $PA$ such that  answer sets of $PA$ correspond to reactive models of $P$.  In this way a notion of reactivity similar to that in KELPS is injected into ASP. We call programs written in this way {\em Reactive ASP}. In order to obtain the correspondence
we  define the new notion of an $n$-distant KELPS framework.

In this section it is convenient to associate a KELPS framework with a set of external events $E$ and we write a framework as $<\R,\C,\Aux>_{E}$.

\begin{definition}\label{ndistantframework}

A KELPS framework $<\R,\C,\Aux>_{E}$ is called $n$-distant, for some $n\geq 0$, if it satisfies the following two properties:
\begin{description}
\item[i)]
In every reactive rule $r$ in $\R$, for every timestamp parameter $Time$ that  is  
a universally quantified time variable there is a condition $Time \leq n$ in the antecedent of $r$, 
and for every timestamp parameter $Time$ that is an existentially quantified time variable  in a disjunct in the consequent  of $r$ there is a condition $Time \leq n$ in that disjunct.
\item[ii)] Any constant timestamp $c$ in  a reactive rule 
must satisfy $c \leq n$.
\item[iii)] There are no external events in $E$ after time $n$; i.e. $ext^*_{i} = \{\ \}$ for all $i > n$.
\end{description}
\noindent
We denote an $n$-distant  KELPS framework by $<\R,\C,\Aux,n>_{E}$. 
A reactive model of a KELPS $n$-distant framework is called an $n$-distant KELPS reactive model (or $n$-distant reactive model for short).
\end{definition}

\noindent We assume in what follows that an $n$-distant KELPS framework can only consider external events that occur at times up to and including $n$ and where it is clear we drop the subscript $E$ in the notation $<\R,\C,\Aux,n>_{E}$.

We can make several observations regarding $n$-distant KELPS frameworks:
\begin{itemize}
\item As a consequence of the properties in Definition~\ref{ndistantframework} $acts^*_{i} = \{\ \}$ for all $i>n$ and for all $i>n$ the non-timestamped set of states $S_{i} = S_{n}$. 
That is, there are no state changes after time $n$.
\item A KELPS framework  that satisfies conditions (ii) and (iii) can be  transformed into an $n$-distant KELPS framework   $F_{n}$, for $n\geq 0$, called its $n$-distant conversion,   by adding the temporal constraints for the time parameters to each of its reactive rules.
For example, let a KELPS framework contain the reactive rule $true \rightarrow a1(T1) \wedge a2(T2) \wedge T2=T1+1$, where $a1$ and $a2$ are events. Then the 2-distant conversion will replace that rule with $true \rightarrow a1(T1) \wedge a2(T2)  \wedge T2=T1+1 \wedge T1 \leq 2 \wedge T2 \leq 2$.
\item Let  $F$ be a KELPS framework and $F_{n}$ be its $n$-distant conversion. Even 
with the same set of external events, not every reactive model of $F$ is a 
reactive model of $F_{n}$. Consider  for example the above reactive rule:
$true \rightarrow a1(T1) \wedge a2(T2) \wedge T2=T1+1$, and suppose $n=2$.
The only  reactive model of $F_{n}$ is $\{a1(1), a2(2)\}$, whereas other  reactive models are possible for $F$; for example $\{a1(3), a2(4)\}$, or more interestingly  $\{a1(1), a2(2), a1(2)\}$.
The latter is possible in $F$ because $a1(2)$ is a supported action  in $F$ (Definition~\ref{reactivemodel}), but it is not supported in $F_{2}$ because the 2-distant framework will not allow the possibility of $a2$ to be executed at time 3, or more specifically the condition $3 \leq 2$ will not be satisfiable.
Note that this clearly  shows that for example $F_{2}$ and $F_{4}$, with the same set of external events, will not necessarily have the same models. This holds in general for any  $F_{n}$ and $F_{m}$, for $m\neq n$.
Moreover, if one has a model the other is not guaranteed to have one too.
\item 
The converse property, that models of $F_{n}$ are models of $F$, also does not hold in general  even  if the external events remain the same. As an example, suppose $F$ consists of the reactive rule  
$a(T1) \wedge p(T) \wedge T=T1+1 \rightarrow a1(T2) \wedge T2=T+1$, where $a$ and $a1$ are events and $p$ is a fluent. 
The corresponding $F_{n}$ would be $a(T1) \wedge p(T) \wedge T=T1+1  \wedge T1\leq n \wedge T\leq n \rightarrow a1(T2)\wedge T2=T+1 \wedge T2\leq n$. Now  suppose also that $n=3$, $a(3)$ is an event that has occurred, $p$ holds initially and no event terminates $p$. The antecedent will be false because $T1+1=4$ and $4\not\leq n$. Thus $F_{n}$ has a reactive  model with only one event $a(3)$, but this is not a model of $F$, as in $F $
the antecedent is true and the consequent is false.
\end{itemize}

However, as the following Lemma~\ref{FnSubsetF} shows, if rules in $F$ are restricted such that the timestamp of every fluent in the antecedent (if any)  is guaranteed to be $\leq$ the timestamp of some event in the antecedent, then models of $F_{n}$ are indeed models of $F$  (again assuming that the external events remain the same). Notice that the above counter-example does not conform to this restriction.\footnote{This restriction captures many cases of reactive rules when the reaction is primarily to an event, that may have happened under certain circumstances (e.g. fluents holding before its occurrence). Violating such a restriction can lead to issues of refraction \cite{daSilva}. For example, the reactive rule: $enter\mbox{-}room(T) \wedge hot(T1) \wedge T1>T \rightarrow open\mbox{-}window(T2) \wedge T2>T1$ violates this restriction and may lead to multiple attempts to open the window if the room remains hot after entering it.}

\begin{lemma}\label{FnSubsetF}
Let  $F$ be a KELPS framework such that  each rule conforms to the restriction that the timestamp of every fluent in the antecedent (if any)  is guaranteed to be $\leq$ the timestamp of some event in the antecedent.
Then for any $n$, if 
 $F_{n}$ is the  $n$-distant conversion of $F$, 
 and   $F$ and $F_{n}$ have the same external events, then 
any $n$-distant reactive model of $F_{n}$ is also a reactive model of $F$.
\begin{proof}
Let $I_{n}$ be a reactive model of $F_{n}$, and suppose for contradiction that $I_{n}$ is not a reactive model of $F$.
There are two cases:  $I_{n}$ must either 
make a precondition false or make a reactive rule false.\\ \\
\noindent {\em Case 1:} If $I_{n}$ makes a precondition $c$ in $F$ false, then for some instance of $c$ it makes the constraint body true. By assumption, any external event true in $I_{n}$ is  timestamped $\leq n$, and by construction of the rules in $F_{n}$ any generated action will also be timestamped $\leq n$. Since precondition constraints include an event, the false instance of $c$ must be at a time $T \leq n$, contradicting the fact that $c$ is true   in $I_{n}$.   \\ \\
\noindent {\em Case 2:} If $I_{n}$ makes a rule $R$ in $F$ false, then for some instance $r$ of $R$ it makes the antecedent  true and the consequent false. 
First, note that all events in $I_{n}$ will have timestamps $\leq n$. Now consider the antecedent of $r$. If it includes no fluents then if $I_{n}$ makes the antecedent true it would also make the antecedent of the rule true in the $n$-distant conversion.
On the other hand, if the antecedent includes a fluent, then by assumption the fluent must be timestamped with a $Time$ that is $\leq$ the timestamp of some event also in the antecedent. In this case, if the antecedent is true in $F$ it would also have been true  in the $n$-distant version $F_{n}$ (again by a similar argument to that in Case 1) and hence the consequent would be true in $F_{n}$ and also in $F$, contradicting the assumption.
\end{proof}
\end{lemma}

\subsection{Basics of Mapping KELPS to ASP}\label{basicMapping}
We next present the basics of the Reactive ASP mapping through a simple example, elaborating  where necessary.

The  KELPS framework in Figure~\ref{simpleKelps} captures a simple narrative for evacuation if an alarm sounds. {\em Initially  at  time 0  the door is locked. Then at time 2 an alarm sounds. Evacuation is impossible while the door remains locked. The door is unlocked at time 4.}

\begin{figure}[hbpt]
\begin{tabular}{ll}\hline
{\textit{R}:}& $alarm(T) \rightarrow evacuate(T_{1}) \wedge T < T_{1}$\\ 
{\textit{C$_{pre}$}:}&$false \leftarrow evacuate(T+1) \wedge door\_locked(T)$\\ 
{\textit{C$_{post}$}:}&$terminates(unlock, door\_locked)$\\ 
$S_0$ = & $ \{door\_locked\}$ and  $ext^{*}=\{alarm(2), unlock(4)\}$
\\ \hline\end{tabular}
\caption{Simple KELPS framework.}
\label{simpleKelps}
\end{figure}
\noindent Consider the $n$-distant version of the above framework with $n = 7$. Definition~\ref{ndistantframework} means that for $n$ = 7 this KELPS framework produces 
sequences of states and events incrementally for each time point up to 7 taking into account the external events\footnote{Remember that state $S_i$ results after occurrence of events $ev_i$.}, of which one such sequence  is:
	$S_0 = \{door\_locked\} = S_1 = S_2 = S_3, S_4 =  S_5=S_6=S_7=\{\}$, 
	$ev_1 = ev_3 =  ev_6 = ev_7 =\{\}$,   $ev_2 = \{alarm\}$, $ev_4 = \{unlock\}$,  $ev_5 = \{evacuate\}$,
and furthermore $\R \cup \C_{pre}$ is true in the Herbrand model  $S^* \cup ev^* \cup \Aux$.
Other 7-distant models are also possible including an $evacuate$ action at time 6 or time 7, and possibly more than one $evacuate$ action.  

The mapping of the above KELPS program  into Reactive ASP is shown in Figure~\ref{singlesimplecorrected}.\footnote{From here onwards, we use Reactive ASP and ASP interchangeably.}
The ASP program uses a number of special predicates with particular meanings relevant to the KELPS program. These are defined in Table~\ref{signature} and further explained below.
\begin{figure}[hbtp]
\begin{center}{\small
\begin{tabular}{ll}
\hline
1. &\texttt{time(0..7).} \ \ \ \textit{\% Time range}\\
2. & \texttt{holds(door\_locked,0).}\ \ \  \textit{\% Initial state S0}\\
3. & \texttt{happens(alarm,2).} \ \ \  \textit{\% External events}\\
4. & \texttt{happens(unlock,4).}\\
& \textit{\% defines `antecedent' and `consequent'}\\
5. &\texttt{ant(1,(Ts),Ts):-happens(alarm,Ts),time(Ts).}\\
6. &\texttt{cons(1,(T),T,Ts):-ant(1,(T),T),happens(evacuate,Ts),T<Ts,time(Ts).}\\
& \textit{\% Constraint enforcing the reactive rule(s)}\\
7. &\texttt{:-ant(ID,X,Ts),not consTrue(ID,X,Ts),time(Ts).}\\
8. &\texttt{consTrue(ID,X,Ts):-cons(ID,X,Ts,Ts1),time(Ts1).}\\
& \textit{\% Supported actions}\\9. &\texttt{0\{happens(Act,Ts)\}1:-supported(Act,Ts),time(Ts),Ts>0.}\\
10. &\texttt{supported(evacuate,Ts):-ant(1,(T),T),T<Ts,time(Ts).}\\
11. &\texttt{terminates(unlock,door\_locked).}\ \ \ \textit{\% \mbox{$C_{post}$}}\\
12. &\texttt{:-happens(evacuate,Ts),holds(door\_locked,Ts-1),time(Ts-1),time(Ts).}\ \textit{\% \mbox{$C_{pre}$}}\\
& \textit{\% Event theory}\\
13. &\texttt{holds(P,Ts):-initiates(E,P),happens(E,Ts),time(Ts).}\\	
14. &\texttt{holds(P,Ts):-holds(P,Ts-1),not broken(P,Ts),time(Ts-1),time(Ts).}\\
15. &\texttt{broken(P,Ts):-terminates(E,P),happens(E,Ts),time(Ts).}\\		
\hline\end{tabular}
}
\caption{Reactive ASP mapping of example in Figure~\ref{simpleKelps}.}
\label{singlesimplecorrected}
\end{center}
\end{figure}
\begin{table}[hbtp]
{\small\begin{tabular}{l|l}\hline
Predicate & Meaning\\ \hline
\texttt{time(X)} & $X$ is a valid timestamp ($0\leq X \leq n$ for $n$-distant KELPS)\\
\texttt{happens(X,Ts)} & Event $X$ occurs in the timestamp interval [$Ts-1,  Ts$)\\
\texttt{holds(X,Ts)} & Fluent $X$ holds at timestamp $Ts$ \\
\texttt{ant(ID,Args,Ts)} & \parbox[t]{3.8in}{Antecedent of rule with identifier $ID$ and arguments $Args$ becomes true at timestamp $Ts$}\\
\texttt{cons(ID,Args,T,Ts)} &  \parbox[t]{3.8in}{(A disjunct of the) Consequent of rule with identifier $ID$  whose antecedent with arguments $Args$ became true at $T$ becomes true at timestamp $Ts$}\\
\texttt{broken(X,Ts)} &  \parbox[t]{3.8in}{Fluent $X$ is terminated by some event that happened in the half-open timestamp interval [$Ts-1,Ts$)}\\
\texttt{supported(X,Ts)} & Action $X$ is supported at timestamp $Ts$\\
\texttt{consTrue(ID,Args,Ts)} &  \parbox[t]{3.8in}{A supplementary predicate expressing that the consequent of rule with identifier $ID$ whose antecedent with arguments $Args$ became true at $Ts$ has become true at some time $Ts1$ (necessarily $Ts1>Ts$)}\\
\hline
\end{tabular}}
\caption{Predicates used in mapping KELPS to Reactive ASP.}
\label{signature}
\end{table}

We capture the $n$-distant notion of KELPS by adding an assertion \texttt{time(0..n)} to the ASP program, here represented  by \texttt{time(0 ..7)} (in Line 1),  an ASP shorthand for  the facts \texttt{time(0)},\ldots,\texttt{time(7)}.  Auxiliary atoms are also mapped to themselves.
In what follows we have often included time atoms  in the body of rules resulting from   the translation to ASP for two reasons.
Firstly it makes for easy comparison with the $n$-distant transformation, and secondly their inclusion can reduce the size of the ASP program grounding, for instance in $C_{pre}$ (Line 12). Unless they are required to guarantee safeness of a rule we have  otherwise minimised their use.

Observe  that in our ASP translation we reify fluent atoms using the meta-predicate \textit{holds} and reify events using the meta-predicate \textit{happens}.
This has several advantages. It allows to capture the event theory that has to be made explicit in Reactive ASP more succinctly. It also allows to define general choice rules and the notion of supportedness which also has to be made explicit in ASP.

In Figure~\ref{singlesimplecorrected} the initial state is captured by \texttt{holds} facts with timestamp 0 (in Line 2), while external events $ext^*=\{alarm(2), unlock(4)\}$ are modelled using the $happens$ meta-predicate (in  Lines 3 and 4). 

\paragraph{Causal Theory.}
In translating the KELPS framework into ASP we have kept as close to the KELPS syntax as possible.
For instance,
the  postcondition part of the causal theory, $C_{post}$,  uses \texttt{initiates} and \texttt{terminates} facts, exactly as in KELPS. 
However, in ASP, in case actions or fluents have arguments these need to be qualified. For instance, the KELPS $C_{post}$ fact
\texttt{initiates(develop\_symptoms(P),ill(P))} would require in ASP the atom \texttt{person(P)} to be added into the body, turning the fact into a clause.

The precondition part of the causal theory, $C_{pre}$, uses constraints.
In Figure~\ref{singlesimplecorrected}, Line 12 states that the evacuate action cannot occur  in the interval $[Ts-1,Ts)$ if the door is locked at time $Ts-1$.
The event theory ET (Equation~(\ref{eq5})), that was an 
emergent property of the KELPS OS,  is included explicitly in the ASP ontology to allow reasoning about fluents that are true in each cycle (Lines 13 to 15 in Figure~\ref{singlesimplecorrected}).  The  predicate \texttt{broken} is introduced to avoid a negated conjunctive condition in Line 14. 
A consequence of an explicit event theory in Reactive ASP programs is that reasoning with frame axioms (via Lines 14 and 15) is needed. This is something that KELPS was designed to avoid for the sake of efficiency. In generating answer sets the ASP program will have to duplicate fluents from state to state with increasing timestamps until 
the fluents are 
terminated by events.

\subsection{Mapping the Reactive Rules}
Before explaining the way we map a reactive rule (see  Lines 5 to 8 in Figure~\ref{singlesimplecorrected}), we first  express the general case of a KELPS reactive rule, rewriting (1), to differentiate between time variables and non-time variables, as follows:\footnote{We replace any time constant $k$ in a reactive rule with a new variable $T$ and add $T=k$ to the conjunct in which $k$ appears.}
\begin{equation}\label{generalrule}\forall \overline{X} \forall \overline{T} [antecedent(\overline{X} \cup \overline{T})\rightarrow \exists \overline{Y} \exists \overline{T_1} consequent(\overline{X'} \cup \overline{T'}, \overline{Y} \cup \overline{T_1})]
\end{equation}
\noindent where $\overline {X}$ ($\overline{T}$) represents all the non-time (time) variables that occur in $antecedent$, and $\overline{Y}$ ($\overline {T_1}$) is the set of all non-time (time) variables that occur only in $consequent$. 
$\overline {X'}$ and $\overline {T'}$ represent subsets of $\overline {X}$ and $\overline {T}$, respectively.

In the mapping to ASP a reactive rule is given an identifier $ID$ and  is represented by several rules and an integrity constraint.  One of these rules captures the antecedent (with head using the \texttt{ant} predicate) and the others capture the consequent (with head using the \texttt{cons} predicate).
The body of the \texttt{ant} rule maps the antecedent of the reactive rule identified by $ID$, while the bodies of the \texttt{cons} rules map the disjuncts in the consequent of the reactive rule identified by $ID$.
Assume that $R_{ID}$ is a KELPS reactive rule of the form given by (\ref{generalrule}). Then the $antecedent$ and each disjunct of the $consequent$ are mapped to rules with the following structures:
\begin{equation}\label{eq16}
{\small \begin{array}{l}
		\texttt{ant(ID,($\overline{X'} \cup \overline{T'}$),\texttt{Ts}):-$antecedent(\overline{X} \cup \overline{T}$),$max(\overline{T}$,\texttt{Ts}$)$,\texttt{time(Ts)}}.\\
		\texttt{cons(ID,($\overline{X'} \cup \overline{T'}$),\texttt{Time},\texttt{Ts}):-ant(ID,($\overline{X'} \cup \overline{T'}$),\texttt{Time}),} \\
		\mytab\mytab consequent_{i}(\overline{X'} \cup \overline{T'}, \overline{Y} \cup \overline{T_1}), max(\overline{T'} \cup \overline{T_1}, \texttt{Ts}), \texttt{time(Ts)}. 
		\end{array}}
\end{equation}

\noindent The body conditions $antecedent(\overline{X} \cup \overline{T})$ and $consequent_{i}(\overline{X'} \cup \overline{T'}, \overline{Y} \cup \overline{T_1})$ represent the conjunction of event, fluent, and auxiliary literals
in the respective KELPS antecedent and each disjunct $consequent_{i}$ of $consequent$.
The condition \texttt{ant} in the definition of \texttt{cons} ensures that the variables in the head of the rule are safe. 
More particularly, the rule ID, $\overline{X'} \cup \overline{T'}$ and the \texttt{ant} timestamp \texttt{Ts}, combined, enables to identify each unique instance of an $antecedent$ having been satisfied and to  identify a corresponding instance of \texttt{cons}.
The variables in $\overline{X}-\overline{X'}$, and $\overline{T}-\overline{T'}$ occur only in $antecedent$ and are not required for this identification. Moreover, avoiding their inclusion   simplifies the grounding of the ASP program.
The timestamp \texttt{Ts} represents the time at which the head atom of either rule becomes  true; note that the \texttt{ant} timestamp is represented by the \texttt{Time} variable in the \texttt{cons} rule.
So for \texttt{ant}, \texttt{Ts} is the maximum of all antecedent time variables in $\overline{T}$,  and for \texttt{cons}, \texttt{Ts} is the maximum of all  time parameters occurring in $consequent_{i}$, i.e. the maximum of the times in \texttt{$\overline{T'}\cup \overline{T_1}$}. 
Note that the combination of $max$ and  $time$ conditions achieves the correspondence to $n$-distant KELPS ensuring that all time parameters are constrained to be $\leq n$.

In practice, to implement the \textit{max} atom in (\ref{eq16}), one or more linked atoms using the  auxiliary predicate \texttt{max/3}, which holds if the third argument is the greater of the first two arguments, are used. Moreover, the $max$ function  is needed only when the time variables in antecedent or consequent are not totally ordered. This is why it is not needed in Figure~\ref{singlesimplecorrected}.
For an example when $max$ is necessary, consider the following reactive rule that states  if  $event$ occurs, then the agent must perform $action_1$ and $action_2$ (in any order and possibly concurrently):  $event(T) \rightarrow action_1(T_1) \wedge action_2(T_2) \wedge T < T_1 \wedge T < T_2$.
In ASP, assuming we give the rule an $ID=1$, the consequent part is represented as:
\[{\small\begin{array}{l}
\texttt{cons(1,(T),T,Ts):-ant(1,(T),T),happens(action1,T1),happens(action2,T2), }\\
\mytab\texttt{T<T1,T<T2,max(T1,T2,Ts),time(Ts).}
\end{array}}\]

For disjunctive consequents, we define \texttt{cons} separately for each disjunct; i.e. there is a \texttt{cons}/4 rule for each course of action the agent could take. For example
the disjunctive reactive rule in (\ref{eq2})
is mapped as:
\[{\small\begin{array}{l}
\texttt{ant(1,(Cust,Item,Ts),Ts):-happens(request(Cust,Item),Ts),time(Ts). }\\
\texttt{cons(1,(Cust,Item,T),T,Ts):-ant(1,(Cust,Item,T),T),}\\
\mytab\mytab\texttt{holds(available(Item,N),T1),happens(allocate(Cust,Item,N),T2),}\\
\mytab\mytab\texttt{T2=T1+1,happens(process(Cust,Item),Ts),T<T2,T2<Ts,Ts<T+4,time(Ts).}\\
\texttt{cons(1,(Cust,Item,T),T,Ts):-ant(1,(Cust,Item,T),T),}\\
\mytab\mytab\texttt{happens(apologise(Cust,Item),Ts),Ts=T+4,time(Ts).}
\end{array}}\]

The generic constraint Equation~(\ref{eq10}) enforces all reactive rules:
\begin{equation}\label{eq10}
{\small\begin{array}{l}
   \texttt{:-ant(ID,Args,Ts),not consTrue(ID,Args,Ts),time(Ts).}\\
   \texttt{consTrue(ID,Args,Ts):-cons(ID,Args,Ts,Ts1),time(Ts1).}
\end{array}}\end{equation}
\noindent The variable \texttt{Ts} represents the time at which the antecedent becomes true, and the last argument of \texttt{cons} represents an existentially quantified timestamp \texttt{Ts1} when the consequent becomes true (for example \texttt{Ts} and \texttt{Ts1} correspond, respectively, to $T$ and $T1$  in the KELPS reactive rule of Figure~\ref{simpleKelps}). 
The constraint ensures all answer sets possess the property that  there is at least one instance of {\small\texttt{cons(ID,Args,Ts,Ts1)}} for every instance of {\small \texttt{ant(ID,Args,Ts)}}.

Note that we do not try to map a reactive rule in \R~directly as an ASP normal rule for several reasons: 
\begin{itemize} 
\item  any ASP rule of the form $consequent(\overline{X},\overline{Y}) \leftarrow antecedent(\overline{X})$ would mis-interpret the quantification of $\overline{Y}$ as universal, whereas it is existential in \R;\footnote{It would also be `unsafe', because the variables in $\overline{Y}$ do not appear in any positive body literals.} 
\item   the consequent is, in general, disjunctive; 
\item   the consequent may contain fluents. The reactive rules in KELPS are goals to be satisfied {\em--} they are not used  to directly allow inference of
fluents. There is a structure to KELPS programs whereby the truth of  fluents is affected only through events;
 and 
 \item  we would like to capture the reactivity of KELPS (as given in Definition~\ref{reactivemodel}).
The representation of KELPS reactive rules by the separation into \texttt{ant} and \texttt{cons} and a generic constraint makes this possible.
\end{itemize}

\subsection{Mapping Supportedness}\label{supportedMapping}
Recall that the KELPS OS produces \textit{reactive} models, in which the agent responds to triggers but does not behave proactively or preemptively. Reactivity is an emergent property of the KELPS OS, but in ASP it has to be stipulated explicitly in the program. In the case of the  example in Figure~\ref{simpleKelps} we determine from \R~that the agent should perform the action $evacuate$ at some time $T1>T$ if  the rule antecedent ($alarm$) has occurred by time $T$. 
As seen in Line 10 of the program in Figure~\ref{singlesimplecorrected} we express this using a meta-predicate \texttt{supported}/2. 
This stipulates that action $evacuate$ is {\em supported} any time (within the $n$-distance) after the antecedent of the reactive rule with ID=1 becomes true.
To achieve in ASP the effect of the KELPS abductive generation of actions to make the reactive rules true, we use a choice rule as seen in Line 9, which
specifies that any action that is supported at time \texttt{Ts}  may  \texttt{happen} at time \texttt{Ts}, or not,
and  ensures that only  supported actions can be added to the answer sets.\footnote{Without the explicit condition  \texttt{supported}  ASP would generate arbitrary actions with no relationship to the reactive rule.}

More generally, according to Definition~\ref{reactivemodel},
an action  $act$ can only be performed if 
there exists  (an instance of) a reactive rule in the form $antecedent \rightarrow [other \vee [earlier \wedge act \wedge rest]]$, where $antecedent$ and $earlier$ are already true, and there is enough time for $rest$ to become true in the future. 
The `future' in an $n$-distant reactive model is capped by the value of $n$.
To model this we define \texttt{supported}/2 for every $act$ in a  reactive rule of the form $antecedent \rightarrow [other \vee [earlier \wedge act \wedge rest]]$. 
The head atom contains the $act$, and the body contains the conjuncts of $antecedent$ and $earlier$. For the $rest$, we check there is a future time when $rest$ can be satisfied
without violating the temporal constraints. 

The general schema of a \texttt{supported}/2 rule is:
\begin{equation}\label{schema}
{\small
\begin{array}{l}
\mbox{\texttt{supported(Act,Ts):-ant(ID,(}$\overline{X'} \cup \overline{T'}$\texttt{),Ts1)},$earlier(\overline{X'},\overline{Y},\overline{T_{1}'}$,\texttt{Ts2}),\texttt{Ts1<=Ts2}},\\
\mytab \mbox{\texttt{Ts2<Ts}}, 
\mbox{$sat\_rest\_time(\overline{T'},\overline{T_{1}'},\overline{T_{2}'}$,\texttt{Ts})},\mbox{\texttt{time(Ts)},\texttt{time(Ts2)},
$time(\overline{T_2'})$.}
\end{array}}
\end{equation}

\noindent where $earlier(\overline{X'}\cup \overline{Y}\cup \overline{T_{1}'}$,\texttt{Ts2}) represents the conjunction of the  event and fluent literals which must be satisfied before \texttt{Act} and their temporal constraints, $\overline{T_{1}'}$ represents the set of time variables belonging to these events and fluents together with some of the times in $\overline{T'}$,
 and \texttt{Ts2} represents the latest of these time variables.\footnote{The atom $time(\overline{T_2'})$ is shorthand for a requirement of all variables in $\overline{T_2'}$ to be less than or equal to the maximum time $n$.}
The predicate $\mathit{sat}\_rest\_time$ represents the conditions under which it will be possible for the $rest$ of that disjunct of the consequent of the rule to be satisfied, without violating time constraints; these conditions may take into account the antecedent time variables ($\overline{T'}$), the time variables in $\overline{T_{1}'}$ and the time variables in $\overline{T_{2}'}$.
The latter are time variables occurring in $rest$ but not elsewhere in the reactive rule. The action \texttt{Act} is supported at time \texttt{Ts} only if all these constraints are satisfiable.\footnote{Note that $\overline{T'}$ includes the time at which the antecedent was satisfied (\texttt{Ts1}) and $\overline{T_{1}'}$ includes \texttt{Ts2}. Also there is no need to specify \texttt{time(Ts1)} since \texttt{Ts1} is constrained by the \texttt{ant} atom.}  
Note that all constraints in an $n$-distant model must be satisfied for times $\leq n$, the upper bound.

To illustrate the \texttt{supported} predicate, consider again reactive rule (\ref{eq2}):
\begin{tabbing}
header\=space\=\kill
\textit{R}:
\>$request(Cust, Item, T)   \rightarrow$ \\
\>\mytab$ [(avail(Item,N,T_1)\wedge allocate(Cust,Item,N,T_2)\wedge T_2=T_1+1$\\
 \>\>\mytab $\wedge process(Cust,Item, T_3) \wedge T < T_2 < T_3 < T+4)$\\
\>\mytab $ \vee (apologise(Cust, Item, T_4) \wedge T_4 = T+4)]$
\end{tabbing}
\noindent The rule includes three different actions, $allocate$, $process$, $apologise$, for each of which there is a \texttt{supported}/2 definition in Reactive ASP:
\[{\small\begin{array}{l}
\texttt{supported(allocate(Cust,Item,N),Ts):-ant(1,(Cust,Item,T),T),T<Ts, }\\
\mytab\mytab\texttt{holds(available(Item,N),Ts-1),Ts<T2,T2<T+4,time(T2),time(Ts).}\\
\texttt{supported(process(Cust,Item),Ts):-ant(1,(Cust,Item,T),T), }\\
\mytab\mytab\texttt{holds(available(Item,N),T1-1),happens(allocate(Cust,Item,N),T1), }\\
\mytab\mytab\texttt{T<T1,T1<Ts,time(T1),Ts<T+4,time(Ts).}\\
\texttt{supported(apologise(Cust,Item),Ts):-ant(1,(Cust,Item,T),T),}\\
\mytab\mytab\texttt{Ts=T+4,time(Ts).}
\end{array}}\]

\noindent The action {\small \texttt{allocate(Customer,Item,N)}} is supported at time \texttt{Ts} if the rule antecedent is true,   the earlier conditions of the relevant disjunct of the consequent are true,
and there exists a time $T_2$ after \texttt{Ts} which is also before $T+4$ (i.e. when the order can be processed). 
The last two time constraints constitute the test for $\mathit{sat}\_rest\_time$ in the schema in (\ref{schema}).
A case where this would not be satisfiable is at timestamp \texttt{Ts}=$T+3$, where $T$ is the time at which the antecedent is satisfied.
Likewise,  the action \texttt{process(Cust,Item)} is supported at time \texttt{Ts} if the rule antecedent is true, 
 the earlier conditions of the  relevant disjunct of the consequent are true, including the allocation of the item, provided that \texttt{Ts < T+4} and \texttt{Ts} is within the time bound. Similarly, the agent may apologise to the customer if the antecedent is true and four cycles have elapsed.

\subsection{Summary}

\begin{table}[hbtp]
{\small \begin{tabular}{ l | l | l }\hline
1 & $n$-distance & Temporal constraints requiring all time parameters to be $\leq n$ 
 \vspace{-5pt}\\ 
\hline
 &  \multicolumn{2}{l}{Replace the explicit temporal constraints in reactive rules  with explicit}\\ 
 &  \multicolumn{2}{l}{\texttt{time} atoms and add a \texttt{time}  range declaration \texttt{time(0..n).} 
  Add explicit \texttt{time} atoms to }\\ 
 & \multicolumn{2}{l}{ASP parts dealing with (supported) actions and event theory - see items 9, 10, 11 below.} \vspace{-5pt}\\
\hline
2 & $\Aux$ & {A set of facts} \vspace{-5pt}\\ \hline
&  \multicolumn{2}{l}{Identical non-temporal facts in ASP syntax, rely on ASP built-ins for temporal facts}\vspace{-5pt}\\ \hline
3 & Time-stamped events & {$e(t_1,\ldots,t_{n},i)$} \vspace{-5pt}\\ \hline
&  \multicolumn{2}{l}{\texttt{happens(e(t1,...,tn),i)}}\vspace{-5pt}\\ \hline
4 & Time-stamped fluents & {$p(t_1,\ldots,t_{n},i)$} \vspace{-5pt}\\ \hline
& \multicolumn{2}{l}{\texttt{holds(p(t1,...,tn),i)}}\vspace{-5pt}\\ \hline
5 & {Initial state  fluents} & $p(t1,...,tn,0)$ \vspace{-5pt}\\ \hline
&  \multicolumn{2}{l}{\texttt{holds(p(t1,...,tn),0)}} \vspace{-5pt}\\ \hline
6 & $C_{post}$ facts & $\mathit{initiates(e,p) /}$  $\mathit{terminates(e,p)}$ \vspace{-5pt} \\ \hline
& \multicolumn{2}{l}{Identical  ASP facts}\vspace{-5pt}\\ \hline
7 & $C_{pre}$ constraints & $false \leftarrow$  $body(T,T+1)$ \vspace{-5pt} \\ \hline
& \multicolumn{2}{l}{
\begin{minipage}[t]{4.5in}{{\small\texttt{:-\mbox{$\mathit{aspbody}$}(Ts-1,Ts),time(Ts-1),time(Ts).}}\\ $aspbody$ is $body$ but with the ASP reified syntax for events and fluents
}\end{minipage}}\vspace{-5pt}\\ \hline
8 & Reactive rules  &
\begin{minipage}[t]{3.5in}{
$\forall \overline{X} \forall \overline{T}$
$ [antecedent(\overline{X} \cup \overline{T})$
$\rightarrow$
$\exists \overline{Y} \exists \overline{T_1}$
$ consequent($
$\overline{X'} \cup \overline{T'}, \overline{Y} \cup \overline{T_1})]$ }\end{minipage} 
 \vspace{-5pt}\\ \hline & 
 \multicolumn{2}{l}{
 \begin{minipage}[t]{5.1in}{Domain dependent part: \\
{\small\texttt{ant(ID},($\overline{X'} \cup \overline{T'}$),\texttt{Ts}):-$antecedent(\overline{X} \cup \overline{T})$,$max(\overline{T}$,\texttt{Ts}),\texttt{time(Ts)}.} (where $\overline{X'}$, $\overline{T'}$ are the \\
variables in $\overline{X}$, $\overline{T}$, respectively, in the antecedent, that also occur in the consequent.)\\
{\small\texttt{cons(ID,(}$\overline{X'} \cup \overline{T'}$),\texttt{Time},\texttt{Ts}):-\texttt{ant(ID,(}$\overline{X'} \cup \overline{T'}$),\texttt{Time}),} \\
\mytab\mytab 
{\small$consequent_{i}(\overline{X'} \cup \overline{T'},\overline{Y} \cup \overline{T_1}), max(\overline{T'} \cup \overline{T_1}$,\texttt{Ts}),\texttt{time(Ts)}.}\\
Domain independent part:\\
{\small\texttt{:-ant(ID,Args,Ts),\mbox{not }consTrue(ID,Args,Ts),time(Ts).}}\\
{\small \texttt{consTrue(ID,Args,Ts):-cons(ID,Args,Ts,Ts1),time(Ts1).}}
}\end{minipage}} 
\vspace{-5pt}\\ \hline 
9 & Event theory & 
An emergent  property of the OS, not
part of the program  \vspace{-5pt} \\ \hline
&  \multicolumn{2}{l}{
\begin{minipage}[t]{4.8in}{Explicitly part of the program\\ 
{\small\texttt{holds(P,Ts):-initiates(E,P),happens(E,Ts),time(Ts).}}\\	
{\small\texttt{holds(P,Ts):-holds(P,Ts-1),not broken(P,Ts),time(Ts-1),time(Ts).}}\\
{\small\texttt{broken(P,Ts):-terminates(E,P),happens(E,Ts),time(Ts).}}}\end{minipage}}
\vspace{-5pt}\\ \hline
10 & Supported actions  & 
An emergent  property of the OS, not
part of the program  \vspace{-5pt} \\ \hline & 
 \multicolumn{2}{l}{
 \begin{minipage}[t]{4.8in}{Explicitly part of the program\\ 
{\small\texttt{supported(Act,Ts):-ant(ID,(}$\overline{X'} \cup \overline{T'}$\texttt{),Ts1)},$earlier(\overline{X'}, \overline{Y}, \overline{T_{1}'}$,\texttt{Ts2)},\texttt{Ts1 <= Ts2},}\\
{\small\mytab \texttt{Ts2<Ts},$\mathit{sat}\_rest\_time(\overline{T'},\overline{T_{1}'},\overline{T_{2}'}$,\texttt{Ts}),\texttt{time(Ts)},\texttt{time(Ts2)},$\mathit{time(\overline{T_2'})}$.}
}\end{minipage}}\vspace{-5pt}\\ \hline
11 &  \begin{minipage}{1.2in}{Abduction of \\ supported actions}\end{minipage} & 
Part of the OS, not part of the program  \vspace{-5pt}\\ \hline & 
 \multicolumn{2}{l}{  
\begin{minipage}[t]{4.5in}{Explicitly part of the program\\ {\small\texttt{0\{happens(Act,Ts)\}1:-supported(Act,Ts),time(Ts),Ts>0.}}}\end{minipage}}\vspace{-5pt}\\ \hline
\end{tabular}}
\caption{Mapping details  of KELPS to Reactive ASP. 
}
\label{mappingdetails}
 \end{table}
We summarise the mapping of an $n$-distant KELPS framework $<\R,\C,\Aux,n>$  into Reactive ASP in Table~\ref{mappingdetails}, in which
 for each item number the first row shows the KELPS feature and its representation in KELPS and the second row shows the ASP representation.
As can be seen, parts of  Reactive ASP rules are identical, or almost identical, to KELPS but for the reified syntax in ASP (items 2, 3, 4, 5, 6). There are two major differences evident between   the two paradigms:  the mapping of the implicit concepts and properties of the operational semantics of KELPS into explicit program rules in ASP (items 9, 10, 11); the mapping of reactive rules, which are conceptually goals in KELPS and become constraints in ASP (item 8).
Note also that the $n$-distance constraints of $n$-distant KELPS are mapped 
into a program rule (item 1), and the addition of \texttt{time} conditions in items  8, 10 and 11.

The translation of KELPS to ASP is as elaboration tolerant \cite{Mccarthy98elaborationtolerance} as the original KELPS.
In particular, the normal rules defining \texttt{ant} and \texttt{cons} can fully capture any expression in the KELPS antecedents and consequents, and the shared parameters between \texttt{ant} and \texttt{cons} ensure that the connection between antecedent and consequent of reactive rules is preserved.

%% file: Formal_Final_corr.tex
\section{Formal Results}\label{sec:theory}

In this section we show 
that the $n$-distant KELPS framework and its mapping to Reactive ASP as defined in Section~\ref{sec:translation} compute the same reactive models.
In particular, we show that the mapping is sound and complete; that is,  any answer set of the resulting ASP program corresponds to an $n$-distant   KELPS reactive model and any $n$-distant reactive model corresponds to an answer set. We first focus on the soundness.

\begin{definition}\label{def:mkelps}
Let $P$ be an $n$-distant  KELPS framework $<\R,\C,\Aux,n>$, with initial state $S_0$ and external events $ext^{*}$. Let $PA$ be the mapping of $P$ into Reactive ASP and $M$ be an answer set for $PA$.
Based on $M$ we define $M_{KELPS}$ as follows.\footnote{Recall that the  facts in $\Aux$ are the same in both a KELPS program and the corresponding ASP program and hence they will be a subset of $PA$.
Note also that $holds(p,i)$ is the reified form of the (shortened) timestamped fluent atom $p(i)$ (see Section~\ref{sec:vocabulary}) and $happens(e,i)$ is the reified form of the (shortened) timestamped event atom $e(i)$.}

$\begin{array}{l}
\mbox{Let } S_i^{*} = \{p(i): holds(p, i) \in M\}, 0\leq i \leq n, \\
\mytab act_i^{*} = \{e(i): happens(e,i) \in M \mbox{ and } e(i)  \not\in ext^{*}\}, 
1\leq i \leq n\mbox{,}\\
\mytab S_{i}= S_{n}, i>n, \mbox{ and }  act_{i}^{*}=\emptyset, i>n\mbox{.}\\
\mbox{Then } M_{KELPS} = S^{*} \cup ext^{*}  \cup act^{*} \cup \Aux, \mbox{ where }\\
\mytab 
S^{*} = S_0^* \cup S_1^* \cup \ldots \cup S_{n}^* \cup S_{n+1}^*\cup\ldots
\mbox{ and }
act^{*} = act_1^* \cup act_2^* \cup \ldots \cup act_{n}^{*}\mbox{.}
\end{array}$
\end{definition}
\noindent We next show that $M_{KELPS}$ is an $n$-distant reactive model of $P$. 
\begin{mytheorem}[Soundness]\label{soundnessTheorem}
Let $P$, $PA$, $M$ and $M_{KELPS}$ be as defined in Definition~\ref{def:mkelps}. Then $M_{KELPS}$ is a reactive model of $P$. 
\begin{proof} 
We need to show the following four properties:
\begin{description}
\item[(i)]  $S_0 =$ the initial state\\
$S_{i+1}=succ(S_{i}, ev_{i+1})$, $0\leq i < n$, where $succ(S_{i}, ev_{i+1})$ =\\
\mytab($S_{i} - \{p:terminates(a,p)\in C_{post}\wedge a \in ev_{i+1}\}$)
$\cup$\\
\mytab\mytab$\{p: initiates(a,p) \in C_{post} \wedge a \in ev_{i+1}\}$.
\item[(ii)]$C_{pre}$ is true in $M_{KELPS}$.
\item[(iii)] \R\   is true in $M_{KELPS}$.
\item[(iv)] Every action in $act^{*}$ is supported in the sense of Definition~\ref{reactivemodel}.
\end{description}

We recall that the inclusion in $PA$ of the \texttt{time} atoms as conditions  in $C_{pre}$ and ET is  simply for safety and grounding. 
We show (i) - (iv) below:

\begin{description}
\item[(i)] Note first that by definition $S_0^{*}$ is the initial timestamped state and hence $S_0$ = the initial state.
Next, recall from subsection~\ref{sec:vocabulary} that events are  assumed to occur independently, even if they occur at the same timestamp.
In terms of the vocabulary of the program $PA$, this property means
\begin{equation}\label{eq:succ}
\begin{array}{l}
holds(P,T+1) \leftrightarrow (\exists E  (initiates(E,P) \wedge happens(E,T+1) \wedge T+1 \leq n)\\
\mytab  
 \vee  (holds(P,T) \wedge\\
 \mytab\mytab \neg\exists E (terminates(E,P) \wedge happens(E,T+1) \wedge T+1 \leq n))
\end{array}
\end{equation}
The property in Equation~(\ref{eq:succ})
is included explicitly as part of $PA$ (the event theory ET) and is thus true in $M$ and hence also in $M_{KELPS}$, since the property only refers to events and fluents in $M$.
\item[(ii)] The $PA$ version of $C_{pre}$ is true in $M$. The only predicates mentioned in that version of $C_{pre}$ are $happens$, $holds$,  auxiliary  and \texttt{time} predicates. 
$M_{KELPS}$ contains exactly the same  set of events as $M$,   any fluents in a rule in $C_{pre}$ have a timestamp earlier than an event in that rule and no event occurs after time $n$. Thus the KELPS $C_{pre}$ (without a temporal constraint)  is also true in $M_{KELPS}$.
\item[(iii)]
Consider a reactive rule $r$  in $P$ of the form shown in Equation~(\ref{generalrule}) (repeated here)
\[\forall \overline{X} \forall \overline{T} [antecedent(\overline{X} \cup \overline{T})\rightarrow \exists \overline{Y} \exists \overline{T_1} consequent(\overline{X'} \cup \overline{T'}, \overline{Y} \cup \overline{T_1})]
\]
and recall that if $r$ has an $ID$=\texttt{id} it is mapped to the following in $PA$ (Equation~(\ref{eq16}) and the general reactive rule constraint Equation~(\ref{eq10})).
{\small\[\begin{array}{l}
		\texttt{ant(id,($\overline{X'} \cup \overline{T'}$),\texttt{Ts}):-$antecedent(\overline{X} \cup \overline{T})$,$max(\overline{T}$,\texttt{Ts}$)$,\texttt{time(Ts)}}.\\
		\texttt{cons(id,($\overline{X'} \cup \overline{T'}$),\texttt{Time},\texttt{Ts}):-ant(id,$\overline{X'} \cup \overline{T'}$),\texttt{Time}),} \\
		\mytab\mytab consequent_{i}(\overline{X'} \cup \overline{T'},\overline{Y} \cup \overline{T_1}), max(\overline{T'} \cup \overline{T_1},\texttt{Ts}),\texttt{time(Ts)}. \\
   \texttt{:-ant(ID,Args,Ts),\mbox{not }consTrue(ID,Args,Ts),time(Ts).}\\
   \texttt{consTrue(ID,Args,Ts):-cons(ID,Args,Ts,Ts1),time(Ts1).}
		\end{array}
		\]}
Suppose for contradiction that $r$ is false in $M_{KELPS}$.
Therefore, it must be the case that for some $\overline{X} \cup \overline{T}$ and timestamp {\small\texttt{ts} }({\small\texttt{ts}} $\leq n$) such that $antecedent(\overline{X} \cup \overline{T})$ is true there is no $\overline{Y}$ and  $\overline{T_1}$ at a timestamp \texttt{ts1} (\texttt{ts1} $\leq n$)  for which $consequent(\overline{X'} \cup \overline{T'},\overline{Y} \cup \overline{T_1})$ is true.
Then by construction of $M_{KELPS}$ from $M$ and the above mapping,
{\small\texttt{ant(id,args,ts)}} (where {\small\texttt{args}}=$\overline{X} \cup \overline{T}$)
is true in $M$, but {\small\texttt{cons(id,args,ts,Ts1)}} is not true in $M$ for any \texttt{Ts1}. Consequently, {\small\texttt{consTrue(id,args,ts)}} is not true in $M$, contradicting that the constraint {\small\texttt{:-ant(ID,Args,Ts),not consTrue(ID,Args,Ts),time(Ts)}}
is satisfied in $M$.
Therefore $r$ is true in $M_{KELPS}$.
\item[(iv)] Actions (i.e. {\small\texttt{happens}} atoms not related to external events) can be in  $M$ only through  instances of the rule
{\small\texttt{0\{happens(Act,Ts)\}1:-supported(Act,Ts),time(Ts),Ts>0}}, hence actions \texttt{Act} at time $Ts$ included in $M$ must satisfy {\small \texttt{supported(Act,Ts)}}.
Moreover, as argued earlier in Section~\ref{supportedMapping}, the definition of \texttt{supported} in $PA$ corresponds exactly to that in the KELPS framework, hence  the actions in $act^{*}$ of $M_{KELPS}$ are also supported.
\end{description}
\end{proof}
\end{mytheorem}

Theorem~\ref{soundnessTheorem} shows that  answer sets of $PA$ correspond to reactive models of the $n$-distant KELPS framework $P$.
In Theorem~\ref{completeTheoremV2} we show that if   $P$ is an $n$-distant  KELPS framework and $PA$ the corresponding  ASP  program, then if $M_{P}$ is   an $n$-distant reactive model  of $P$ there will be a corresponding answer set $A$ of $PA$.

\begin{mytheorem}[Completeness]\label{completeTheoremV2}
Let $P$ be an  $n$-distant KELPS framework $<\R,\C,\Aux,n>$ with initial state $S_0$ and external events $ext^{*}$, and $PA$ be the 
Reactive ASP mapping of $P$.
If $M_{P}$ is an $n$-distant reactive model  of $P$, then the program 
 $PA$ has an answer set $M$ such that $M_{KELPS}$ = $M_{P}$.
\end{mytheorem}

Before giving the proof of Theorem~\ref{completeTheoremV2} we define some notation.
\begin{definition}\label{def:completeNotation}
Let $P$, $PA$, $M_{P}$ and  $ext^{*}$ be as given in the statement of the theorem and 
$acts^{*}$ be the set of timestamped actions in $M_{P}$.
Then \\
- $PA_{ncc}$  is the program consisting of  the normal rules of  $PA$,   but neither  the constraints\\
 nor  the choice rule, augmented by the set of 
facts $H = \{$   {\small\texttt{happens(a,t)}} $\mid$   {\small\texttt{a(t)}}  $\in acts^{*}\}$.\\
- $A_{ncc}$ is the answer set of $PA_{ncc}$.\footnote{In Step 1 we show $A_{ncc}$ exists and is unique.}\\
- For $0\leq i\leq n$, $F_i$ is the set of fluents  given by $F_i$ = $\{p(i): holds(p,i) \in A_{ncc}\}$.\\
- $PA^{-}$ is the program   $PA$   without the constraints.
\end{definition} 
The proof of Theorem~\ref{completeTheoremV2} has several steps that are outlined next.
\begin{description}
\item[Step 1:] Note that $PA$ includes \texttt{happens} facts corresponding to the external events $ext^{*}$ of the KELPS program $P$. We   construct from $PA$ a (reduced) program $PA_{ncc}$ that {\em excludes} constraints and the choice rule, but {\em includes} the set $H$ of \texttt{happens} facts corresponding to the actions $acts^{*}$ in $M_{P}$. We  show that $PA_{ncc}$ is locally stratified by constructing a stratification \cite{GelfondChapter}  and therefore conclude it has a unique answer set, denoted $A_{ncc}$.
\item[Step 2:] We show that  the set of states of $M_{P}$ up to $S_{n}$  and the set of fluents in the answer set $A_{ncc}$ of $PA_{ncc}$ (expressed through \texttt{holds/2} atoms)  are the same.
\item[Step 3:] We show that for every \texttt{happens(a,t)} fact in $A_{ncc}$, where $a$ is not an external event,  the fact \texttt{supported(a,t)} is also in $A_{ncc}$.
\item[Step 4:] By considering the program $PA^{-}$, derived from $PA_{ncc}$ by reinstating the choice rule and removing the \texttt{happens} facts in $H$, we show, through iterative application of the Splitting Theorem as described in Lemma~\ref{answerSetconstruction},
that the answer set  of $PA_{ncc}$ is an answer set of $PA^{-}$. 
\item[Step 5:] Finally, we show that the answer set of $PA_{ncc}$ is an answer set of $PA$. 
\end{description}

\begin{proof}
\paragraph{Step 1:} 
We show that $PA_{ncc}$ is locally stratified. Hence $PA_{ncc}$  has a unique answer set  $A_{ncc}$ \cite{GelfondChapter}. Later, in Step 4, this answer set will be constructed.

The stratification, given in Table~\ref{stratification},  is based on the following observations. $PA_{ncc}$ has a finite grounding and the ground instances of its rules can be placed into strata   based on the timestamp argument $Ts$. 
The lowest stratum (denoted 0), includes  atoms in $C_{post}$ and  $\Aux$, together with \texttt{time} atoms. 
For each timestamp $Ts\geq 0$ there are strata   $Ts\mbox{-}i$, $Ts\mbox{-}ii$ and $Ts\mbox{-}iii$, which are ordered according to the value of $Ts$, such that each $Ts\mbox{-}i$ is in the stratum immediately preceding $Ts\mbox{-}ii$,  each $Ts\mbox{-}ii$ is in the stratum immediately preceding $Ts\mbox{-}iii$, and stratum 0 is least in the order. 
(Note that \texttt{happens} and \texttt{supported} atoms can only occur at timestamps $> 0$, so in fact there is no  stratum labelled $0\mbox{-}i$.)
There is also a strata $n$ for all \texttt{consTrue} atoms. The strata are ordered (lowest to highest) by  $0 <0\mbox{-}ii < 0\mbox{-}iii <1\mbox{-}i <1\mbox{-}ii < 1\mbox{-}iii < 2\mbox{-}i <\ldots < n\mbox{-}i < n\mbox{-}ii < n\mbox{-}iii < n$.\footnote{The final stratum $n$ is so named as all \texttt{consTrue} atoms, regardless of their timestamp, can refer to timestamped \texttt{cons} atoms with timestamps up to $n$.}
It can be checked that each ground rule instance only refers to positive body atoms in the same or a lower stratum, and only refers to negative body atoms in a lower stratum.
\begin{table}[htbp]
\begin{tabular}{l|l}
\hline 
Strata & Rules\\ \hline
0 & \texttt{initiates}, \texttt{terminates}, \texttt{time} and $\Aux$ facts,\\
$0\mbox{-}ii$  &  facts for \texttt{holds} at timestamp 0 (i.e. initial state)\\
$0\mbox{-}iii$ &  rules for \texttt{ant} or \texttt{cons}  at timestamp 0\\
$Ts\mbox{-}i$ &  rules for \texttt{broken}, \texttt{happens} and  \texttt{supported} at timestamp $Ts$, $Ts>0$\\
$Ts\mbox{-}ii$ & rules for \texttt{holds} at timestamp $Ts$, $Ts>0$\\
$Ts\mbox{-}iii$ & rules for \texttt{ant} and  \texttt{cons}  at timestamp $Ts$, $Ts>0$\\
$n$ & rules for \texttt{consTrue}\\
\hline
\end{tabular}
\caption{Strata in Reactive ASP.}
\label{stratification}
\end{table}

\paragraph{Step 2:} 
Lemma~\ref{lem:SameState} shows that for $0\leq i \leq  n$ the state $S_i$ of the KELPS model $M_P$ and the set of fluents holding at time $i$ in  the answer set $A_{ncc}$ are the same. \\

\begin{lemma}\label{lem:SameState}
Using notation in Definition~\ref{def:completeNotation}, for each $i$, $0\leq i\leq n$,  $F_i$ =   the state $S_i$ of  $M_P$.
\begin{proof}
Note first that by construction $M_P$ and $A_{ncc}$ have the same set of events, both user actions and external events, the same initial state and the same definition of $C_{post} $.
Initially, $F_0 = S_0$ by definition of $M_{P}$ and $A_{ncc}$. Assume as inductive hypothesis (IH) that $F_i = S_i$, for some $i$, $0 \leq i <  n$. We argue $F_{i+1}=S_{i+1}$.
By Definition \ref{def:modelstructure}, $S_{i+1}=$\\
$(S_i-\{p:terminates(e,p) \in C_{post} \wedge e \in ev_{i+1}\})$$ \cup$$\{p:initiates(e,p) \in C_{post} \wedge e \in ev_{i+1}\}$. 
From the ET in  $PA$ (and thus in $PA_{ncc}$), and noting that \texttt{time(i+1)} and \texttt{time(i)} are true,  \texttt{holds(p,i+1)}  is  true  if and only if
$\exists$\texttt{e: initiates(e,p)} and \texttt{happens(e,i+1)} or 
  \texttt{holds(p,i)} and \texttt{not} ($\exists$ \texttt{e : terminates(e,p)} and \texttt{happens(e,i+1)}).
Thus, $F_{i+1}$ = $\{p:$ \texttt{holds(p,i+1)}$\in A_{ncc}\}$ = 
$\{p:(\exists  e:\texttt{initiates(e,p)} \wedge$\texttt{happens(e,i+1)}) $\vee$\\
 $($\texttt{holds(p,i)} $\wedge  \neg (\exists e:\mbox{\texttt{terminates(e,p)}} \wedge $\texttt{happens(e,i+1)}))$\}$.
By (IH)  and because ET and KELPS have the same $C_{Post}$ and same user actions and external events, $F_{i+1}$ 
= $\{p: (\exists  e:initiates(e,p) \wedge e\in ev_{i+1}) \vee  (p(i) \wedge \neg ( \exists e : terminates(e,p) \wedge e\in ev_{i+1} )) \} = S_{i+1}$.
\end{proof}
\end{lemma}

\paragraph{Step 3:}
By the properties of an $n$-distant reactive KELPS model,  it holds that $\R \cup \C$ is true in $M_{P}$ and every user action in $M_{P}$  is supported, satisfying conditions of Definition~\ref{reactivemodel}. 
Furthermore, both the actions in $H$ and  external events  are common to $M_{P}$ and $A_{ncc}$, by definition, respectively,  of  $M_{P}$ and  $PA_{ncc}$,  and, as shown in  Lemma~\ref{lem:SameState}, the values of fluents at each time point up to $n$ will therefore be the same.

By construction, because $M_P$ is an $n$-distant reactive model of KELPS, the maximum time of occurrence of any \texttt{happens} fact in $H$ is $n$. 
Partition the  \texttt{happens} facts in $H$ by their time of occurrence, and form a sequence  of sets of actions. Denote by $H(i)$ the set of  actions occurring at time $i$, i.e. $H(i)=\{a: $ \texttt{happens(a,i)} $ \in H\}$ and consider an action $u$ in $H(i)$. 
We show that  $u$ is supported at time $i$, namely that \texttt{supported(u,i)} is in $A_{ncc}$. 
By definition of $M_P$, all  actions in $acts^{*}$ are supported, which by Definition~\ref{reactivemodel} 
means there is a rule $r$ and instance $r_u$ such that (i) the antecedent of $r_u$ occurs before $i$,  (ii) actions and fluents in some disjunct of the consequent of $r_u$ can be separated into those that occur at times $t1$, where $t1 < i$,  the instance of $u$ at $i$, and those that should occur at or after $i$, and (iii) that the   temporal constraints for the the latter actions and fluents are satisfiable.

As explained below Equation~(\ref{schema}), these  conditions are captured in Reactive ASP, respectively, by the conditions \texttt{ant(ID,(}$\overline{X'} \cup \overline{T'}$\texttt{),Ts1)},  $earlier(\overline{X'},\overline{Y},\overline{T_{1}'}$,\texttt{Ts2}), \texttt{Ts1<=Ts2}, \texttt{Ts2<Ts},
and $sat\_rest\_time(\overline{T'},\overline{T_{1}'},\overline{T_{2}'}$,\texttt{Ts)},
where the relevant instantiation(s) would be  that ID is the identifier of rule $r$, $Ts$ is $i$,  earlier times $t1$ in $\overline{T_{1}'}$ are all less than $Ts$, and  times  later than $Ts$ are in the set 
$\overline{T_{2}'}$.

\paragraph{Step 4:}

Now  consider   the program $PA^{-}$.  By considering the reduct of $PA^{-}$ w.r.t the interpretation $A_{ncc}$, we next show that $A_{ncc}$ is an answer set of $PA^{-}$ by iteratively applying the Splitting Set Theorem as described in Lemma~\ref{answerSetconstruction}.\footnote{In what follows we work with the grounding of $PA^{-}$ and $PA_{ncc}$.} 
Since $A_{ncc}$ is the answer set of  $PA_{ncc}$, by Definition~\ref{MarkReduct}  the answer set of the reduct of $PA_{ncc}$ w.r.t $A_{ncc}$ will be $A_{ncc}$. In Lemma~\ref{4.2Step4} we show that the answer set of the reduct of $PA^{-}$ w.r.t. $A_{ncc}$ is also $A_{ncc}$ by iteratively constructing the answer sets according to Lemma~\ref{answerSetconstruction} to both reducts and showing that they are the same. 

\begin{lemma}\label{4.2Step4}
The answer set $A_{ncc}$ of $PA_{ncc}$ is an answer set of  $PA^{-}$.
\begin{proof}
Consider the reducts of $PA_{ncc}$ and $PA^{-}$ w.r.t. $A_{ncc}$, the answer set of $PA_{ncc}$. First,  note that the   $PA_{ncc}$ and $PA^{-}$ are almost the same, differing only in the following way: the rules with \texttt{happens} in the head related to actions are facts in $PA_{ncc}$ and choice rules in $PA^{-}$.~\footnote{Other rules with \texttt{happens} in the head, i.e. external event \texttt{happens}, are facts in both $PA_{ncc}$ and $PA^{-}$.}
Second, observe that the answer set $A_{ncc}$ will include all atoms of the form \texttt{happens(event,t)} (where event may be an external  event or a generated action), that appear as facts in $PA_{ncc}$. These observations guarantee that the reducts of the two programs will therefore differ in only one respect, namely the rules for \texttt{happens} atoms.
In the reduct of $PA_{ncc}$ these are simply facts, whether \texttt{event} is an external event or an action, whereas in the reduct of $PA^{-}$ they are either also facts for external events, or ground rules of the form 
\begin{equation}\label{eq:happens}
\begin{array}{l}
\texttt{happens(act,t):-supported(act,t),time(t),t>0.} 
\end{array}
\end{equation}
\noindent for those actions \texttt{act} and timestamps \texttt{t} where \texttt{happens(act,t)} is a fact in the reduct of $PA_{ncc}$.  These latter rules are derived from the choice rules in $PA^{-}$, by item 3 of Definition~\ref{MarkReduct}, given the observation above.\footnote{In KELPS there are no occurrences of negated \texttt{happens} literals in the antecedent or consequent of rules, hence the application of item 1 of Definition~\ref{MarkReduct} will yield the same results.} 
The  reduct of $PA^{-}$ can be stratified according to the strata   given in Table~\ref{stratification}; in particular, the rules of the form in Equation~(\ref{eq:happens})
will be in the strata $Ts\mbox{-}i$ for $Ts=\mbox{\texttt{t}}$,   the same strata as the \texttt{happens} facts at time \texttt{t} in $PA_{ncc}$. Therefore the iterative answer set construction of Lemma~\ref{answerSetconstruction} can be applied to both reducts in parallel. 

Throughout the  lemma we will use the following  notation.
We denote the programs corresponding to $\pi_j$ in Lemma~\ref{answerSetconstruction} derived from the iterative splitting of the reducts of $PA_{ncc}$ and $PA^{-}$ by, respectively, $B_{j}$ and $C_{j}$, where $j$ is one of the   strata in Table~\ref{stratification}, and the corresponding answer sets  $S_j$  of $bot_{{\cal H}_j}(\pi_j)$ by $AB_{j}$ (answer set of $bot_{{\cal H}_j}(B_j)$)  and $AC_{j}$ (answer set of $bot_{{\cal H}_j}(C_j)$).
Recall that by construction, $\pi_k=ev_{{\cal H}_{k-1}}(top_{{\cal H}_{k-1}}(\pi_{k-1}),S_{k-1})$,
where, depending on the reduct $\pi_{k-1}$ is either  $B_{k-1}$ or $C_{k-1}$ and $S_{k-1}$ is either $AB_{k-1}$ or $AC_{k-1}$.
Finally, when we refer to $k$ as a stratum, we mean the set of rules in that stratum (i.e. $k = bot_{{\cal H}_{k}}(\pi_{k})$).

Let $k$ be a stratum
and assume as inductive hypothesis that for all preceding strata $j < k$ the   answer sets  $AB_{j}$ and  $AC_{j}$
are identical. 
There are several cases depending on the type of  strata  of $k$.
 \begin{description}
\item[$k$ = stratum 0:] The strata in the two reducts are identical by construction and hence the programs $bot_{{\cal H}_0}(B_0)$ and $bot_{{\cal H}_0}(C_0)$ (=$k$) will be the same. Hence the answer sets $AB_0$ and $AC_0$ will also be the same. Note that after applying partial evaluation to $top_{{\cal H}_0}(B_0)$ or $top_{{\cal H}_0}(C_0)$ all 
\texttt{time} atoms  will be   eliminated as \texttt{time} atoms occur only positively in clauses in Reactive ASP. 
Moreover, in any clause in which a true $\Aux$ atom occurs positively in $top_{{\cal H}_0}(C_0)$ or $top_{{\cal H}_0}(B_0)$ the atom will be removed after partial evaluation, whereas if it occurs negatively the clause will be eliminated. Similarly, in any clause in which an $\Aux$ atom that is not true occurs negatively in $top_{{\cal H}_0}(C_0)$ or $top_{{\cal H}_0}(B_0)$ the atom will be removed after partial evaluation, whereas if it occurs positively the clause will be eliminated.
In particular, this means that in program $C_1 = ev_{{\cal H}_{0}}(top_{{\cal H}_{0}}(C_0),AC_0)$ the \texttt{time} atom and $\Aux$ atom \texttt{t>0} will have been eliminated from the clauses of the form given by Equation~(\ref{eq:happens}), where \texttt{t} will be a particular ground timestamp value\footnote{See explanation below Equation~(\ref{eq:happens}).}
because they are all true facts in $bot_{{\cal H}_0}(C_0)$ and hence will be in the answer set $AC_0$.

\item[$k$= stratum \texttt{t}-$i$:] 
By hypothesis  $AB_{k-1}=AC_{k-1}$ and as noted above the difference in the strata is only in the rules for \texttt{happens}.
In the case of program $B_k$ the answer set $AB_k$ of $bot_{{\cal H}_k}(B_k)$ will include \texttt{happens}, \texttt{supported} and \texttt{broken} atoms at timestamp \texttt{t}. The \texttt{happens} atoms come directly from the \texttt{happens} facts, and the other atoms will be derived, through partial evaluation with answer sets of previous splits, from rules where the body atoms are true in those answer sets.
In the case of program $C_k$, the \texttt{supported} and \texttt{broken} atoms at timestamp  \texttt{t} in the answer set $AC_{k}$ of  $bot_{{\cal H}_k}(C_k)$ will similarly be derived through partial evaluation with answer sets of previous splits, and the \texttt{happens} atoms will be derived for exactly those supported actions. 
Since, as we have shown in Step 3, all such atoms are indeed supported, the answer set $AC_k$ will include exactly the same \texttt{happens} atoms as $AB_k$. 
Thus the answer sets $AB_k$ and $AC_k$ are the same.
\item[$k$ = stratum  \texttt{t}-$ii$ or  \texttt{t}-$iii$:] 
Similar, but simpler, reasoning to the previous case   allows to conclude that 
the rules in  $bot_{{\cal H}_k}(B_k)$ and  $bot_{{\cal H}_k}(C_k)$ are  the same because the 
rules in strata \texttt{t}-$ii$ or \texttt{t}-$iii$ of the two reduct programs $AB_0$ and $AC_0$ are the same. Hence the answer sets of strata $k$, $AB_k$ and $AC_k$, are also the same.
\item[$k$ = stratum $n$:] Arguing as in the previous cases, the answer sets $AB_k$ and $AC_k$ are the same. That is, $AB_n=AC_n$. As this is the highest strata, there is no partial evaluation to be made.
\end{description}
Finally, the answer sets of the two reducts w.r.t. $A_{ncc}$, namely $B_0$ (reduct of $PA_{ncc}$)  and $C_0$ (reduct of $PA^{-}$) are both equal to 
$\bigcup_{j=0}^{j=n} AB_{j} = \bigcup_{j=0}^{j=n} AC_{j}$, where $j$ ranges over the strata $0, 0\mbox{-}ii, 0\mbox{-}iii, 1\mbox{-}i, \ldots, n\mbox{-}iii,n$.
\end{proof}
\end{lemma}


\paragraph{Step 5:}
Step 4 showed that $A_{ncc}$ is an answer set of   $PA^{-}$. Consider now program $PA$, i.e. by reinstating the constraints into $PA^{-}$. If none of the constraints is violated by $A_{ncc}$ then $A_{ncc}$ will be an answer set of $PA$.
 Suppose  for contradiction that $A_{ncc}$ is not an answer set of $PA$.
This means that one or more of the constraints in $PA$ must have been violated (i.e. the respective constraint body is satisfied) by $A_{ncc}$. There are two categories of constraints in $PA$, the pre-conditions in $C_{pre}$ and the reactive rule constraint \texttt{:- ant(ID,X,Ts),not consTrue(ID,X,Ts),time(Ts).}
There are two cases.\\ \\
\noindent {\em Case 1}:  $A_{ncc}$  does not satisfy one of  the constraints in $C_{pre}$. 
By Step 2, $A_{ncc}$ has the same fluents as  $M_{P}$ up to time $n$. Also, $A_{ncc}$ and $M_{P}$ have the same set of external events and by construction the same set of actions (the set $H$).
Thus if $A_{ncc}$ does not satisfy a pre-condition constraint, nor will $M_{P}$, contradicting that $M_{P}$ is a reactive model of $P$. 
Moreover, for a constraint that inhibits co-occurrence of events, all the events in the constraint have the same timestamp and if they all occur in $A_{ncc}$, then they all occur in $M_{P}$. 
For a constraint that imposes a precondition on an event,
because the fluents in $C_{pre}$ have timestamps earlier than an event in the same $C_{pre}$ rule, the fact that $A_{ncc}$ has states only up to $n$ does not matter.\\ \\
\noindent {\em Case 2}: $A_{ncc}$  does not satisfy the reactive rule constraint.
That is, for some \texttt{id}, \texttt{x}  and \texttt{t}, \texttt{ant(id,x,t)} is true  and \texttt{consTrue(id,x,t)} is false in $A_{ncc}$. This means, by the definition of \texttt{consTrue} that there can be no \texttt{Ts1}$\geq$\texttt{t} in the range [$0, \ldots, n$] such that \texttt{cons(id,x,t,Ts1)} is true. 
That is,  there is a reactive rule in $P$ with its antecedent true but for no time $\leq n$ can its consequent be made true. But that means there is a reactive rule in $P$ that is not true in $M_{P}$, which is not the case. 
\end{proof}

%% file: Remainder_Final_corr.tex
\section{Reactive ASP Functionality Beyond KELPS}\label{sec:extensions}
In this section we describe how some features of ASP inherited by Reactive ASP can be exploited to enhance its flexibility and functionality beyond those of KELPS.
The   features we exploit are the model generation paradigm, weak constraints and the explicit representation of choice.
More specifically, model generation allows reasoning to take into account any ramifications of actions (for {\em prospective} behaviour), in which  Reactive ASP can look ahead to reason about possible evolutions of its current state, thus informing current decisions.
Explicit choice and weak constraints
allow to specify the type of model ({\em preemptive}, {\em proactive} or {\em reactive}) preferred and to rule out unwanted models and rank those models that are not ruled out.
All of these new behaviours are possible  because reactive ASP  generates complete answer sets, and thus complete complete courses of actions and resulting fluents,
up to a maximum time-range $n$.
In particular, we explain how to achieve preemptive, proactive and prospective reasoning, and introduce a hybrid system that combines  the two frameworks of KELPS and ASP into one enhanced integrated framework that uses the best features of both.

\paragraph{Reasoning with Priorities}
In KELPS all constraints are hard constraints and   are used only to express preconditions of actions and to restrict co-occurrences of events.
But in ASP constraints can be hard or soft (weak constraints) and can be used to express many other features, including preferences.

One useful application of preferences would be where the consequent is disjunctive and we would like to express  preferences amongst the disjuncts. 
This can be achieved systematically when mapping to ASP as follows: Suppose the disjuncts in KELPS are written in order of preference from high to low. In the mapping to ASP, first give \texttt{cons} atoms an additional argument representing the position of a disjunct in the consequent of a rule; second, add the following generic weak constraint, which allows to prefer answer sets that achieve the lowest possible indexed disjunct.
\[{\small\begin{array}{l}
\texttt{\mbox{$:\sim$}cons(ID,I,Args,T,Ts).[1@I,ID,I,Args,T,Ts]}
\end{array}}\]
\noindent where the second argument of cons represents the position of the disjunct in the consequent of the reactive rule.
For instance, the head of the second ($apologise$) disjunct in Equation~(\ref{eq2}) might be written as the atom \texttt{cons(1,2,(Cust,Item,T),T,Ts)}, where $T$ represents the timestamp of the associated request.
In this case the weak constraint ensures a preference for allocating items that are requested (\texttt{I}=1), if possible, rather than apologising (\texttt{I}=2).
 
We briefly mention some other typical examples, based around the bookstore narrative from Section~\ref{eq2}, that demonstrate how  constraints might be used to enhance decisions in  Reactive ASP.
As another example of using a weak constraint, we might want to prioritise allocation of items to a particular customer (say Tom) over any others in a situation where  two customers requested the same item at the same time. This could be achieved by the constraint 
\[{\small\begin{array}{l}
\texttt{\mbox{$:\sim$}happens(request(tom,Item),T),happens(request(C,Item),T),C!=tom,}\\
\mytab \texttt{happens(allocate(tom,Item,\_),T2),happens(allocate(C,Item,\_),T1), }\\
\mytab \texttt{time(T),time(T2),time(T1),T1<T2.[1@1,T,T1,T2,Item,C]}
\end{array}}\]
\noindent which penalises allocating the Item to customer C before allocating to Tom.
Another example could be to prefer not to process more than one book  at any  time.
This could be expressed by\footnote{For simplicity, we assume customers do not request a particular item more than once in the timescale of the program.}
\[{\small\begin{array}{l}
\texttt{\mbox{$:\sim$}happens(process(C1,Item1),T),happens(process(C2,Item2),T),}\\
\mytab \texttt{time(T),Item1<Item2.[1@2,T,Item1,Item2,C1,C2]}
\end{array}}\]
One could even express a preference not to allocate items to the same customer within (say) three timesteps. Other kinds of preferences could include: allocate as early as possible, or if an item 
requested by a customer is not in stock but is on order to be re-stocked 
then schedule the response to the customer's order
as late as possible (to avoid apologising unnecessarily). One could also consider more specific preferences, such as the one above, preferring to allocate books one at a time, but within that preferring to allocate children's books as early as possible. This can be achieved by adding a second weak constraint at level 1. For instance
\[{\small\begin{array}{l}
\texttt{\mbox{$:\sim$}happens(request(C,Item),T1),child\_book(Item)},\\
\mytab\mytab\texttt{happens(allocate(C,Item,\_),T2)},\\
\mytab \mytab \texttt{time(T1),time(T2).[(T2-T1)@1,Item,T1,T2,C]}.
\end{array}}\]
\noindent will aim to minimise the time difference between a request and allocation of a children's book.

In the next subsection we will show how to achieve preemptive, proactive and prospective behaviours in Reactive ASP.


\subsection{Relaxing Reactivity to Provide a Variety of Other Models}\label{proactive}
Recall that in KELPS the fact that all actions are supported  is an emergent feature but in Reactive ASP  this has  to be formalised explicitly in the program.
Below we explain how,
by relaxing that actions be supported, 
a wider range of models can be generated by Reactive ASP
leading to either preemption of  a reactive rule, or  proactive behaviour to satisfy a reactive rule. 
For instance, in case of an alarm in some building, the following KELPS reactive rule will produce  a reactive model including the $evacuate$ action only  if no guard is present at the time of the alarm.
\begin{equation}\label{security}
\begin{array}{l}
alarm(T1) \wedge  \neg present\_guard(T1) \\
\mytab \rightarrow evacuate( T2) \wedge T1+1 < T2 \wedge T2< T1+4
\end{array}\end{equation}
\noindent However, another possible model could be to preempt the alarm and send a guard to the building even before any alarm, so evacuation would not  be needed. 
On the other hand, a proactive model might, as a precaution, evacuate a building even before any alarm! 
A more practical example of proactive behaviour could be to buy a ticket in advance of entering a bus to save time.
Thus  given the reactive rule $enter\_bus(T) \rightarrow have\_ticket(T_1) \wedge T \leq T_1\wedge T_1\leq T+1$
the ticket could be bought  before entering the bus, ensuring that the fluent  $have\_ticket$ holds when necessary.
Neither of these behaviours is possible in KELPS as they are not supported by earlier conditions in a reactive rule, but by relaxing the \texttt{supported} condition for actions both  behaviours can be achieved in our Reactive ASP formalism.

Firstly, any action that can be either preemptive or proactive is defined 
by a clause of the form  
\texttt{action(act($\bar{X}$)):-body($\bar{X}$)}
where \texttt{act}  names the action, and \texttt{body($\bar{X}$)} represents a conjunction of auxiliary atoms used to ground $\bar{X}$, the set of arguments (if any) pertaining to that action.  

\begin{figure}[hbtp]
{\small\begin{tabular}{ll}\hline
1. &\texttt{time(0..7).}\\
2. & \texttt{action(send\_guard).}\ \ \  \texttt{action(evacuate).}\\
3. & \texttt{happens(alarm,3).}\\
4. &\texttt{initiates(send\_guard,present\_guard).}\\		
5. &\texttt{ant(1,(Ts),Ts):-happens(alarm,Ts),not\ holds(present\_guard,Ts),time(Ts).}\\
6. &\texttt{cons(1,(T),T,Ts):-ant(1,(T),T), }\\
& \mytab\mytab\mytab \texttt{happens(evacuate,Ts),T+1<Ts,Ts<T+4,time(Ts).}\\
7. &\texttt{:-ant(ID,X,Ts),not consTrue(ID,X,Ts),time(Ts).}\\
8. &\texttt{consTrue(ID,X,Ts):-cons(ID,X,Ts,Ts1),time(Ts1).}\\
9. &\texttt{0\{happens(Act,Ts)\}1:-action(Act),Ts>0,time(Ts).}\\
10. &\texttt{:-happens(send\_guard,Ts),holds(present\_guard,Ts-1),time(Ts-1),time(Ts).}\\
& \textit{\% Event theory as before}
\\ \hline\end{tabular}}
\caption{Security guard proactive and preemptive behaviour.}
\label{guard}
\end{figure}

Secondly, the choice rule is simplified to
enable  actions of this kind to take place at any time $T$, whether supported or not:
{\small\texttt{0\{happens(Act,Ts)\}1:-action(Act),time(Ts),Ts>1.}}
\noindent Actions that cannot be proactive or preemptive (i.e. are not defined by an \texttt{action} clause) are enabled, as before, by use of the \texttt{supported} predicate.
As an example, in Figure~\ref{guard}  we illustrate how we can make actions \texttt{send\_guard} and \texttt{evacuate} potentially proactive and preemptive. The figure also includes the ASP version of reactive rule (\ref{security}).\footnote{If \texttt{evacuate} is not allowed to be proactive then the second fact of Line 2 would be omitted and appropriate clauses (similar to those in Lines 9-10 of Figure~\ref{singlesimplecorrected}) would be needed.}
This ASP program exhibits (or results in) several possible distinguishable behaviours via the models it generates.
In some models a guard is sent at or before time 3 (preemptive behaviour); in some  a guard is not sent and evacuation takes place at  time 5 or 6 (reactive behaviour); and in others evacuation takes place at times from time 1 onwards (proactive behaviour).

In the preemptive models the sending of the guard causes the antecedent condition {\small\texttt{not holds(present\_guard,Ts)}} to be false by causing {\small\texttt{holds(present\_guard,Ts)}} and avoiding evacuation.
To prefer such behaviour a weak constraint can be used, such as the following, which minimises the number of \texttt{evacuate} actions, the best being zero:
{\small\texttt{\mbox{$:\sim$}happens(evacuate,T),time(T).[1@2,T]}}.
By making the priority level 2, this constraint will be minimised before any at a lower priority.
 Line 10 ensures that multiple occurrences of sending a guard are avoided, as a guard is sent only if  one is not already present. Alternatively, the weak constraint 
{\small\texttt{\mbox{$:\sim$}happens(send\_guard,T),time(T).[1@1,T]}}
could be added to minimise the occurrences of the sending of guards (in addition to minimising the number of evacuates). 
This constraint can also be used on its own to minimise occurrences of sending a guard if  proactive behaviour is preferred.

\subsection{Prospective Reasoning}\label{Prospection}

KELPS agents are capable of non-deterministic self-evolution. At any given time, they may have several different possible future trajectories depending on what actions they take, when those actions are taken and what external situations arise. 
This provides an opportunity to explore what Pereira and others \cite{prospective,prospective2} call Prospection or Evolution Prospection. The challenge stated in \cite{prospective} was `how to allow such evolving
agents to be able to look ahead, prospectively, into such hypothetical futures, in order to determine the best courses of evolution from their own present, and thence to prefer amongst them'.

The advantage of  Reactive ASP compared to the original KELPS is that it naturally incorporates Prospective reasoning and no further machinery is required. In particular, Recative ASP determines
{\em every} possible course of actions 
{\em and} the corresponding set of ramifications (within a given timeframe), and thus allows $n$-distant Prospective KELPS.
 It can also easily accommodate constraints and expected future events within that timeframe.
 Thus each answer set would represent the agent's possible future evolution within  a fixed timeframe of $n$ cycles, with expected external scenarios. 
 Furthermore, we  can express `a priori' preferences for certain outcomes using strong and weak constraints, allowing to filter action plans  and to highlight others in  order of optimality. 
We will illustrate some of these features by an example.

Consider the following scenario related to decisions about what to drink 
and when to go to bed. There are three reactive rules:
(i) if the agent drinks wine they must retire (to bed) within one cycle, due to drowsiness; (ii) if the sun sets the agent must also retire, but within three cycles; and (iii) 
if the agent is thirsty, they  must have a drink  (coffee, wine or water) before three cycles. 
Furthermore, there are some action preconditions: the agent cannot perform any action while asleep, nor, if the agent feels energetic, can they go to bed.
Finally, there are some  postconditions: drinking coffee makes the agent energetic and going to bed induces sleep.
Ideally, the agent wants to go to bed as late as possible. 
The scenario is expressed in the KELPS framework as shown in Figure~\ref{prospective}.\footnote{
In order to avoid clutter  we omit the temporal constraints related to 
$n$-distance, but will assume they are present for our choice of $n$.}
Note that the above preference cannot be represented in KELPS.

\begin{figure}[hbpt]
\begin{tabular}{ll}\hline
{\textit{\Aux}:} &$isDrink(coffee)$\ \ \ \ \ \ $isDrink(wine)$\ \ \ \ \ \ $isDrink(water)$\\
{\textit{\R}:} & $drink(wine, T) \rightarrow gotoBed(T+1)$\\
& $sunset(T) \rightarrow gotoBed(T_1) \wedge T < T_1 \wedge T_1 \leq T+3$\\
& $thirsty(T) \rightarrow drink(Liquid, T_1) \wedge isDrink(Liquid) \wedge T < T_1 \wedge T_1 < T+3$\\
{\textit{C$_{post}$}:} & $initiates(drink(coffee),energetic)$\\
 & $initiates(gotoBed, asleep )$\\
{\textit{C$_{pre}$}:} & $false \leftarrow asleep(T) \wedge drink(L, T+1) \wedge  isDrink(L)$\\
& $false \leftarrow asleep(T) \wedge gotoBed(T+1)$\\
& $false \leftarrow   energetic(T) \wedge gotoBed(T+1) $\\
\\ \hline\end{tabular}
\caption{Example for prospection.}
\label{prospective}
\end{figure}

The corresponding mapping to Reactive ASP is standard and is in Figure~\ref{drinking} (in the Appendix).
Imagine now  that $S_0 = \{\}$ and $ext^*=\{thirsty(1)\}$. So the agent must drink something and could choose  coffee, wine, or water.
Suppose we also know that sunset will occur at time 2.  Then we can extend  $ext^*=\{thirsty(1), sunset(2)\}$
and look ahead beyond the first 2 cycles, say up to time 5.
Given the preference of going to bed as late as possible, the latest bedtime is time 5, after the sunset that occurs at  time 2. As soon as the agent goes to bed it falls asleep and so can no 
longer drink. Thus the drinking can happen at the latest by time 5; in fact the latest is time 3 as it must be before time 4 according to the third reactive rule.
Depending on which drink is chosen, there are various consequences.
If coffee is chosen as the drink at time 2 or 3, then due to becoming  energetic the agent will not be able to go to bed in time (and there will be no 
model). If wine is chosen at time 2 or 3, then the agent will have to go to bed at 
time 3 or 4, respectively,  because of the first reactive rule. If water is chosen at time 2 or 3 then the agent need not go to 
bed until time 5. Of course, it is also possible to have multiple drinks, with similar 
consequences as above.

Reactive ASP  provides all this information in the answer sets it produces, and the agent can then choose the best drinking option in the 
light of this. It is also possible to express preferences {\em a priori}, for example to  minimise the number of drinks,  or the aforementioned preference of going to bed as late as possible. 
These are achieved by weak constraints as in Lines 17 and 18 in Figure~\ref{drinking}, repeated here.
\[{\small\begin{array}{ll}
17\mbox{.} & \texttt{\mbox{$:\sim$}happens(drink(L),T),isDrink(L),time(T).[1@1,T,L]}\\
18\mbox{.} & \texttt{\mbox{$:\sim$}happens(gotoBed,T),time(T).[-T@2,T]}\\
\end{array}}\]

%% file: Hybrid_corr.tex
\section{An Integrated KELPS and Reactive ASP Framework}\label{sec:combined}
\begin{figure}[htbp]
\begin{center}
\vspace{-0.4cm}
 \includegraphics[width=12cm]{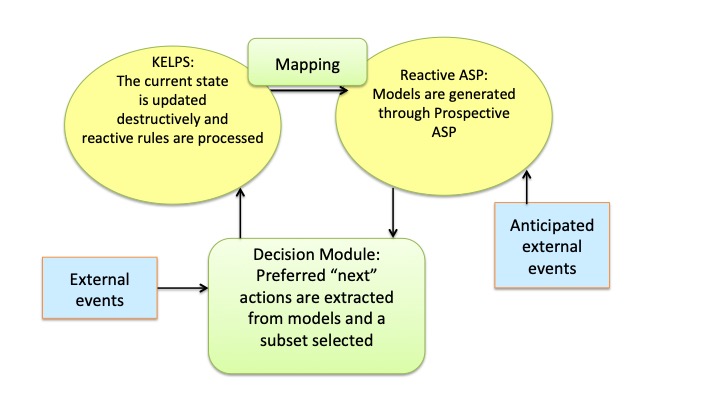}
\end{center}
\vspace{-1.0cm}
\caption{HKA: a combined architecture for reactive and prospective control.}
\label{combined}
\end{figure}

The discussions in the paper up to this point have highlighted the strengths of each of the two paradigms of Reactive ASP and KELPS. The major strength of Reactive ASP is that it easily allows a variety of reasoning behaviours and functionalities, such as prospective reasoning and reasoning with preferences via weak constraints. On the other hand major strengths of the KELPS operational semantics are that it updates the state
destructively and incrementally simplifies the reactive rules. Thus it does not reason with frame axioms, nor does it need to access past information about states or events.
These respective strengths suggest a potential new architecture for reactive, prospective agents that combines  KELPS and Reactive ASP. Such an architecture is summarised  in Figure~\ref{combined}. The distribution of the work between the two paradigms in this architecture   is informed by their relative strengths. We describe the architecture, henceforth abbreviated to HKA, and illustrate its behaviour through two examples.

In HKA the KELPS operational semantics retains from Section~\ref{sec:background}  the parts of updating the state and triggering and simplifying the reactive rules in the light of changes to states and events, whereas the part of generation of plans to solve the goals is passed on to (prospective) Reactive ASP.

That is, in the KELPS module, at time $T$, the state is updated according to the events executed during the interval 
$[T-1,T)$. Then the reactive rules are processed given the updated state at time $T$.
The KELPS module then passes  to Reactive ASP the  updated state and all  the processed reactive rules.
Furthermore, if there happen to be anticipated external events in the future, these are also input to the Reactive ASP module.
This module will have a timeframe starting at $T$ and ending at $T+k$ (\texttt{time(T..T+k)}), for some chosen k, meaning that the search can consider up to  $k$-distant models beyond the current time $T$. The value of $k$ can  vary in different iterations of the cycle if required.

The current state is modelled by {\small \texttt{holds(Args,T)}} and the future anticipated events at some time \texttt{t>0} as {\small\texttt{happens(Args,t):-time(t)}}.
Since KELPS has processed all previous occurred events, there is no need for ASP to start its time range from earlier than \texttt{T}. 
We can illustrate this with a simple example.
Let  \[a(T) \wedge b(T+3) \rightarrow c(T1) \wedge T+3<T1 \wedge T1<T+6\] be a reactive rule in KELPS, where $a$, $b$ and $c$ are events, and suppose that  $a$ occurs at time 2.  KELPS will partially process this rule to give a rule $b(5) \rightarrow c(T1) \wedge 5<T1 \wedge T1<8$.
It is this rule that is mapped into Reactive ASP for the time interval starting at time 2.

The rest of  the Reactive ASP program, such as the causal theory and event theory remain as described in Section~\ref{sec:translation}. Weak constraints formalising priorities and preferences can also be changed if desired with different iterations of the cycle.

The output from running the Reactive ASP part of HKA  will be a set of optimal answer sets, according to the weak constraints, and looking forward in time from $T$ to $T+k$.
From these answer sets 
and taking account of the external events in the next time interval [T,T+1), the best set of actions in this time interval to execute can be chosen. These are executed and combined with the external events and fed into the KELPS module for the next iteration at time $T+1$.

Note that in  HKA  the ASP program has no need to reason about the past. In particular it will never need to instantiate the frame axiom in the event theory with time prior to the current cycle time,  i.e. prior to $T$ in its timeframe of \texttt{time(T..T+k)}. It also does not need to reason with past events and states in respect to the reactive rules.
We consider some simple examples of how this interaction  works. For the first example, shown  in Figure~\ref{Both1}, we just describe  the results.
\begin{figure}[hbpt]
\begin{tabular}{ll}\hline
{\textit{R1}:} & $a1(T) \rightarrow (a3(T1) \wedge T<T1<T+4) \vee (a4(T1) \wedge T<T1<T+4)$\\
{\textit{R2}:} & $a2(T) \rightarrow (a6(T1) \wedge T<T1<T+9) \vee (q(T1) \wedge a7(T1) \wedge T<T1<T+9)$\\
{\textit{C$_{post}$}:} & $\mathit{terminates}(a3,p)$\ \ \ \ \ $\mathit{initiates}(a8,q)$\\ 
{\textit{C$_{pre}$}:} & $false \leftarrow a7(T+1) \wedge \neg p(T)$\\
Initial: & $p(0)$\\
Events: & $a1(1)$\ \ \ \ \ $a2(1)$\ \ \ \ $a8(6)$
\\ \hline\end{tabular}
\caption{First Example for HKA interaction ($a1 - a8$ are events and $p$ and $q$ are fluents).}
\label{Both1}
\end{figure}

Informally, from Figure~\ref{Both1} it can be seen that to satisfy rule $R1$ either $a3$ or $a4$ can occur at any time in the range [$2,\ldots,4$], and to satisfy rule $R2$ either $a6$ may occur at any time in the range [$2,\ldots,9$], or $a7$ can occur, but at or after time 6, after $q$ is initiated by event $a8$.   
Suppose now that it is also the case that $a7$ is preferred to $a6$ in order to satisfy rule $R2$.  It can be seen that in this situation rule $R1$ can only be satisfied by $a4$, because $a3$ terminates the precondition $p$ of $a7$ before $a7$ can be usefully executed (because $q$ would not be true). But after $a4$, $a7$ can occur at times   6, 7, 8 or 9.
When translated into ASP at time 1 and run prospectively up to time 10, ASP will return as optimal answer sets the above results.
The optimal answer sets are returned to KELPS, and KELPS makes a choice about the next action. It is worth noting that not only can KELPS by itself not deal with preferences, it cannot deal with foreknowledge of events such as $a8$ in the example when reasoning at earlier times.

Suppose an additional rule $R3: a7(T) \rightarrow a9(T1)\wedge T1=T+1$  is added to the example of Figure~\ref{Both1}, and  that it is also desired for  $a9$ to be executed as late as possible. Then in a timeframe extending from 1 to 10, $a4$ can occur as before and $a7$ should be executed at time 9, the latest time it can be to 
satisfy rule $R2$, to allow $a9$ to occur at time 10 and rule $R3$ to be satisfied.

In the next example, shown in Figure~\ref{Both3}, we illustrate the  evolution of the ASP translation through several cycles. 
 By way of illustration we assume that KELPS requests ASP to consider a timeframe of 3 steps beyond the current time (i.e. $k$=3). 
  \begin{figure}[hbpt]
\begin{tabular}{ll}\hline
{\textit{R}:} & $a(T) \rightarrow a1(T1)  \wedge T<T1 \wedge  a2(T2) \wedge T2=T1+1$\\
{\textit{C$_{pre}$}:} & $false \leftarrow a2(T) \wedge a3(T)$\\
Events: & $a(1)$
\\ \hline \end{tabular}
\caption{Second Example for HKA interaction (here $a$ and $a1-a3$ are events).}
\label{Both3}
\end{figure}

\begin{figure}[hbtp]
{\small\begin{tabular}{ll}\hline
& \textit{At time 0 (initial state) before any actions}\\
0.1 &\texttt{time(0..3).}\\
0.2 &\texttt{ant(1,(Ts),Ts):-happens(a,Ts),time(Ts).}\\
0.3 &\texttt{cons(1,(T),T,Ts):-ant(1,(T),T),happens(a1,T1),T<T1,}\\
 & \mytab\mytab \texttt{happens(a2,Ts),Ts=T1+1,time(T1),time(Ts).}\\
0.4 & \texttt{supported(a1,Ts):-ant(1,(T),T),T<Ts,time(Ts),time(Ts+1).}\\
0.5 & \texttt{supported(a2,Ts):-ant(1,(T),T),happens(a1,T1),}\\
  &  \mytab\mytab \texttt{T<T1,Ts=T1+1,time(T1),time(Ts).}\\
0.6 & \texttt{:-happens(a2,Ts),happens(a3,Ts),time(Ts).}\\
0.7 & \texttt{0\{happens(Act,Ts)\}1:-supported(Act,Ts),0<Ts,time(Ts).}\\
0.8 & \texttt{:-ant(ID,X,Ts),not consTrue(ID,X,Ts),time(Ts).}\\
0.9 & \texttt{consTrue(ID,X,Ts):-cons(ID,X,Ts,Ts1),time(Ts1).}\\
\\
& \textit{After \texttt{happens(a,1)} and KELPS processes rule $R$, Reactive ASP}\\
& \ \ \ \textit{ will include all rules from time 0 except 0.1 and 0.7 together with the following:}\\
1.1 &\texttt{time(1..4).}\\
1.3 &\texttt{cons(1,(1),1,Ts):-happens(a1,T1),1<T1,happens(a2,Ts),}\\
 & \mytab\mytab \texttt{Ts=T1+1,time(T1),time(Ts).}\\
1.4 & \texttt{supported(a1,Ts):-1<Ts,time(Ts),time(Ts+1).}\\
1.5 & \texttt{supported(a2,Ts):-happens(a1,T1),1<T1,Ts=T1+1,time(T1),time(Ts).}\\
1.7 & \texttt{0\{happens(Act,Ts)\}1:-supported(Act,Ts),1<Ts,time(Ts).}\\
1.8 &   \texttt{:- not consTrue(1,(1),1).}\\
\\
&\textit{At time 2 after \texttt{happens(a1,2)} and knowing about \texttt{happens(a3,3)}, Reactive ASP}\\ 
&\textit{will have all rules from time 1 except 1.1 and 1.7, together with the following:}\\
2.1 &\texttt{time(2..5).}\\
2.3 &\texttt{cons(1,(1),1,3):-happens(a2,3).}\\
2.5 & \texttt{supported(a2,3).}\\
2.7 & \texttt{0\{happens(Act,Ts)\}1:-supported(Act,Ts),2<Ts,time(Ts).}\\
2.10 & \texttt{happens(a3,3):-time(3).}\\
\\
&\textit{At time 3 after \texttt{happens(a3,3)} and \texttt{happens(a1,3)}, Reactive ASP will have}\\
& \ \ \ \ \ \textit{all rules from time 2 except 2.1, 2.5, 2.7 and 2.10, together with the following:}\\
3.1 &\texttt{time(3..6).}\\
3.3 &\texttt{cons(1,(1),1,4):-happens(a2,4).}\\
3.5 & \texttt{supported(a2,4).}\\
3.7 & \texttt{0\{happens(Act,Ts)\}1:-supported(Act,Ts),3<Ts,time(Ts).}\\
\\
& \textit{Other parts as standard}
\\ \hline\end{tabular}}
\caption{In the numbering notation $x$.$y$, $x$    is the timestamp of the start of the timeframe and $y$ is an identifier for  a reactive ASP rule. Thus rules with $x=0$ are the mapping of $KELPS(0;3)$ into Reactive ASP. Rules  with $x>0$ are the mapping of $KELPS(x;x+3)$.}
\label{Both2}
\end{figure}

The prospective answer set from ASP in the timeframe $\{1,\dots,4\}$ will suggest  to execute $a1$ at time 2
or at time 3, allowing for $a2$ to be executed at time 3 or at time 4, respectively.
Suppose that indeed  $a1(2)$ is executed,  and furthermore foreknowledge of the external event $a3(3)$  becomes available. Then it will become impossible to execute $a2(3)$ because it would violate \textit{C$_{pre}$}. 
The only course of action to satisfy the reactive rule is for $a1$  to be executed again.\footnote{Note this is possible because $a1$ in the consequence of the rule remains supported, and ASP can still generate it through the choice rule.}   
Assuming that the timeframe has now increased  to $\{2,\dots,5\}$, then ASP will recommend re-executing $a1$ at time 3 (or time  4) to allow for $a2$ to be executed at time 4 (or time 5) in order to satisfy the reactive rule.

The operation of HKA relies on the following key observation:
at each step of KELPS processing the reactive rules, the causal theory and the current state form a KELPS   framework. 
In effect, the current state $S_{T}$ becomes the initial state  for the next cycle (in Figure~\ref{combined}) and the conversion to $k$-distant  KELPS and to Reactive ASP can be carried out as described in Section~\ref{sec:translation}. 
In the sequel we will use the notation $KELPS(T;T+k)$ to refer to the $k$-distant KELPS starting at time $T$.
In Figure~\ref{Both2} we show the ASP rules resulting from mapping $KELPS(T;T+3)$ for Figure~\ref{Both3} at times $T = 0, 1, 2, 3$.
In this  figure the numbering notation $x$.$y$ indicates that $x$    is the timestamp of the start of the timeframe and $y$ is an identifier for  a reactive ASP rule. Thus rules with $x=0$ are the mapping of $KELPS(0;3)$ into Reactive ASP. Rules  with $x>0$ are the mapping of $KELPS(x;x+3)$. 
For example, at  time 2  rule 2.3 is the result of processing the consequent of $R$ due to the event $a1(2)$, the antecedent of $R$ having already become true because of the earlier event $a(1)$.
Original rules from previous timestamps   are carried forward, unless otherwise indicated.

By way of further illustration we highlight some of the results of these mappings. 
The rules 0.1 to 0.9 (called set 0) are a mapping of the  framework $KELPS(0;3)$ from Figure~\ref{Both3}.
These rules, except  0.1 and 0.7,  are maintained in subsequent frameworks.
For example the constraints 0.8 and 0.6 are present in set 2, that maps $KELPS(2;5)$.
The mapping of framework $KELPS(1;4)$ (set 1) includes in addition the rules 1.1, 1.3-1.5, 1.7 and 1.8.
The rules for \texttt{time} (e.g. rules 1.1) replace the old ones (e.g. 0.1) because of the shifting of the time window, that is  rule \texttt{time(0..k)} is replaced by rule  \texttt{time(1..1+k)}.
The new  choice rule 1.7 replaces the old one (0.7)  for the same reason, increasing the bound on $Ts$ to the new start time of the framework.
Similarly for transitions between times 1 and 2, etc.
Rule 1.3 is the mapping of the  processing by KELPS of rule $R$ after the event $a(1)$, which makes the antecedent of the rule true at time 1, leaving the instantiated consequent  $a1(T1)  \wedge 1<T1 \wedge  a2(T2) \wedge T2=T1+1$.
The occurrence of event $a(1)$ also results in the new  \texttt{supported} rules 1.4 and 1.5.
Since the antecedent of the reactive rule has been made true at time 1 the reactive rule constraint for that instance of the rule becomes the constraint 1.8  \texttt{:-not consTrue(1,(1),1)} indicating that rule $R$ has not yet been satisfied. 
Similar explanations for the rules in set 2 and set 3 can be given.
The rules in set 3 are the mapping of $KELPS(3;6)$, after action $a1(3)$ and event $a3(3)$.
Notice also, that although some rule may have  been satisfied, while actions in its consequent remain supported, ASP will still be able to generate those actions through the choice rule.
This is why action $a1(3)$ can occur as it is still supported (see rule 1.4).

We end this section by discussing the relationship between HKA and KELPS. First we define some terminology to use in our discussion.

\begin{definition}\label{hybridTranslation}
Let $T\geq 0$ be a time and  $KELPS(T;T+k)$ = $<R_{T},\C,\Aux,k>$ be a $k$-distant KELPS   framework  starting at time $T$ with state $S_{T}$. 
Let  $ev^{*}_{T+1}$ be  events that take place during the time interval $[T,T+1)$. 
Let these events transform the KELPS state $S_{T}$ to $S_{T+1}$ and process the reactive rules  $R_{T}$ to $R_{T+1}$.
We denote by $KELPS(T+1;T+1+k)$ = $<R_{T+1},\C,\Aux,k+1>$, the $k$-distant KELPS   framework  starting at time $T+1$ with state $S_{T+1}$. 
We denote the mapping of $KELPS(T;T+k)$ into ASP by $ASP(T;T+k)$ and the mapping ot $KELPS(T+1;T+1+k)$ into ASP by $ASP(T+1;T+1+k)$.

\end{definition}

\begin{figure}[h]
\begin{center}
\includegraphics[height=3.5cm,width=12cm]{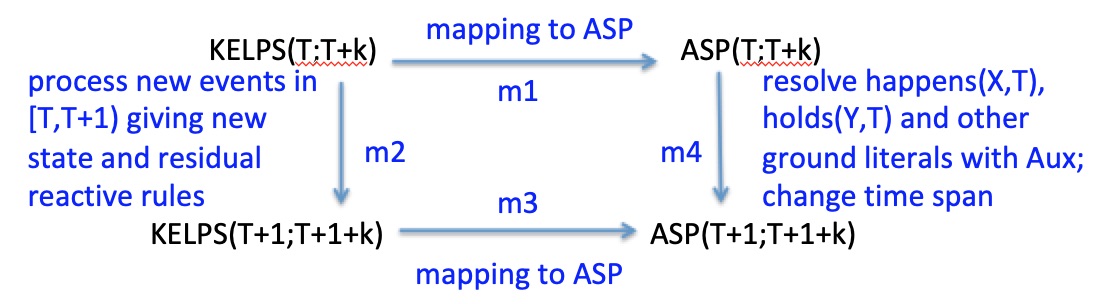}
\end{center}
\vspace{-0.4cm}
\caption{Relationship between KELPS($T;T+k$) and ASP($T;T+k$).}
\label{commute}
\end{figure}

We focus the discussion of the relationship between HKA and KELPS on the commutative diagram shown in Figure~\ref{commute}.
In that figure the HKA approach we have described so far involves the mapping of $KELPS(T;T+k)$ to $ASP(T;T+k)$ (arrow {\em m1}), processing of the rules in $KELPS(T;T+k)$ in the light of events that take place during the time interval $[T;T+1)$ and the resulting updated state to obtain $KELPS(T+1;T+1+k)$ (arrow {\em m2}), and a mapping of this to $ASP(T+1;T+1+k)$ (arrow {\em m3}). In addition, of course, as indicated in Figure~\ref{combined},  $ASP(T;T+k)$ feeds back to $KELPS(T;T+k)$ information about computed answer sets, as does $ASP(T+1;T+1+k)$ to $KELPS(T+1;T+1+k)$.

It is worth noting that the soundness and completeness results of Section~\ref{sec:theory} naturally extend to the mappings in {\em m1} and {\em m3}, based on the aforementioned key observation that at each step of KELPS processing the reactive rules, the causal theory and the current state form a KELPS framework.
But as can be seen in Figure~\ref{commute} there is an alternative, more economical, approach to realising HKA, namely the following. There is an initial mapping of $KELPS(0;k)$ to $ASP(0;k)$, but after that, at each time $T>0$ after KELPS provides the executed events and the updates state to ASP, it is possible to mirror in ASP itself the (KELPS-style) processing of the reactive rules to obtain the new $ASP(T;T+k)$. This is indicated by arrow {\em m4} in Figure~\ref{commute}. This avoids the need for mappings of KELPS into ASP after $T>0$.
In the following we define the direct mapping between $ASP(T;T+k)$ and $ASP(T+1;T+1+k)$.
This correspondence between KELPS and its mapping to ASP relies on the fact KELPS processing of the reactive rules is based on resolution. Moreover, since both ASP and KELPS are logic-based languages   the resolution step is defined identically for the two, and thus each mirrors the other. We formalise this idea next in Definition~\ref{directASPmapping}.

\begin{definition}\label{directASPmapping}
Given $ASP(T;T+k)$, a set of events in the form \texttt{happens(e, T+1)}, in the time interval $[T,T+1)$ and a set of fluents, in the form \texttt{holds(f,T+1)}, $ASP( T+1;T+1+k)$ is constructed through 5 steps as follows: 
\begin{description}
\item[Step 1: Increment time.] Replace time facts \texttt{time(T..T+k)} with \texttt{time(T+1..T+1+k)} and replace the condition $T<Ts$ in the choice rule with $T+1<Ts$.
\item[Step 2: Reason with events and fluents at time $T+1$.] Apply inference to rules in $ASP(T;T+k)$ using facts of the form \texttt{happens(e,T+1)} and \texttt{holds(f,T+1)}.
\item[Step 3: Simplify using $\Aux$ and \texttt{time} facts.] Remove from rule bodies all ground and true \texttt{time} atoms (i.e. inference with the \texttt{time} facts added in Step 1). 
Remove all true negated \texttt{happens} and \texttt{holds} literals, and negated $\Aux$ literals.
\item[Step 4: Propagate new facts timestamped $T+1$.] Apply inference to rules using any newly generated facts timestamped $T+1$ (e.g. \texttt{ant} and \texttt{cons}) with rules.
\item[Step 5: Remove processed facts and constraints.] Except for any newly generated \texttt{consTrue} fact, remove facts timestamped $T+1$ which were reasoned with in Step 4. Remove rules which now have unsatisfiable   body conditions. This latter step may apply to ground constraints of the form \texttt{:-not consTrue(ID,Args,Ts)} which were generated at a  time $Ts$ earlier than the current time $T+1$. 
\end{description}
\end{definition}

We use Figure~\ref{Both2} to exemplify some of the reasoning described in Definition~\ref{directASPmapping}.
The rules 1.3 to 1.5 and 1.8 are   instances of rules in set 0 
that have been derived using facts at time 1.
Rule 0.2 is resolved with the facts \texttt{happens(a,1)} and \texttt{time(1)} (Steps 2 and 3) to give
\texttt{ant(1,(1),1)}, which in turn is resolved with rule 0.3 to give 1.3, and with rule 0.8 to give 1.8 (Step 4). Similar derivations yield the other rules. 
If the constraint 0.6 were not present, action $a2$ could occur at time 3 making \texttt{cons(1,(1),1,3)} true. This would allow \texttt{consTrue(1,(1),1)} to be derived from 0.9 (Step 4) and the constraint 1.8 to be satisfied, whence it can be removed because it would have an unsatisfiable body condition (Step 5).
In the case of the actual example in  Figure~\ref{Both2}, since 0.6 is present, this cannot occur and constraint 1.8 remains in set 3  as shown.

%% file: Related_Final_corr.tex
\section{Discussion and Related Work}\label{sec:related}
In this section we review the translation of KELPS to ASP and the variations considered, before discussing related work.
 
\subsection{Brief Review of Reactive  ASP}\label{sec:compare}
The basic concepts of KELPS were described in Section~\ref{sec:KELPS}, 
in which, to progress towards achieving the goals specified by the  reactive rules, and in the light of new observations, at each timepoint  supported actions can be selected for execution provided their 
preconditions hold.
KELPS models are not time limited, so for translating to ASP, which produces finite models, the notion of an $n$-distant KELPS framework was introduced (see Definition~\ref{ndistantframework}). The basic restriction is that time is limited to the interval 
$[0, \ldots, n]$ 
for some given value of $n$ and no events can occur after $n$, nor can the notion of supportedness require that time extend beyond $n$.

The translation of $n$-distant KELPS into ASP 
was described in subsections~\ref{basicMapping} to \ref{supportedMapping}, and proven sound and complete in Section~\ref{sec:theory}.
A feature of the translation is the ability to include preemptive and proactive, as well as reactive, behaviour.
This is facilitated by the flexibility of how to  specify explicitly in the program what actions can be selected, as well  as by prioritising models with weak constraints, as we saw in Section~\ref{sec:extensions}.
Another feature of the translation we illustrated in Section~\ref{sec:extensions} is the 
ability to include prospective behaviour.
This is essentially the basic mapping, but the focus is different. 
In this case
foreknowledge of possible future events is made available to improve the decision making at the current time. This feature is not possible in KELPS, but simply emerges as a  consequence of the ASP translation. 

The two paradigms KELPS and Reactive ASP differ in their  \textit{operational behaviour}. In KELPS the framework reasons partially about the consequences of reactive rules and interleaves the reasoning with action execution. Thus a KELPS framework may execute an action as part of a partial evaluation of a consequent, and simply verify that 
 the temporal constraints will   allow a model to exist potentially.  Of course, future (as yet unknown) events, or consequences of future actions, might still prevent the existence of such a potential model.
On the other hand, Reactive ASP, being a model generation paradigm, generates complete answer sets and thus complete plans up to a maximum time  $n$, and it is this that makes prospective reasoning possible. Even so, it is possible that an answer set of a reactive ASP program  exists at some time $t$, but no answer set including it exists for longer timeframes because of new observations made after  time $t$.  
Another major difference between the two paradigms is that KELPS  requires no frame reasoning, but Reactive ASP requires reasoning with the  explicit frame axiom in the event theory (rule 14 in Figure~\ref{singlesimplecorrected}).

These differences encouraged the design  of a hybrid KELPS/ASP paradigm
that benefits from  the destructive update of actions and the consequences on fluents that is inherent in KELPS, yet also uses prospection as a way of providing information about   the `best' actions to be selected for execution at each increasing time point. 
In an architecture described in Section~\ref{sec:combined} for this hybrid
the initial KELPS program is translated into ASP, and then at each later 
timestep ASP recommends the next set of actions to be executed through 
prospective reasoning taking into account any anticipated future events 
up to a required timepoint, and KELPS executes the actions and returns 
to ASP via the translation the updated state and partially evaluated 
reactive rules. The period of prospection can vary as required, for 
example at each timestep.
This hybrid system is able to simulate an extended KELPS 
framework incorporating prospective reasoning over some possibly varying set $k$ of future times, 
the prospective timeframe being 
$[currentTime \dots  currentTime+k ]$, where $currentTime$ increases by 1 at each iteration. 
This system is quite close to the operational semanitics of  
KELPS in the sense of committing to actions as time progresses.
In fact,  the models computed by the ASP program at cycle $t$ can be characterised as follows.
They are the KELPS $t+k$-distant models of the initial state of KELPS together with the executed events (observed events and user actions) up to time $t$ and the anticipated external events at times $t+1$ up to $t+k$.

\subsection{A Possible Alternative Incremental ASP Mapping}\label{sec:inc}

The operational semantics of KELPS, as depicted in Figure~\ref{fig1}, is inherently incremental in nature. As time progresses the database is updated destructively,
new events are observed and actions are considered to make the consequents of reactive rules true, once their antecedents are satisfied. 
In effect, KELPS (resp. $n$-distant KELPS) attempts to construct model structures (resp. $n$-distant model structures) based on current knowledge, for each time step $k$, $k \geq 0$ (resp. $0 \leq k\leq n$).  

We initially had a concern that in the basic translation of Section~\ref{sec:translation}, here called  the {\em standard} mapping,  the frame axiom in the event theory of Reactive ASP might render the system unscalable, as regards the size of the grounding of the program, and the time taken to generate the grounding and find solutions.

Therefore, in addition to what we have described so far we implemented a slightly different mapping  of $n$-distant KELPS to ASP utilising the incremental variant of $clingo$~4. The models returned by the implementation  are found incrementally, from inital time 0 up to final time $n$. Assuming the same external events, the effect is the same as if the standard mapped program were run iteratively in a loop for times varying from 0 through $n$. That is, for each time $t$  in the range $[0, \ldots, n ]$   there is an answer set  if and only if KELPS has a $t$-distant model, and the  answer sets correspond to the KELPS models in the sense of Section~\ref{sec:theory}. Recall from Section~\ref{sec:translation} that the sets of models of $n$-distant frameworks for different values of $n$ are generally not the same, even when the external events are unchanged. Therefore,  in the incremental approach the models must be, and are,  re-computed for each $t$ in the range $[0, \ldots, n ]$.

To compare the two approaches, we ran two simple experiments. 
The first experiment recorded the rate of time increase for a program resulting from the standard mapping, with external events, but with no reactive rule or reactive rule integrity constraint, and so the only reasoning involved the causal theory, including its frame axiom. The experiment was made for increasing numbers of fluents and/or steps. The second experiment compared the overall time for running  individual calls to the standard mapped program for $n$ varying from 0 up to a given maximum with that for running the incremental mapped program for the same given maximum. The results are shown in Tables~\ref{Tab:Expt1} and \ref{Tab:Expt2} in the Appendix.
It is clear from the results  of the first experiment that the overhead of using the frame axiom   is linear in the value of the maximum timestamp and linear in the number of fluents, which is as expected,  but acceptable.  The results of the second experiment show  that there is no gain in using the incremental mapping. 

Note also that  the Hybrid HKA framework would not require long timeframes in each run of the ASP part, and moreover it would reduce the need for reasoning with the frame axiom by limiting it only to the future events. 
Another motivation for considering the incremental  approach was that it could allow  for external events to be added by the user at each time step at the time they happen. However, the Hybrid HKA achieves this in a more flexible way, and moreover incorporates prospective reasoning as well.
Therefore, although the use of incremental clingo seemed an obvious mode for the translation,   we did not find any particular advantage, and did not pursue  it further, but for completeness we have described it in the Appendix.

\subsection{Other Approaches to Reactivity in Logic Programming}
 Reactivity, in the context of KELPS, has a specific meaning (see Section~\ref{reactivity}). 
In \cite{LPSReactive} (Section 7) and \cite{kelps_paper} (Section 6)
KELPS has been compared extensively with related work, such as abductive logic programming, event calculus, MetateM, constraint handling rules, production systems, transaction logic, active databases, agent languages and reactive systems programming languages.

 Reactivity has also been explored in Action Logics.
The paper \cite{ActionTheories} describes reactive control theories, where each control rule has on the left hand side (analogous to our antecedent) a conjunction of fluents all referring to the same time, and on the right hand side (analogous to our consequent) a conjunction of actions all to be performed at the same time. The language is quite restrictive in comparison to KELPS, and its purpose is to show the correctness of such reactive control with respect to causal theories of action. 
Reactivity has been incorporated in two extensions of the Action Language ${\cal A}$, which added triggers, initially to give a language   ${\cal A}_{o}^{T}$ \cite{BaralKR}, and then further extended to a language ${\cal A}_{\infty}^{T}$ \cite{Triggers}. The triggers in ${\cal A}_{o}^{T}$ have the restricted syntax of fluents holding in a single state triggering an action that must take place at the time it is triggered. ${\cal A}_{\infty}^{T}$ provides more flexibility than ${\cal A}_{o}^{T}$ by allowing the triggered action to take place at the same time it is triggered or later. 
Separately from, and independently of, the reactive rules, ${\cal A}_{\infty}^{T}$ allows for event orderings. However, in KELPS this is incorporated in the consequents of reactive rules making the event orderings related to the context of the triggers and reactions. In addition KELPS allows histories of states and events in antecedents and consequents of reactive rules.
One similar aspect between ${\cal A}_{\infty}^{T}$ and KELPS is that the definition of state transitions in ${\cal A}_{\infty}^{T}$ captures the intuition that if an action occurs then it must have been triggered. This intuition is similar to the notion of supportedness of actions in KELPS.
 
In the work described in this paper we  use ASP to generate KELPS reactive
models, and to our knowledge there is no other work that incorporates
such reactivity in ASP. But there is other work that uses ASP to
incorporate  different perspectives of reactivity. We briefly review these below.

The oclingo system \cite{oclingo}
was an early
`reactive' answer set solver, whose technology has been
incorporated into \textit{clingo} 4. This  solver generates answer
sets incrementally, taking into account new information in
real-time.
It is reactive to the extent
that the answer sets grow and change, as new information is
acquired. \cite{maritime} suggest using this technology as a
`situational awareness' tool in maritime traffic domains. This
allows new information to be acquired (and inferred) over time,
and thus it allows managing the histories that form dynamically
over several time-steps.

In \cite{ribeiro} the authors also propose using ASP for reactive reasoning.
However, their notion of reactivity is about the efficiency of
reasoning and generation of models. They achieve this efficiency by
dividing the agent knowledge into modules. Only relevant modules
are solved at any given step, avoiding the computational cost of
generating irrelevant knowledge. In this sense the agent can then
`react' more quickly to its given situation. The ASP modules form
a tree-like hierarchy, starting at a `root' and spreading to
`leaves'. The root and the internal nodes represent the agent's
knowledge about itself (meta-knowledge), and the leaves represent
`elementary knowledge' (such as what actions should be taken).
At every step, the reasoning process starts at the root, which
determines the next module(s) to solve; this process continues
until leaf nodes are reached. The meta-reasoning determines what
the agent needs to know in the given context. It is not necessary
to reason about all available knowledge at every step.

In \cite{Costantini2011} the author uses ASP in conjunction with Observe-Think-Act agents.
For each possible external event,
a `reactive ASP module' defines the different ways in which the agent
might react to it. Each module contains
a constraint, satisfied only when the relevant external event occurs.
When the relevant event occurs, the
resulting set of models indicates possible reaction strategies.
%
In this work the  ASP modules can be triggered
only by single events. By contrast, in KELPS, and consequently in Reactive ASP, a reactive rule can
be triggered by a conjunction of events and conditions. In fact,  in KELPS and in Reactive ASP a
reactive rule is in general triggered by a history of events and
state conditions, and the reactions, in general, are plans
extending over periods of time. 
In Reactive ASP,  moreover, using weak constraints 
and prospection we
can compare different models to determine which action
plan has the most desirable future ramifications. 

In \cite{ConstantiniExploration} the authors describe a combined  architecture between DALI  agents \cite{DALI} and an answer set program that   performs planning tasks and determines what actions the agent can take.
However,
these modules do not indicate which action will lead to the best
outcome, nor do they provide the reactive or prospective reasoning of the KELPS-Reactive ASP hybrid.  

In a more recent position paper \cite{ConstantiniSurvey}   the authors argue for a modular design also exploiting features of ASP in combination with agent architectures. No specific design is given, but in principle our hybrid architecture can be said to be motivated by similar considerations. 
In fact, the authors seem to be of the opinion  that ASP is not a `fully appropriate modelling tool for 
the dynamic flexible functioning of agents as concerns reactivity, proactivity and communication'.

In Section~\ref{sec:extensions} we discussed how our formalisation of reactivity
in ASP lends itself to prospection. We saw that no further
notation or functionality was required to cater for this.
The use of \textit{clingo} and the approach to modelling KELPS
agents allows us to look into the future for a given timeframe
and also to take into account not only the present information and
its ramifications, but also any known information about the future
and its ramifications.

A logic programming system called ACORDA for prospective
reasoning is proposed in \cite{prospective} and applied to modelling multi-agent intention recognition  in \cite{prospective2} and morality in \cite{prospective3}. In the latter the authors formalise classic trolley problems, where action or inaction in diverting a trolley has implications in terms of lives saved. 
In their system abductive hypotheses are generated for
solving goals. Then some reasoning is performed from the
abducibles to obtain relevant consequences, which can then be
compared according to specified preferences. This work does not
use temporal information and does not specify how far into the
future one wishes to prospect. Moreover, it does not incorporate
knowledge about known future events or states. Another significant
difference with our work is that, as well as exploring the
consequences of  abductions\footnote{In our framework these are potential actions that are considered for execution.}, we explore their other ramifications
in terms of triggering a chain of new reactions in the future, via
the reactive rules and changes of state.

Specific applications of reactivity that KELPS/LPS can be and has  been considered for are formalisation and monitoring of policies as well as active updates in databases. It would be interesting to consider how Reactive ASP might be exploited for such applications, for example as described in \cite{Eiter2004}. 
Moreover, KELPS/LPS provides an aspect of stream processing, as described in Section~\ref{sec:Framework}. The KELPS OS monitors the stream of states and incoming and self-generated events and actions, to check whether an instance of a reactive rule antecedent has become true and to determine if any actions have become necessary. It processes the stream on the fly, i.e. as it receives it. It would be interesting to see how Reactive ASP might be exploited to do stream reasoning, for example as in \cite{Streams}.

A different formalism using equilibrium temporal logic to model reactive rules is described in~\cite{temporal}.
A recent system called {\em tclingo}, presented   in \cite{temporal} and based on temporal Equilibrium Logic \cite{TEL}, has some similarities to KELPS. Specifically, the temporal rules are restricted as in KELPS such that  antecedent atoms contain no future operators and  consequent atoms contain no past operators. However, the logic is propositional and thus appears to be less expressive than KELPS.

\section{Conclusion and Future Work}\label{sec:conclusion}
This paper has described how the form of reactivity captured within the KELPS system \cite{kelps_paper}  can be implemented in ASP. Moreover, it has shown that the resulting ASP representation is, in some respects, more flexible than KELPS in that it allows for proactive and preemptive behaviour and prospective reasoning. In addition, a Hybrid KELPS/ASP  that combines the advantages of both frameworks was described.

Although the mapping to ASP required to include explicitly a frame axiom, this has not caused particular runtime problems.
We acknowledge that in case of simulations with a very large number of constants and/or a very long time span, the running time or grounding of the answer set program could become a problem. 
However, our intention is not to deal with long timespans. For example in the proposed  Hybrid the  timespan would relate to the prospective future.
Moreover, one of our motivations to model KELPS in ASP was to facilitate the implementation of  various analysis tools by exploiting the representation, for instance to detect inconsistencies in a Reactive ASP program. This too does not need a long timestamp.
Such analysis tools would be useful because the presence of temporal constraints and action preconditions in KELPS makes it difficult to judge a priori if the reactive rules are satisfiable, and there is a need for an automated system that makes this decision. 
For example, the reactive rule constraint 
could be modified by adding a head atom 
as in 
\[{\small\begin{array}{ll}
\texttt{badRule(ID,Args,Ts):-ant(ID,Args,Ts),\mbox{not }consTrue(ID,Args,Ts),time(Ts).}
\end{array}}\]
to detect 
cases where an antecedent of some reactive rule is true but the corresponding consequent cannot be satisfied. Minimising occurrences of \texttt{badRule} atoms would show if this kind of inconsistency can happen and under what circumstances.

We also will extend the implementation of Reactive ASP to include the full LPS, including conditional clauses in the causal theory
and complex events. Recent work, such as~\cite{mehul}, has addressed the problem of detecting complex events from video. Such approaches could contribute to a longer term goal, namely a more complex reasoning system encompassing detection of external events as well as reasoning about the ramifications of the events.
Reactive ASP will also allow us to have more expressive clauses in the causal theory, whereby for example preconditions of actions can refer to histories of past events and states.
Of interest also, is  the potential to learn reactive rules due to the systematic structure of the resulting ASP program.  In our future work we will use the state-of-the-art inductive learning system ILASP \cite{ILASP} to learn reactive rules given example models of the expected behaviour.

%% file: Iclingo_appendix_corr.tex
\newpage
\section{Prospective Reasoning Example}
Figure~\ref{drinking} shows the mapping to ASP of  the example in Figure~\ref{prospective} in Section~\ref{Prospection}. Line 3 corresponds  to $ext^*$ and lines 17 and 18 represent the preferences mentioned in Section~\ref{Prospection}.

\begin{figure}[h]
{\small
\begin{tabular}{ll}\hline
1. &\texttt{time(0..5).}\\
2. & \texttt{isDrink(coffee).}\ \ \ \texttt{isDrink(wine).}\ \ \ \texttt{isDrink(water).}\\
3. & \texttt{happens(thirsty,1).}\ \ \ \texttt{happens(sunset,2).}\\
4. & \texttt{initiates(drink(coffee),energetic).}\ \ \ \texttt{initiates(gotoBed,asleep).}\\
5. &\texttt{ant(1,(Ts),Ts):-happens(drink(wine),Ts),time(Ts).}\\
6. &\texttt{cons(1,(T),T,Ts):-ant(1,(T),T), }\\
 & \mytab\mytab\mytab \texttt{happens(gotoBed,Ts),T+1=Ts,time(Ts).}\\
7. &\texttt{ant(2,(Ts),Ts):-happens(sunset,Ts),time(Ts).}\\
8. &\texttt{cons(2,(T),T,Ts):-ant(2,(T),T), }\\
 & \mytab\mytab\mytab \texttt{happens(gotoBed,Ts),T<Ts,Ts<=T+3,time(Ts).}\\
9. &\texttt{ant(3,(Ts),Ts):-happens(thirsty,Ts),time(Ts).}\\
10. &\texttt{cons(3,(T),T,Ts):-ant(3,(T),T),happens(drink(Liquid),Ts), }\\
 & \mytab\mytab\mytab \texttt{ isDrink(Liquid),T<Ts,Ts<T+3,time(Ts).}\\
11. & \texttt{supported(gotoBed,Ts):-ant(1,(T),T),time(Ts),T+1=Ts.}\\
12. & \texttt{supported(gotoBed,Ts):-ant(2,(T),T),T<Ts,Ts<=T+3,time(Ts).}\\
13. & \texttt{supported(drink(L),Ts):-ant(3,(T),T),isDrink(L),T<Ts,Ts<T+3,time(Ts).}\\
14. &\texttt{:-holds(asleep,Ts-1),happens(drink(Liquid),Ts),isDrink(Liquid),}\\
& \mytab \mytab \mytab \texttt{time(Ts),time(Ts-1).}\\
15. & \texttt{:-holds(asleep,Ts-1),happens(gotoBed,Ts),time(Ts), time(Ts-1).}\\
16. & \texttt{:-holds(energetic,Ts-1),happens(gotoBed,Ts),time(Ts), time(Ts-1).}\\
17. & \texttt{\mbox{$:\sim$}happens(drink(L),T), isDrink(L),time(T).[1@1,T,L]}\\
18. & \texttt{\mbox{$:\sim$}happens(gotoBed,T),time(T).[-T@2,T]}\\
& \textit{\% Reactive rule constraint, choice rule for actions, and the event theory are as before.}\\
\hline\end{tabular}
}
\caption{Prospective behaviour choosing a drink.}
\label{drinking}
\end{figure}

\section{Alternative KELPS Simulation}\label{incrementalASP}

This section presents an alternative incremental style mapping from $n$-distant KELPS to ASP (called {\em multi-shot}).  As in Section~\ref{sec:translation} we will consider $n$-distant 
KELPS frameworks and throughout we use as an example the KELPS 
framework shown in  Figure~\ref{incrementalExampleV2}. We assume that for the $n$-distance version 
appropriate temporal constraints are added to the two reactive rules, 
according to Definition~\ref{ndistantframework}.

\begin{figure}[hbpt]
\begin{tabular}{ll}\hline
Reactive Rules \textit{R1} and \textit{R2}\\
{\textit{R1}:} & $a(T) \rightarrow a1(T1) \wedge T<T1 \wedge T1<=T+10$\\
 & \mytab $\wedge$ $a2(T2) \wedge T2>T1\wedge T2<=T1+5$\\
{\textit{R2}:} & $b(T) \rightarrow b1(T1) \wedge T<T1$\\
{\textit{C$_{post}$}:} & $\mathit{terminates}(a1,p)$\\ 
 & $\mathit{initiates}(c,p)$\\
{\textit{C$_{pre}$}:} & $false \leftarrow b1(T+1) \wedge \neg p(T)$\\
Initial: & $p(0)$\\
Events: & $a(1)$ \ \ \ \ \ $b(5)$\ \ \ \ \ $c(9)$\\
& In the above $a$, $a1$, $a2$, $b$,  $b1$ and $c$ are events and $p$ is a fluent.\\
\hline \end{tabular}
\caption{KELPS Framework for exemplifying incremental simulation.}
\label{incrementalExampleV2}
\end{figure}

\subsection{An Incremental (Multi-shot) Mapping for Computing $n$-distant Models}\label{incrementalmap}
In this appendix we make use of \textit{clingo}~4 to incrementally increase  the timeframe, and to incrementally generate  groundings.
We use a predicate \texttt{maxRange}/1 to represent the horizon for the increasing timespan. 

The mapping from $n$-distant KELPS to Reactive ASP given in Section~\ref{sec:translation} employed a global timeframe from time 0 to 
time  $n$. We refer to that mapping as $f_{global}$.  
In this section we describe a second, incremental mapping, which we refer  to as $f_{inc}$.  
The difference between the two mappings can most clearly be seen at a qualitative level by considering the translation of a very simple reactive rule of the form  
\begin{equation}\label{verysimplerule}
{\textit{R}:}
\ \  event(T) \rightarrow action(T_1) \wedge T+1 < T_1
\end{equation}
\noindent by \texttt{ant} and \texttt{cons} rules as well as a reactive rule constraint.
First of all, recall that using $f_{global}$ the translation would include the rules
 
\begin{equation}\label{eq72}
{\small \begin{array}{l}
\texttt{ant(id,(Ts),Ts):-happens(event,Ts),time(Ts).}\\
\texttt{cons(id,(T),T,Ts):-ant(id,(T),T),happens(action,Ts),T+1<Ts,time(Ts).}\\
\texttt{:-ant(Id,X,Ts),not consTrue(Id,X,Ts),time(Ts).}\\
\texttt{consTrue(Id,X,Ts):-cons(Id,X,Ts,Ts1),time(Ts1).}\\
 \texttt{supported(action,Ts):-ant(id,(T),T),T+1<Ts,time(Ts).}
 \end{array}}
\end{equation}

\noindent Note that (in this example) in the definition of \texttt{cons} the head of the rule has a timestamp that  is at least two timestamps after the timestamp of the matching \texttt{ant} in the body, but all appropriate groundings of the rules are considered at once. In particular, this means that the time variables are constrained to be in the range [$0,\ldots,n$], as given by the facts \texttt{time(0..n)}.

On the other hand, the basic mapping $f_{inc}$ (for a fixed value of $n$ given by \texttt{maxRange(n)}) enlists the parameterised subprogram feature of \textit{clingo}~4. 
The resulting (multi-shot) program consists of two subprograms (or modules), which we  call \texttt{base} and \texttt{cycle(t)} declared by the \texttt{\#external program} command and whose grounding is under control of a procedure `Control' 
(shown in Figure~\ref{alg:test} and described in subsection~\ref{specificprocedure}).
The \texttt{base} subprogram is a standard  ASP program representing 
the initial state, the auxiliary facts $\Aux$~and $C_{post}$.
The \texttt{cycle(t)} subprogram consists of rules for \texttt{ant}, \texttt{cons} and \texttt{supported}, the  reactive rule integrity constraint, and the choice rule, as well as the event theory $ET$ and $C_{pre}$. The timestamp argument for all these is \texttt{t}.
The procedure  Control will generate an instantiation of \texttt{cycle(t)}  for each value of \texttt{t} in the program's overall timeframe (i.e. \texttt{t} =1,2,3,\ldots n), where $n$ is fixed  by a \texttt{maxRange} fact in the module \texttt{base}.
Each new instantiation is added incrementally to the pre-existing program. So, if the program models $n$ cycles, it will eventually include the modules \texttt{base} and \texttt{cycle(1)}, \texttt{cycle(2)},..., \texttt{cycle($n$)}. 
The (ground) program is re-solved after each cumulative expansion. 
Considering the reactive rule in (\ref{verysimplerule}), but now using the mapping $f_{inc}$, the mapped rules, which will be in  \texttt{cycle(t)}, will have the form in (\ref{eq71})

\begin{equation}\label{eq71}
{\small \begin{array}{l}
\texttt{ant(id,(t),t):-happens(event,t).} \\
\texttt{cons(id,(T),T,t):-ant(id,(T),T),happens(action,t),T+1<t.}\\
\texttt{:-query(t), ant(Id,X,T),not consTrue(Id,X,T,t).}\\
\texttt{consTrue(Id,X,T,t):-cons(Id,X,T,T1).}\\
\texttt{supported(action,t):-ant(id,(T),T),T+1<t.}
 \end{array}}
\end{equation}

\noindent As before, in the definitions of \texttt{ant} and \texttt{cons} the final parameter (\texttt{t}) is the timestamp at which the head atom of rule becomes true.
The atom \texttt{query(t)} is an external atom declared in ASP by  the command \texttt{\#external query(t)} and whose truth value can be manipulated by the procedure Control (see Figure~\ref{alg:test}).
Consider, for example, the grounding of the constraint when \texttt{t}=3 and \texttt{query(3)}=$True$, which will require that for every previously true ground instance of \texttt{ant(ID,X,T)} (i.e. \texttt{T}$\leq 3$),  there must be a true atom \texttt{cons(ID,X,T,T1)}, where \texttt{T1} takes a value in the range [$0,\ldots,3$].
There was a previous grounding of the constraint, when \texttt{t}=2 and \texttt{query(2)}= $True$, which required for every previously true ground instance of \texttt{ant(ID,X,T)} (i.e. \texttt{T}$\leq 2$), that there must be a true atom \texttt{cons(ID,X,T,T1)}, where \texttt{T1} takes a value in the range [$0,\ldots,2$]. It can be seen that for each value of \texttt{t} the constraint requires that for all previous true atoms  of \texttt{ant(ID,X,T)} (\texttt{T}$\leq $\texttt{t}) there must be a true atom  \texttt{cons(ID,X,T,T1)}, where \texttt{T1} takes a value in the range \texttt{[0,...,t]}.
Eventually, when \texttt{t}=$n$ the same set of instances of the constraint will have been considered as for the global mapping.

Using this incremental grounding, $f_{inc}$, the mapped rules for the example in Figure~\ref{incrementalExampleV2} are shown in Figure~\ref{incsimpleV2}. To map the observed external events $ext^*=\{a(1), b(5), c(9)\}$,   the facts \texttt{happens(a,1)},  \texttt{happens(b,5)}  and \texttt{happens(c,9)}  can be made available to the modules \texttt{cycle(1)}, \texttt{cycle(5)}  and \texttt{cycle(9)}, as shown in Lines 6a-6c  in Figure~\ref{incsimpleV2}.  The maximum timespan  is fixed by Lines 1a and 1b (for illustration we used 10). This can be overrridden by another value on the program call to \textit{clingo}. The \texttt{extratime} atom   used  in the body of the definition for \texttt{supported} in Line 14  ensures that there is adequate future time for the remaining parts of the (consequent of the reactive) rule to be made true. Here, the existential variable \texttt{T2} is a time in the future of \texttt{t}, and is constrained to be within the maximum range of 10. 

\begin{figure}[hbtp]
{\small
\begin{tabular}{ll} \hline
0. & \texttt{\#program base
\ \ \ \ \ \ \ \ \ \ \ \ \ \ \ \ \ \ \ \ \ \ \ \ \ \ \ \ \ \ \ \ \ \ \ \ \ \ \ \ \ \ \ \ \ \ \ \ \ \ \ \ \ \ \ \ \ \ \ \ \ \ \ \ \ \ }\\
1a. & \texttt{\#const m=10.}\\
1b. & \texttt{maxRange(m).}\\
2. & \texttt{extratime(0..X):-maxRange(X).}\\
3. & \texttt{holds(p,0).}\\
4a.  &\texttt{terminates(a1,p).}\\
4b. &\texttt{initiates(c,p).}\\
5. & \texttt{\#program cycle(t).}\\
6a. & \texttt{happens(a,t):-t=1.} \\
6b. & \texttt{happens(b,t):-t=5.}\\
6c. & \texttt{happens(c,t):-t=9.}\\
7. & \texttt{\#external query(t).}\\		
8a. &\texttt{:-query(t),ant(ID,X,T),not consTrue(ID,X,T,t).}\\
8b. &\texttt{consTrue(ID,X,T,t):-cons(ID,X,T,T1).}\\
9. &\texttt{0\{happens(Act,t)\}1:-supported(Act,t).}\\
10. &\texttt{ant(1,(t),t):-happens(a,t).}\\
11. &\texttt{cons(1,(T1),T1,t):-ant(1,(T1),T1),happens(a1,T2),T1<T2,}\\
& \mytab\mytab  \texttt{T2<=T1+10,T2<t,t<=T2+5,happens(a2,t).}\\
12. &\texttt{ant(2,(t),t):-happens(b,t).}\\
13. &\texttt{cons(2,(T1),T1,t):-ant(2,(T1),T1),happens(b1,t),T1<t.}\\
14. &\texttt{supported(a1,t):-ant(1,(T1),T1),T1<t,t<=T1+10,}\\
 & \mytab\mytab \texttt{T2>t,T2<=t+5,extratime(T2).}\\
15. &\texttt{supported(a2,t):-ant(1,(T1),T1),happens(a1,T2),}\\
& \mytab\mytab  \texttt{T1<T2,T2<t,T2<=T1+10,t<=T2+5.}\\
16. &\texttt{supported(b1,t):-ant(2,(T1),T1),T1<t.}\\
17. &\texttt{:-happens(b1,t),not holds(p,t-1).}\\
18a. &\texttt{holds(P,t):-initiates(E,P),happens(E,t).}\\	
18b. &\texttt{holds(P,t):-holds(P,t-1),not broken(P,t).}\\
18c. &\texttt{broken(P,t):-terminates(E,P),happens(E,t).}\\		
\hline\end{tabular}
}
\caption{Translation by $f_{inc}$ of example in Figure~\ref{incrementalExampleV2}.}
\label{incsimpleV2}
\end{figure}

\subsection{The Procedural Control}\label{specificprocedure}

To implement the incremental variant of Reactive ASP we have used clingo~4 
\cite{clingo_guide}, with a simple Lua script, similar to that given in \cite{multishot} (see Figure~\ref{alg:test}).
The input $max$ to the procedure is extracted from the constant $m$ on Line 1a of the ASP program.%

\begin{figure}[htbp]
\small\begin{algorithm}[H]
	\SetAlgoLined
	{\sc PROCEDURE} Control ($max$);\\
	$t \leftarrow 0;$\\
	\While{$t \leq max$}{
         \eIf{$t = 0$}{
{\textit{INSERT `base' subprogram to  program;}}\\
}{
\textit{INSERT `cycle(t)' subprogram to  program;}\\
}
	$GROUND\ the \ program$;\\
	\If{$t > 0$}{
$query(t) \leftarrow TRUE$; \\
	}
$SOLVE\ the\ program's\ answer\ sets$; \\
		\If{$t > 0$}{
$query(t) \leftarrow FALSE; $\\
		}
$ t \leftarrow t+1$; 	
}
\end{algorithm}
\caption{Pseudocode procedural control.}\label{alg:test}
\end{figure}
\normalsize

In the first cycle ($t=0$) the \texttt{base} program is solved, capturing the initial state, $\Aux$ and information about postconditions of events. 
In the next cycle  ($t=1$) the procedure does the following steps.
It sets \texttt{query(1)} to $True$, effectively `activating' the reactive rule constraint  for the timeframe 0 to 1 (see Lines 8a and 8b in Figure~\ref{incsimpleV2}) and attempts to find the program's answer sets. 
To look for answer sets in subsequent timeframes ($t=1,\ldots, max$) it is necessary to `switch off' the constraint for the timeframe \texttt{0} to \texttt{1} and to re-try it with an expanded timeframe. To this end, the procedural control sets \texttt{query(1)} to $False$, sets \texttt{query(2)} to $True$ and adds the instantiated \texttt{cycle(2)}, and tries to find an answer set without redoing the grounding of the previous iteration.
This loop continues  until \texttt{t}=$max$.\footnote{Notice that an added advantage of such a procedure is that from \texttt{t}=1  onwards external events 
at time \texttt{t} may be added within the loop to \texttt{cycle(t)}, as the knowledge of their occurrence becomes available, allowing a more reactive and situated behaviour as in KELPS. As we saw in Section~\ref{sec:combined} this was also possible in the Hybrid KELPS/ASP variant.}

By way of illustration, some relevant parts of the  output   of the program in Figure~\ref{incsimpleV2} with the control in Figure~\ref{alg:test} are described next for the value of the constant $m$ set to 7.
There is one answer set for $t=0$, namely {\small\texttt{\{holds(p,0)\}}}.\footnote{Note that the answer set includes facts in the \texttt{base} subprogram. Also, for ease of reading we only show the facts for \texttt{holds} and \texttt{happens} in the answer sets.} There is no answer set for $t=1$ or $t=2$ because the agent cannot yet satisfy the reactive rule triggered by {\small\texttt{happens(a,1)}}. 
However, for  $t=3$ the agent can perform the actions \texttt{a1} and \texttt{a2} and satisfy the rule consequent of the reactive rule with identifier 1.   The resulting answer set includes the atoms {\small\texttt{\{happens(a,1),happens(a1,2),happens(a2,3),holds(p,0),}} {\small\texttt{holds(p,1)\}}}. 
For  $t=4$ there are two answer sets, depending on whether action \texttt{a1} occurs at time 2, or time 3. For $t=5$ there is no answer set since there is no time to perform \texttt{b1} to satisfy the reactive rule with identifier 2. Moreover, in order for \texttt{b1} to occur the fluent \texttt{p} must be true in the previous time instant, hence action  \texttt{a1} must not have occurred at least until time 6. This prevents any answer set  for $t=6$ as there is no time to satisfy reactive rule with identifier 1. 
There is an answer set for $t=7$, namely {\small\texttt{\{happens(a,1),happens(a1,6),happens(a2,7)},  \texttt{happens(b,5),happens(b1,6),holds(p,0),holds(p,1),holds(p,2),holds(p,3),holds(p,4)}, \texttt{holds(p,5)\}}}.
Note that for each value of $t$, where there is an answer set it 
corresponds to a $t$-distant KELPS reactive model - and vice versa.

\subsection{Experiments}\label{Experiments} 
We show here results of two experiments. The first experiment checks the impact of the explicit frame axiom in the standard mapping of Section~\ref{sec:translation}, while the second experiment compares the standard mapping with the incremental version of Section~\ref{incrementalmap}.

The first experiment (actually two sub-experiments 1a and 1b), involved a standard mapping program with the potential to vary the number of fluents. Specifically, the fluents belonged to $\{p(X),q(X),r(X),s(X),t(X),j(X) : 1\leq X\leq 200\}$. There were 6 external events,  $\{c, d, e, f, g, h\}$, each having 9 occurrences at varying times between 1 and 100. The action postconditions were $initiates(c,p(X))$, $initiates(d,q(X))$, $initiates(e,r(X))$, $initiates(f,s(X))$, $initiates(g,t(X))$, $initiates(h,j(X))$,   and $terminates(c,q(X))$, \break $terminates(d,r(X))$, $terminates(e,p(X))$, $terminates(f,j(X))$, $terminates(g,s(X))$, \break $terminates(h,t(X))$,  for $1\leq X\leq 200$.\footnote{For example, given the facts \texttt{index(1..20)}, the rule \texttt{initiates(c,p(X)):-index(X)} will generate \texttt{initiates(c,p(1))}, \texttt{initiates(c,p(2))} up to \texttt{initiates(c,p(20))}.} There were no reactive rules.

Experiment 1a    fixed the maximum timestamp   at 200  and varied the number of fluents by changing the number of indices ($X$) in steps of 20 up to 200. Experiment 1b fixed the number of fluents at 1200 ($X=200$) and ran for maximum timestamps ($n$) varying from 100 up to 1000 in steps of 100. Results of both experiments are shown in Table~\ref{Tab:Expt1}. It is clear that, as expected, the total runtime varies linearly with the number of fluents and with the maximum timestamp.
\begin{table}[hbtp]
$\begin{array}{c c}
\fbox{
$\begin{array}{ll | ll}
\mbox{\#fluents} & \mbox{Time(s)}&\mbox{\#fluents} & \mbox{Time(s)}\\
\hline
20 & \mbox{0.038} & 120 & \mbox{0.232} \\
40 & \mbox{0.074}& 140 & \mbox{0.267} \\
60 & \mbox{0.109}& 160 & \mbox{0.303} \\
80 & \mbox{0.140} & 180 & \mbox{0.333} \\
100 & \mbox{0.184}  & 200 & \mbox{0.366}
\end{array}$}
&
\fbox{
$\begin{array}{ll | ll}
\mbox{$n$} & \mbox{Time(s)} & \mbox{$n$} & \mbox{Time(s)}\\
\hline
100 & \mbox{0.199} & 600 & \mbox{1.193}\\
200 & \mbox{0.377} & 700 & \mbox{1.394}\\
300 & \mbox{0.579} & 800 & \mbox{1.698}\\
400 & \mbox{0.784} & 900 & \mbox{1.816}\\
500 & \mbox{0.997} & 1000 & \mbox{2.06}
 \end{array}$}
\end{array}$
\caption{Results of Experiment 1a (left) and Experiment 1b (right).}
\label{Tab:Expt1}
\end{table}

The second experiment compared run times of the standard mapping from Section~\ref{sec:translation} with those of the incremental version. The program was an extended version of that shown in Figure~\ref{incsimpleV2}, with external events $a(1) \ldots a(3)$ and $b(1) \ldots b(3)$ occurring at various times. In this experiment, as in Figure B2,  there is again one nullary fluent $p$. Just as in Figure~\ref{incsimpleV2}, where the event $a$ triggers events $a1$ and $a2$, in this extended version event $a(I)$ triggers $a1(I)$ and $a2(I)$.
The reactive rules are the same except that the events $a$, $a1$, $a2$, $b$ and $b1$ are modified by adding an index ranging from 1 to 3 and adjustments made accordingly.
The experiment ran the standard Reactive ASP code for every value of maximum timestamp 
between 1 and 50 and accumulated the execution times in groups of 10 or 20. It compared the results for the incremental translation  for maxRange  ($m$) values in $\{10,20,40,50\}$. The results are shown in Table~\ref{Tab:Expt2}.

The external events occurred at the following times: $\{(a(1),1), (b(1), 5), (a(2), 11), \break (b(2), 15), (a(3), 32), (b(3), 35), (c,9),(c,19),(c,29),(c,39),(c,49)\}$. All indexed $a1$ events terminate fluent $p$ and all indexed $b1$ events require the fluent $p$ to hold at the previous time.  The purpose of event $c$ was to (re)initiate the fluent $p$ after it had been terminated, ready for each subsequent pair of events $a$ and $b$. Additionally, some constraints were imposed to limit the number of occurrences of actions $a1(I), a2(I), b1(I)$, $I \in \{1,2,3\}$, to a maximum of one each and to ensure that the reactions  $a1(I)$, $a2(I)$ and $b1(I)$ to an  occurrence of  events $a(I)$ and $b(I)$ occur before the  occurrence of $a$ and $b$ with the next index ( i.e. $a(I+1)$ and  $b(I+1)$) {\em --} see equation~(\ref{eqIncConstraints}), shown for the incremental version.
\begin{equation}\label{eqIncConstraints}
{\small \begin{array}{l}
\texttt{:-happens(a1(X),T),happens(a1(X),t),T<t,index(X).}\\
\texttt{:-happens(a2(X),T),happens(a2(X),t),T<t,index(X).}\\
\texttt{:-happens(b1(X),T),happens(b1(X),t),T<t,index(X).}\\
 \texttt{:-happens(b1(X),t),happens(a2(X),T1),t>T1,index(X).}
 \end{array}}
\end{equation}

\begin{table}[hbtp]
\fbox{$\begin{array}{l | l | l}
\mbox{$m$} & \mbox{Total Standard Time(s)}& \mbox{Incremental Time(s)}\\
\hline
10 & \mbox{0.07}& \mbox{0.04} \\
20 & \mbox{0.185}  & \mbox{0.21}\\
40 & \mbox{6.065} &  \mbox{8.28} \\
50 & \mbox{171.865}  & \mbox{287.8} 
\end{array}$}
\caption{Results of Experiment 2.}
\label{Tab:Expt2}
\end{table}

The results show that the cumulative time for $n$ standard runs is lower than the incremental time. As mentioned earlier, it is to be expected that the times be comparable as the incremental program has to re-evaluate the set of models for each value of $t$ up to $n$. The results show there is a small additional overhead.


%% file: ms.bbl
\begin{thebibliography}{}

\bibitem[\protect\citeauthoryear{Aguado, Cabalar, Dieguez, Perez, and
  Vidal}{Aguado et~al\mbox{.}}{2013}]{TEL}
{\sc Aguado, F.}, {\sc Cabalar, P.}, {\sc Dieguez, M.}, {\sc Perez, G.}, {\sc
  and} {\sc Vidal, C.} 2013.
\newblock Temporal equilibrium logic: a survey.
\newblock {\em Journal of Applied Non-Classical Logics\/}~{\em 23,\/}~1-2,
  2--24.

\bibitem[\protect\citeauthoryear{Alferes, Banti, and Brogi}{Alferes
  et~al\mbox{.}}{2006}]{Alferes}
{\sc Alferes, J.}, {\sc Banti, F.}, {\sc and} {\sc Brogi, A.} 2006.
\newblock An event-condition-action logic programming language.
\newblock In {\em 10th European Conference on Logics in Artificial
  Intelligence}. 29--42.

\bibitem[\protect\citeauthoryear{Anh and Pereira}{Anh and
  Pereira}{2011}]{prospective2}
{\sc Anh, H.} {\sc and} {\sc Pereira, L.} 2011.
\newblock Intention-based decision making with evolution prospection.
\newblock In {\em Progress in Artificial Intelligence, 15th Portuguese
  International Conference on Artifical Intelligence (EPIA 2011)}. 254--267.

\bibitem[\protect\citeauthoryear{Baral and Son}{Baral and
  Son}{1998}]{ActionTheories}
{\sc Baral, C.} {\sc and} {\sc Son, T.~C.} 1998.
\newblock Relating theories of actions and reactive control.
\newblock {\em Electronic Transactions on Artificial Intelligence\/}~{\em 2},
  211--271.

\bibitem[\protect\citeauthoryear{Beck, Dao{-}Tran, and Eiter}{Beck
  et~al\mbox{.}}{2018}]{Streams}
{\sc Beck, H.}, {\sc Dao{-}Tran, M.}, {\sc and} {\sc Eiter, T.} 2018.
\newblock {LARS:} {A} logic-based framework for analytic reasoning over
  streams.
\newblock {\em Artificial Intelligence\/}~{\em 261}, 16--70.

\bibitem[\protect\citeauthoryear{Berstel-Da~Silva}{Berstel-Da~Silva}{2012}]{daSilva}
{\sc Berstel-Da~Silva, B.} 2012.
\newblock Formalizing both refraction-based and sequential executions of
  production rule programs.
\newblock In {\em Rules on the Web: Research and Applications}, {A.~Bikakis}
  {and} {A.~Giurca}, Eds. Springer Berlin Heidelberg, 47--61.

\bibitem[\protect\citeauthoryear{Brewka}{Brewka}{2013}]{Brewka}
{\sc Brewka, G.} 2013.
\newblock Towards reactive multi-context systems.
\newblock In {\em Logic Programming and Nonmonotonic Reasoning. LPNMR 2013}.

\bibitem[\protect\citeauthoryear{Brewka, Eiter, and Truszczy\'{n}ski}{Brewka
  et~al\mbox{.}}{2011}]{ASP_Glance}
{\sc Brewka, G.}, {\sc Eiter, T.}, {\sc and} {\sc Truszczy\'{n}ski, M.} 2011.
\newblock Answer set programming at a glance.
\newblock {\em Commun. ACM\/}~{\em 54,\/}~12, 92--103.

\bibitem[\protect\citeauthoryear{Cabalar, Kaminski, Schaub, and
  Schuhmann}{Cabalar et~al\mbox{.}}{2018}]{temporal}
{\sc Cabalar, P.}, {\sc Kaminski, R.}, {\sc Schaub, T.}, {\sc and} {\sc
  Schuhmann, A.} 2018.
\newblock Temporal answer set programming on finite traces.
\newblock {\em Theory and Practice of Logic Programming\/}~{\em 18,\/}~3-4,
  406--420.

\bibitem[\protect\citeauthoryear{Calimeri, Faber, Gebser, Ianni, Kaminski,
  Krennwallner, Leone, Maratea, Ricca, Schaub, and et~al.}{Calimeri
  et~al\mbox{.}}{2020}]{ASPCore2}
{\sc Calimeri, F.}, {\sc Faber, W.}, {\sc Gebser, M.}, {\sc Ianni, G.}, {\sc
  Kaminski, R.}, {\sc Krennwallner, T.}, {\sc Leone, N.}, {\sc Maratea, M.},
  {\sc Ricca, F.}, {\sc Schaub, T.}, {\sc and} {\sc et~al.} 2020.
\newblock Asp-core-2 input language format.
\newblock {\em Theory and Practice of Logic Programming\/}~{\em 20,\/}~2,
  29--309.

\bibitem[\protect\citeauthoryear{Clark}{Clark}{2018}]{Keith1}
{\sc Clark, K.} 2018.
\newblock Rule control of teleo-reactive, multi-tasking, communicating robotic
  agents.
\newblock In {\em Proceedings of 15th International Conference on Informatics
  in Control, Automation and Robotics, {ICINCO} 2018}. 5--15.

\bibitem[\protect\citeauthoryear{Clark and Robinson}{Clark and
  Robinson}{2015}]{Keith2}
{\sc Clark, K.} {\sc and} {\sc Robinson, P.} 2015.
\newblock Robotic agent programming in teleo{R}.
\newblock In {\em Proceedings of IEEE International Conference on Robotics and
  Automation}. 5040--5047.

\bibitem[\protect\citeauthoryear{Costantini}{Costantini}{2011}]{Costantini2011}
{\sc Costantini, S.} 2011.
\newblock Answer set modules for logical agents.
\newblock In {\em Datalog Reloaded: First International Workshop, Datalog 2010,
  Oxford, UK, March 16-19, 2010. Revised Selected Papers}. 37--58.

\bibitem[\protect\citeauthoryear{Costantini, De~Gasperis, and
  Nazzicone}{Costantini et~al\mbox{.}}{2015}]{ConstantiniExploration}
{\sc Costantini, S.}, {\sc De~Gasperis, G.}, {\sc and} {\sc Nazzicone, G.}
  2015.
\newblock Exploration of unknown territory via dali agents and asp modules.
\newblock In {\em Distributed Computing and Artificial Intelligence, 12th
  International Conference}. Springer, 285--292.

\bibitem[\protect\citeauthoryear{Costantini and Tocchio}{Costantini and
  Tocchio}{2004}]{DALI}
{\sc Costantini, S.} {\sc and} {\sc Tocchio, A.} 2004.
\newblock The {DALI} logic programming agent-oriented language.
\newblock In {\em 9th European Conference on Logics in Artificial
  Intelligence}. 685--688.

\bibitem[\protect\citeauthoryear{Deane}{Deane}{2016}]{Graham}
{\sc Deane, G.} 2016.
\newblock Preferential description logics: Reasoning in the presence of
  inconsistencies.
\newblock Ph.D. thesis, Imperial College London.

\bibitem[\protect\citeauthoryear{Dyoub, Costantini, and De~Gasperis}{Dyoub
  et~al\mbox{.}}{2018}]{ConstantiniSurvey}
{\sc Dyoub, A.}, {\sc Costantini, S.}, {\sc and} {\sc De~Gasperis, G.} 2018.
\newblock Answer set programming and agents.
\newblock {\em The Knowledge Engineering Review\/}~{\em 33}.

\bibitem[\protect\citeauthoryear{Eiter, Fink, Sabbatini, and Tompits}{Eiter
  et~al\mbox{.}}{2004}]{Eiter2004}
{\sc Eiter, T.}, {\sc Fink, M.}, {\sc Sabbatini, G.}, {\sc and} {\sc Tompits,
  H.} 2004.
\newblock {\em Declarative Update Policies for Nonmonotonic Knowledge Bases}.
\newblock Springer Berlin Heidelberg, 85--129.

\bibitem[\protect\citeauthoryear{Erdem, Gelfond, and Leone}{Erdem
  et~al\mbox{.}}{2016}]{applications}
{\sc Erdem, E.}, {\sc Gelfond, M.}, {\sc and} {\sc Leone, N.} 2016.
\newblock Applications of answer set programming.
\newblock {\em AI Magazine\/}~{\em 37,\/}~3, 53--68.

\bibitem[\protect\citeauthoryear{Fernandes, Williams, and Paton}{Fernandes
  et~al\mbox{.}}{1997}]{Fernandes}
{\sc Fernandes, A.}, {\sc Williams, M.}, {\sc and} {\sc Paton, N.} 1997.
\newblock A logic-based integration of active and deductive databases.
\newblock {\em New Generation Computing\/}~{\em 15,\/}~2, 205--244.

\bibitem[\protect\citeauthoryear{Fr{\"u}hwirth}{Fr{\"u}hwirth}{1998}]{fruhwirth}
{\sc Fr{\"u}hwirth, T.} 1998.
\newblock Theory and practice of {C}onstraint {H}andling {R}ules.
\newblock {\em J. Logic Programming, Special Issue on Constraint Logic
  Programming\/}~{\em 37,\/}~1--3, 95--138.

\bibitem[\protect\citeauthoryear{Gebser, Grote, Kaminski, and Schaub}{Gebser
  et~al\mbox{.}}{2011}]{oclingo}
{\sc Gebser, M.}, {\sc Grote, T.}, {\sc Kaminski, R.}, {\sc and} {\sc Schaub,
  T.} 2011.
\newblock Reactive answer set programming.
\newblock In {\em Logic Programming and Nonmonotonic Reasoning}, {J.~P.
  Delgrande} {and} {W.~Faber}, Eds. 54--66.

\bibitem[\protect\citeauthoryear{Gebser, Kaminski, Kaufmann, Lindauer,
  Ostrowski, Romero, Schaub, Thiele, and Wanko}{Gebser
  et~al\mbox{.}}{2019}]{clingo_guide}
{\sc Gebser, M.}, {\sc Kaminski, R.}, {\sc Kaufmann, B.}, {\sc Lindauer, M.},
  {\sc Ostrowski, M.}, {\sc Romero, J.}, {\sc Schaub, T.}, {\sc Thiele, S.},
  {\sc and} {\sc Wanko, P.} 2019.
\newblock Potassco user guide, version 2.2.0.
\newblock {\em Institute for Informatics, University of Potsdam, second
  edition\/}.

\bibitem[\protect\citeauthoryear{Gebser, Kaminski, Kaufmann, and Schaub}{Gebser
  et~al\mbox{.}}{2019a}]{multishot}
{\sc Gebser, M.}, {\sc Kaminski, R.}, {\sc Kaufmann, B.}, {\sc and} {\sc
  Schaub, O.} 2019a.
\newblock Multi-shot asp solving with clingo.
\newblock {\em Theory and Practice of Logic Programming\/}~{\em 19,\/}~1,
  27--82.

\bibitem[\protect\citeauthoryear{Gebser, Kaminski, Kaufmann, and Schaub}{Gebser
  et~al\mbox{.}}{2013}]{ASP}
{\sc Gebser, M.}, {\sc Kaminski, R.}, {\sc Kaufmann, B.}, {\sc and} {\sc
  Schaub, T.} 2013.
\newblock {\em Answer set solving in practice}.
\newblock Morgan \& Claypool.

\bibitem[\protect\citeauthoryear{Gebser, Kaminski, Kaufmann, and Schaub}{Gebser
  et~al\mbox{.}}{2019b}]{Recenticlingo}
{\sc Gebser, M.}, {\sc Kaminski, R.}, {\sc Kaufmann, B.}, {\sc and} {\sc
  Schaub, T.} 2019b.
\newblock Multi-shot {ASP} solving with clingo.
\newblock {\em Theory and Practice of Logic Programming\/}~{\em 19,\/}~1,
  27--82.

\bibitem[\protect\citeauthoryear{Gelfond}{Gelfond}{2007}]{GelfondChapter}
{\sc Gelfond, M.} 2007.
\newblock Chapter 7 answer sets.
\newblock In {\em Handbook of Knowledge Representation}, {F.~van Harmalen},
  {V.~Lifschita}, {and} {B.~Porter}, Eds. Elsevier Science.

\bibitem[\protect\citeauthoryear{Gelfond and Lifschitz}{Gelfond and
  Lifschitz}{1988}]{Gelfond1988}
{\sc Gelfond, M.} {\sc and} {\sc Lifschitz, V.} 1988.
\newblock The stable model semantics for logic programming.
\newblock In {\em ICLP/SLP}. Vol.~88. 1070--1080.

\bibitem[\protect\citeauthoryear{Gurevich}{Gurevich}{2000a}]{gurevich}
{\sc Gurevich, Y.} 2000a.
\newblock Sequential abstract-state machines capture sequential algorithms.
\newblock {\em ACM Trans. Comput. Logic\/}~{\em 1,\/}~1, 77--111.

\bibitem[\protect\citeauthoryear{Gurevich}{Gurevich}{2000b}]{ASMs}
{\sc Gurevich, Y.} 2000b.
\newblock Sequential abstract-state machines capture sequential algorithms.
\newblock {\em ACM Transactions on Computational Logic\/}, 77--111.

\bibitem[\protect\citeauthoryear{Kowalski and Sadri}{Kowalski and
  Sadri}{1999}]{Early}
{\sc Kowalski, R.} {\sc and} {\sc Sadri, F.} 1999.
\newblock From logic programming towards multi-agent systems.
\newblock {\em Annals of Mathematics and Artificial Intelligence\/}~{\em 25},
  391--419.

\bibitem[\protect\citeauthoryear{Kowalski and Sadri}{Kowalski and
  Sadri}{2011}]{LPSAbductive}
{\sc Kowalski, R.} {\sc and} {\sc Sadri, F.} 2011.
\newblock Abductive logic programming agents with destructive databases.
\newblock {\em Annals of Mathematics and Artificial Intelligence\/}~{\em
  62,\/}~1-2, 129--58.

\bibitem[\protect\citeauthoryear{Kowalski and Sadri}{Kowalski and
  Sadri}{2015}]{LPSReactive}
{\sc Kowalski, R.} {\sc and} {\sc Sadri, F.} 2015.
\newblock Reactive computing as model generation.
\newblock {\em New Generation Computing\/}~{\em 33,\/}~1, 33--67.

\bibitem[\protect\citeauthoryear{Kowalski and Sadri}{Kowalski and
  Sadri}{2016}]{kelps_paper}
{\sc Kowalski, R.} {\sc and} {\sc Sadri, F.} 2016.
\newblock Programming in logic without logic programming.
\newblock {\em Theory and Practice of Logic Programming\/}~{\em 16,\/}~3,
  269--295.

\bibitem[\protect\citeauthoryear{Kowalski and Sergot}{Kowalski and
  Sergot}{1986}]{KowalskiSergot}
{\sc Kowalski, R.} {\sc and} {\sc Sergot, M.} 1986.
\newblock A logic-based calculus of events.
\newblock {\em NewGeneration Computing\/}~{\em 4}, 67--95.

\bibitem[\protect\citeauthoryear{Lausen, Ludäscher, and May}{Lausen
  et~al\mbox{.}}{1998}]{Statelog}
{\sc Lausen, G.}, {\sc Ludäscher, B.}, {\sc and} {\sc May, W.} 1998.
\newblock On active deductive databases: The statelog approach.
\newblock {\em Transactions and Change in Logic Databases\/}, 69--106.

\bibitem[\protect\citeauthoryear{Law, Russo, and Broda}{Law
  et~al\mbox{.}}{2015}]{MarkLawReduct}
{\sc Law, L.}, {\sc Russo, A.}, {\sc and} {\sc Broda, K.} 2015.
\newblock Simplified reduct for choice rules in {ASP}, technical report
  {DTR}2015-2, imperial college london.

\bibitem[\protect\citeauthoryear{Law, Russo, and Broda}{Law
  et~al\mbox{.}}{2016}]{ILASP}
{\sc Law, M.}, {\sc Russo, A.}, {\sc and} {\sc Broda, K.} 2016.
\newblock Iterative learning of answer set programs from context dependent
  examples.
\newblock {\em Theory and Practice of Logic Programming\/}~{\em 16}, 834--848.

\bibitem[\protect\citeauthoryear{Lifschitz and Turner}{Lifschitz and
  Turner}{1994}]{Splitting}
{\sc Lifschitz, V.} {\sc and} {\sc Turner, H.} 1994.
\newblock Splitting a logic program.
\newblock In {\em Proceedings of the Eleventh International Conference on Logic
  Programming}. 23--37.

\bibitem[\protect\citeauthoryear{Mancarella, Terreni, Sadri, Toni, and
  Endriss}{Mancarella et~al\mbox{.}}{2009}]{CIFF}
{\sc Mancarella, P.}, {\sc Terreni, G.}, {\sc Sadri, F.}, {\sc Toni, F.}, {\sc
  and} {\sc Endriss, U.} 2009.
\newblock The {CIFF} proof procedure for abductive logic programming with
  constraints: Theory, implementation and experiments.
\newblock {\em Theory and Practice of Logic Programming\/}~{\em 9,\/}~6,
  691--750.

\bibitem[\protect\citeauthoryear{McCarthy}{McCarthy}{1998}]{Mccarthy98elaborationtolerance}
{\sc McCarthy, J.} 1998.
\newblock Elaboration tolerance.
\newblock In {\em In Working Papers of the Fourth International Symposium on
  Logical Formalizations of Commonsense Reasoning, Commonsense-1998.}

\bibitem[\protect\citeauthoryear{Nam and Baral}{Nam and Baral}{2007}]{Triggers}
{\sc Nam, T.} {\sc and} {\sc Baral, C.} 2007.
\newblock Reasoning about non-immediate triggers in biological networks.
\newblock {\em Annals of Mathematics and Artificial Intelligence\/}~{\em
  51,\/}~2-4, 267--293.

\bibitem[\protect\citeauthoryear{Nilsson}{Nilsson}{1994}]{Nilsson}
{\sc Nilsson, N.} 1994.
\newblock Teleo-reactive programs for agent control.
\newblock {\em Journal of Artificial Intelligence Research\/}~{\em 30}.

\bibitem[\protect\citeauthoryear{Paschke, Boley, Zhao, Teymourian, and
  Athan}{Paschke et~al\mbox{.}}{2012}]{RuleML}
{\sc Paschke, A.}, {\sc Boley, H.}, {\sc Zhao, Z.}, {\sc Teymourian, K.}, {\sc
  and} {\sc Athan, T.} 2012.
\newblock Reaction rule{ML} 1.0: Standardized semantic seaction sules.
\newblock In {\em Rules on the Web: Research and Applications}. Springer Berlin
  Heidelberg, 100--119.

\bibitem[\protect\citeauthoryear{Pereira and Lopes}{Pereira and
  Lopes}{2009}]{prospective}
{\sc Pereira, L.} {\sc and} {\sc Lopes, G.} 2009.
\newblock Prospective logic agents.
\newblock {\em International Journal of Reasoning-based Intelligent
  Systems\/}~{\em 1,\/}~3-4, 200--208.

\bibitem[\protect\citeauthoryear{Pereira and Saptawijaya}{Pereira and
  Saptawijaya}{2011}]{prospective3}
{\sc Pereira, L.} {\sc and} {\sc Saptawijaya, A.} 2011.
\newblock Modelling morality with prospective logic.
\newblock 98--421.

\bibitem[\protect\citeauthoryear{Rao}{Rao}{2009}]{Agentspeak}
{\sc Rao, A.} 2009.
\newblock Agentspeak (l): {BDI} agents speak out in a logical computable
  language.
\newblock {\em Agents Breaking Away\/}, 42--55.

\bibitem[\protect\citeauthoryear{Rao and Georgeff}{Rao and
  Georgeff}{1995}]{BDI}
{\sc Rao, A.} {\sc and} {\sc Georgeff, M.} 1995.
\newblock {BDI} agents: From theory to practice.
\newblock In {\em International Conference on Multiagent Systems}. 312--319.

\bibitem[\protect\citeauthoryear{Ribeiro, Inoue, and Bourgne}{Ribeiro
  et~al\mbox{.}}{2013}]{ribeiro}
{\sc Ribeiro, T.}, {\sc Inoue, K.}, {\sc and} {\sc Bourgne, G.} 2013.
\newblock Combining answer set programs for adaptive and reactive reasoning.
\newblock {\em Theory and Practice of Logic Programming\/}~{\em 13,\/}~4-5.

\bibitem[\protect\citeauthoryear{Russell and Norvig}{Russell and
  Norvig}{2003}]{Russell}
{\sc Russell, S.} {\sc and} {\sc Norvig, P.} 2003.
\newblock {\em Artificial intelligence - a modern approach, 2nd Edition}.
\newblock Prentice Hall series in artificial intelligence. Prentice Hall.

\bibitem[\protect\citeauthoryear{Sanchez, Alvarez, Morales, and
  Navarro}{Sanchez et~al\mbox{.}}{2016}]{Sanchez}
{\sc Sanchez, P.}, {\sc Alvarez, B.}, {\sc Morales, J.}, {\sc and} {\sc
  Navarro, P.~J.} 2016.
\newblock From teleo-reactive specifications to architectural components: A
  model-driven approach.
\newblock {\em Journal of Systems and Software\/}~{\em 117}, 317--333.

\bibitem[\protect\citeauthoryear{Suchan and Bhatt}{Suchan and
  Bhatt}{2019}]{mehul}
{\sc Suchan, J.} {\sc and} {\sc Bhatt, M.and~Varadarajan, S.} 2019.
\newblock Out of sight but not out of mind: An answer set programming based
  online abduction framework for visual sensemaking in autonomous driving.
\newblock In {\em Proceedings of the Twenty-Eighth International Joint
  Conference on Artificial Intelligence, IJCAI-19}. International Joint
  Conferences on Artificial Intelligence Organization, 1879--1885.

\bibitem[\protect\citeauthoryear{Tran and Baral}{Tran and
  Baral}{2004}]{BaralKR}
{\sc Tran, N.} {\sc and} {\sc Baral, C.} 2004.
\newblock Reasoning about triggered actions in ansprolog and its application to
  molecular interactions in cells.
\newblock In {\em Principles of Knowledge Representation and Reasoning:
  Proceedings of the Ninth International Conference (KR2004), Whistler,
  Canada}, {D.~Dubois}, {C.~A. Welty}, {and} {M.~Williams}, Eds. {AAAI} Press,
  554--564.

\bibitem[\protect\citeauthoryear{Vaseqi and Delgrande}{Vaseqi and
  Delgrande}{2013}]{maritime}
{\sc Vaseqi, Z.} {\sc and} {\sc Delgrande, J.} 2013.
\newblock An application of answer set programming for situational analysis in
  a maritime traffic domain.
\newblock {\em Advances in Artificial Intelligence. 26th Canadian Conference on
  Artificial Intelligence\/}, 315--22.

\bibitem[\protect\citeauthoryear{Wielemaker, Riguzzi, Kowalski, Lager, Sadri,
  and Calejo}{Wielemaker et~al\mbox{.}}{2019}]{SWISH}
{\sc Wielemaker, J.}, {\sc Riguzzi, F.}, {\sc Kowalski, R.}, {\sc Lager, T.},
  {\sc Sadri, F.}, {\sc and} {\sc Calejo, M.} 2019.
\newblock Using swish to realize interactive web-based tutorials for
  logic-based languages.
\newblock {\em Theory and Practice of Logic Programming\/}~{\em 19,\/}~2,
  229--261.

\bibitem[\protect\citeauthoryear{Zaniolo}{Zaniolo}{2003}]{activedatabases}
{\sc Zaniolo, C.} 2003.
\newblock On the unification of active databases and deductive databases.
\newblock {\em Advances in Databases\/}, 23--39.

\end{thebibliography}
